\documentclass{article} % For LaTeX2e
\usepackage{conference,times}
\usepackage{threeparttable}
\usepackage{graphicx}
\usepackage{wrapfig}
\usepackage{amsmath}
\usepackage{amssymb}
\usepackage{subcaption}
\usepackage{pifont}
\usepackage{svg}
\usepackage{bbm}
\usepackage{makecell}
\usepackage{booktabs, colortbl, xcolor, multirow, soul}
\usepackage{amsmath,amsfonts,bm}

\def\eqref#1{equation~\ref{#1}}
\def\1{\bm{1}}

\DeclareMathAlphabet{\mathsfit}{\encodingdefault}{\sfdefault}{m}{sl}
\SetMathAlphabet{\mathsfit}{bold}{\encodingdefault}{\sfdefault}{bx}{n}

\usepackage{hyperref}
\usepackage{url}
\usepackage[nameinlink,capitalize]{cleveref}
\usepackage{xspace}

\definecolor{oursblue}{RGB}{218,232,252}
\definecolor{groupgray}{RGB}{240,240,240}
\definecolor{oursblue1}{RGB}{245,248,254}
\definecolor{oursblue2}{RGB}{236,243,253}
\definecolor{oursblue3}{RGB}{227,238,252}
\definecolor{oursblue4}{RGB}{218,232,252}  % same as oursblue
\definecolor{vitalblue}{HTML}{6C8EBF}
\definecolor{nonvitalgreen}{HTML}{82B366}
\definecolor{transitionpurple}{HTML}{9673A6}
\definecolor{addblue}{RGB}{30,80,160}\definecolor{addpill}{RGB}{220,235,255}%

\newcommand{\prog}{\mkern-4mu\blacktriangleright\mkern-4mu}

\title{ProgressNet: Sketching and Prompting\\with a Frozen Text-to-Image Model}

\author{
Arkaprabha Basu\textsuperscript{1,2}, \hspace{.2cm} Chaitat Utintu\textsuperscript{1}, \hspace{.2cm} Yi-Zhe Song\textsuperscript{1} \\
\textsuperscript{1} SketchX, CVSSP, University of Surrey \\
\textsuperscript{2} CoSTAR National Lab \\
{\tt\small \{a.basu, chaitat.u, y.song\}@surrey.ac.uk}\\
}

\iclrfinalcopy % Uncomment for camera-ready version, but NOT for submission.
\begin{document}

\maketitle

\begin{figure}[!ht]
    \vspace{-2em}
    \centering
    \includegraphics[width=\textwidth]{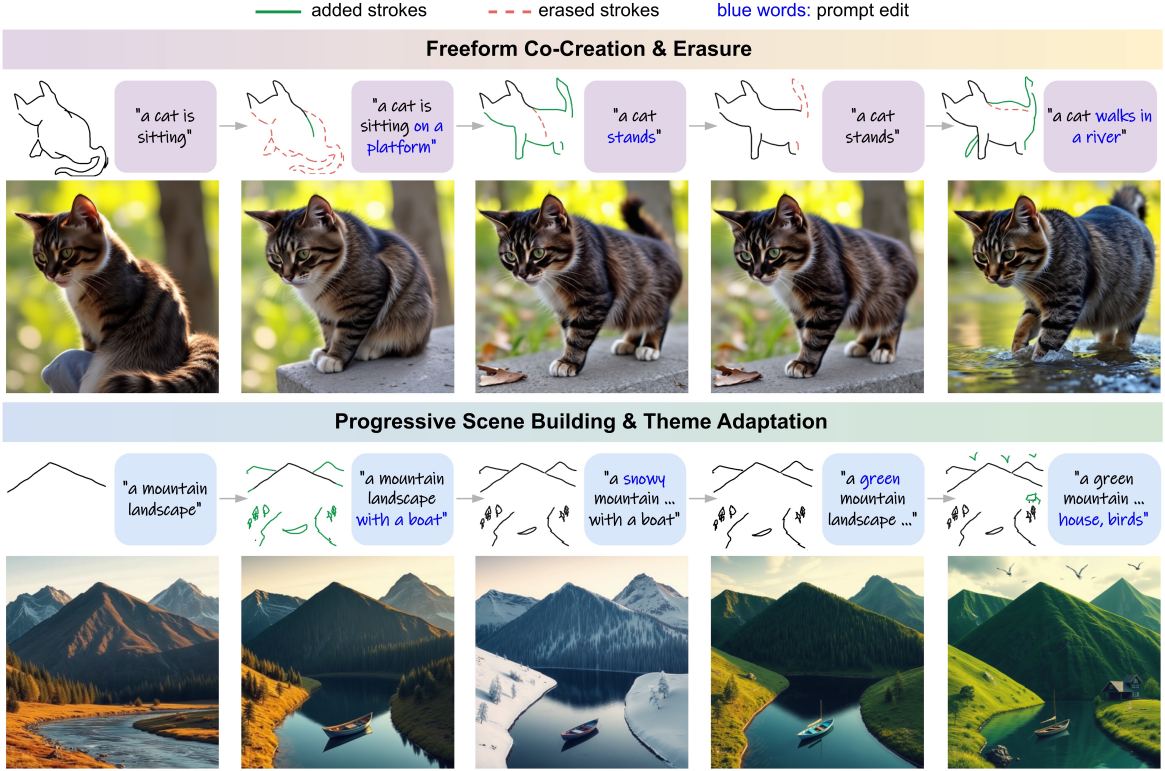}
    \vspace{-1.8em}
    % \caption{ProgressNet enables freeform, iterative sketch-to-image co-creation: users draw, erase, and refine prompts freely. \textbf{Top}: \emph{Freeform Co-Creation} through continuous stroke \textcolor{blue}{addition}, \textcolor{red}{removal} (\texttt{-}\texttt{-}), and prompt edits across turns. \textbf{Bottom}: \emph{Scene Building} with stroke \textcolor{blue}{insertion}, \textcolor{red}{removal} (\texttt{-}\texttt{-}), and theme reinterpretation via prompt. All training-free on frozen FLUX.}
    % \caption{ProgressNet supports freeform, iterative sketching to image co-creation. \textbf{Top}: a cat is progressively reimagined; from sitting to standing and walking in a river. \textbf{Bottom}: a mountain scene is evolved in composition and theme from a landscape to several environments. Both examples freely adopt stroke addition, removal, and prompt refinement across turns, all training-free on FLUX.}
    \caption{ProgressNet follows a drawing session turn by turn, training-free on frozen FLUX. \textbf{Top}: the cat’s body is erased and it moves onto a platform, stands once legs are drawn, loses its tail when it is erased, and walks into a river once the tail is redrawn. \textbf{Bottom}: a scene is built stroke by stroke, re-themed by the prompt from snowy to green, and finished with a house and birds.}
    % \textbf{Bottom}: \emph{Fine-Grained Control} from minimal, imperfect strokes. All training-free on frozen FLUX.}
    % \vspace{-0.5em}
    \label{fig:index}
\end{figure}

\begin{abstract}
\par\vspace{-0.5em}
Humans draw progressively: a few strokes, a look at the result, a stroke erased, a prompt revised. Image generators do not work this way. They typically take a finished sketch and produce the image in a single pass, so every edit starts the picture again, and the models that do keep state across turns are driven by text, cannot take a stroke, and are too slow to draw with. We present \emph{ProgressNet}, a training-free framework that lets a frozen text-to-image model follow a drawing session as it unfolds: strokes are added and erased, the prompt is revised, and the image keeps up at about a second per turn. It needs no new parameters because the frozen model already has what a progressive generator needs, a pathway through which the previous turn can be remembered, layers that can carry appearance forward without freezing structure, and an internal signal of how far to trust an unfinished sketch; three inference-time mechanisms (\emph{Previous-Concept Memory}, \emph{Layer-Selective K/V Injection} and \emph{Banded Adaptive Control}) use each in turn. As a sketch fills in, every existing method degrades, the FID of the FLUX+ControlNet baseline doubling between 10\% and 100\% completion on FS-COCO, while ProgressNet's barely moves; it maintains strong fidelity and progressive coherence across three sketch domains and is preferred by users over five competitors, most widely on erasure.
\end{abstract}

\vspace{-1em}
\section{Introduction}
\label{sec:intro}
\vspace{-0.5em}
% ── 1. Humans draw; generation has gained control, but takes the sketch as finished ──
Humans are visual creatures. We drew before we wrote, and drawing remains the most natural way to think through a visual idea and share it. Generative AI has opened a remarkable channel for this: a sentence now produces a picture~\citep{rombach2022ldm,flux2024}, in well under a second with few-step distillation~\citep{luo2023latent,sauer2024fast}, and because a sentence is a poor description of a picture, control has steadily grown beyond text, to edges, depth and layout~\citep{zhang2023controlnet,mou2024t2iadapter}, to reference images~\citep{ye2023ipadapter,bfl2025fluxkontext}, and most recently to a sketch drawn beside the prompt box~\citep{openai2026images25}. Sketch is the natural end of that road: it is how people convey shape and intent when words run out~\citep{eitz2012humans,koley2024itsallaboutyoursketch}. Yet every one of these systems takes the sketch as a finished input and produces the image in a single pass, and nobody draws that way. A drawing is made over time: a few strokes, a look, a stroke erased, a line redrawn, often a change of mind about what the picture is of (\cref{fig:index}).

% ── 2. What a person needs; ProgressNet ──
What a person needs is not a picture from a sketch but a picture that keeps up with the sketching: one that follows each added stroke, lets go of each erased one, absorbs a revised prompt, leaves everything else untouched, and answers fast enough to feel like a response rather than a resubmission. This paper is about that interaction. \emph{ProgressNet} lets a frozen text-to-image model support all of it, drawing, erasing and prompt revision, at about a second per turn and with no training.

% ── 3. Why existing systems do not, including editing models ──
\begin{wrapfigure}{r}{0.42\linewidth}
  \vspace{-1.2em}
  \centering
    \includegraphics[width=\linewidth]{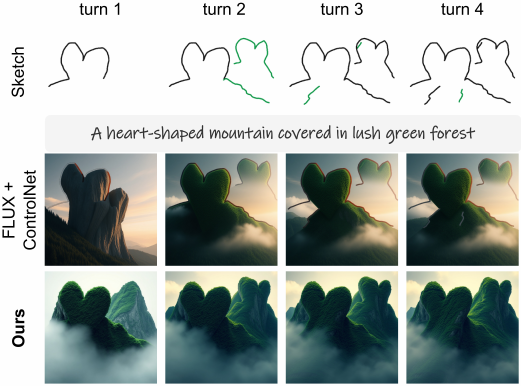}
  \vspace{-1.6em}
  \caption{Flux+ControlNet imprints raw stroke edges and visually drifts across progressive turns; ours adapts smoothly while maintaining temporal coherence.}
  \label{fig:introduction}
  \vspace{-1em}
\end{wrapfigure}
To the best of our knowledge, no existing system does all. Sketch-conditioned generation~\citep{zhang2023controlnet,koley2024itsallaboutyoursketch,li2024controlnetplusplus} is stateless: each turn regenerates the scene from nothing, so appearance drifts with every stroke and an erased stroke leaves a ghost, because nothing withdraws the evidence it added (\cref{fig:introduction}). Interactive sketch systems lower the cost of regeneration without removing its statelessness~\citep{sarukkai2024block,song2024sdxs}, and Subjective Camera~\citep{chen2025subjective} follows a drawing order by latent optimisation that takes minutes per scene. Image editing models~\citep{bfl2025fluxkontext,zhou2025multiturn,liao2025freqedit,xu2026anchoredit} do handle turn-by-turn interaction, but for a different kind of turn: the image enters as a second token stream in joint attention and the change is given in words, so there is no pathway for a stroke map and no way to say that a stroke is gone, and tens of denoising steps over a doubled sequence make each turn take seconds rather than the fraction of a second a drawing person will wait. ProgressNet therefore builds on the architecture that accepts strokes, a frozen flow-matching backbone with a ControlNet, and stays in the few-step regime that makes interaction possible.

% ── 4. Three problems, three intuitions ──
Following a drawing session with a frozen model raises three problems, and each has a simple answer inside the model. \textit{Remembering.} The model must remember what it generated last turn. The ControlNet's tile branch accepts an RGB image, so the previous output can be fed back as an explicit appearance memory, and because that memory lives in pixel space the user's operations map onto it directly: an added stroke keeps it, an erased stroke white-fills its region, a new prompt relaxes the regions the user has not committed to (\emph{Previous-Concept Memory}, PCM; \cref{sec:pcm}).
\textit{Carrying forward.} The model must carry that memory without freezing what should change. Training-free editors inject cached attention at the layers most vital to forming the image~\citep{avrahami2025stableflow}; under progression those are exactly the layers that must absorb new strokes, and locking them yields images that stop responding as the drawing grows. Injecting at the least vital layers instead keeps appearance while structure stays free (\emph{Layer-Selective K/V Injection}, LS-KVI; \cref{sec:lskvi}).
\textit{Trusting the sketch.} The model must balance adherence to an unfinished sketch with preservation of previous appearance. We regulate this balance indirectly, using changes in cached ControlNet memory features between the two preceding completed turns to modulate memory
strength across layer bands and solver steps, while structural control remains fixed (Banded Adaptive Control, BAC; \cref{sec:bac}). Together the three add $0.068$\,s to a two-step FLUX generation.

% ── 5. Payoff, PFC, contributions ──
The result is a generator whose quality does not decay as the drawing fills in: on FS-COCO~\citep{chowdhury2022fs} the FID of FLUX+ControlNet doubles between 10\% and 100\% completion and every baseline degrades the same way, while ProgressNet's moves by under one point. 
% Because frame-to-frame similarity alone rewards a method that ignores the new stroke, we also introduce a progressive fidelity--coherence score (PFC) that rewards staying coherent and tracking the intended image at once. 
Because frame-to-frame similarity alone rewards a method that ignores the new stroke, we introduce a progressive fidelity-coherence score (PFC) combining temporal coherence with completion-level distributional quality. Our contributions are
\textit{(i)} the first training-free system for progressive generation from sketch and text that supports stroke addition, erasure and prompt revision at interactive rates;
\textit{(ii)} PCM, LS-KVI and BAC, with the findings that vital-layer injection over-locks under progression and that inter-turn ControlNet feature redistribution is a usable online control signal, validated across two backbones and four sketch domains;
\textit{(iii)} a protocol and score for progressive generation, with results on three sketch domains and 
% a 42-participant 
user study in which ProgressNet is preferred over five competitors, most widely on erasure. None of the three mechanisms is sketch-specific.
\vspace{-0.5em}
\section{Related Work}
\label{sec:relatedworks}
\vspace{-0.5em}

Controllable image generation spans ControlNet-style residual conditioning~\citep{zhang2023controlnet}, multi-condition control~\citep{zhao2023unicontrolnet}, lightweight adapters~\citep{mou2024t2iadapter,ye2023ipadapter}, layout grounding~\citep{li2023gligen,bar2023multidiffusion}, and training-free feature steering~\citep{mo2024freecontrol,tumanyan2023plug}, recently ported to DiT and rectified-flow backbones~\citep{esser2024sd3,flux2024,tan2025ominicontrol}. Inversion-based editing~\citep{hertz2022prompttoprompt}, attention injection~\citep{avrahami2025stableflow}, and few-step distillation~\citep{song2024sdxs,sauer2024fast} improve editability and latency, bringing sketch interaction near real time~\citep{sarukkai2024block}. Others mitigate text-sketch conflict by re-weighting cross-attention~\citep{chefer2023attend,phung2023attentionrefocusing} or rescaling the control residual~\citep{liu2024smartcontrol,cohen2025conditional}. All operate single-shot: none tracks how sketch--text conflict evolves across turns, nor preserves prior content as the condition changes. ProgressNet targets this non-stationary setting via training-free temporal memory, selective feature reuse, and ControlNet-driven adaptive control, detailed in Appendix~\ref{sec:supp_related}.

\vspace{-0.5em}
\section{Preliminaries}
\label{sec:background}
\vspace{-0.5em}
%%% --------------ECCV --------------------------
 % ProgressNet builds on FLUX~\citep{flux2024}, a flow-matching model~\citep{lipman2022flow} with a DiT backbone~\citep{peebles2023scalable} of 19 joint transformer blocks and 38 single-stream blocks, conditioned via CLIP~\citep{radford2021learning} and T5~\citep{raffel2020exploring}. Each joint block applies \emph{joint} attention: latent and context tokens are concatenated into $X{=}[H;C]$ and projected through shared $W_{Q,K,V}$, making keys and values the natural locus for temporal repurposing (\cref{sec:lskvi}). Spatial control enters via a \textcolor{black}{lightweight} 5-layer ControlNet stack producing per-block residuals injected as $\mathbf{h}' \leftarrow \mathbf{h} + \lambda\,\mathbf{r}$. Under Union ControlNet, multiple \textcolor{black}{heterogeneous} conditioning signals (e.g.\ canny edges, tile memory) share weights with mode embeddings and their residuals are \textcolor{black}{simply} summed at injection. Since injection occurs at every \textcolor{black}{single} solver step, correct modulation of $\lambda$ per layer and per step is \textcolor{black}{particularly} essential under few-step inference (\cref{sec:bac}).
%%% --------------ECCV --------------------------
ProgressNet builds on flow-matching backbones~\citep{flux2024,bandyopadhyay2025sd3,lipman2022flow}
with Union ControlNet. Joint attention over latent and context tokens exposes the
K/V features later reused by LS-KVI (\cref{sec:lskvi}), while spatial conditions
such as Canny edges and tile memory are introduced through ControlNet residuals,
$\mathbf{h}'=\mathbf{h}+\lambda\mathbf{r}$. This formulation also makes the
control strength $\lambda$ a critical factor under few-step inference, where
over- or under-conditioning cannot be corrected over many denoising steps
(\cref{sec:bac}).

\begin{figure}[!t]
\centering
\begin{subfigure}{0.42\linewidth}
  \centering
  \includegraphics[width=\linewidth]{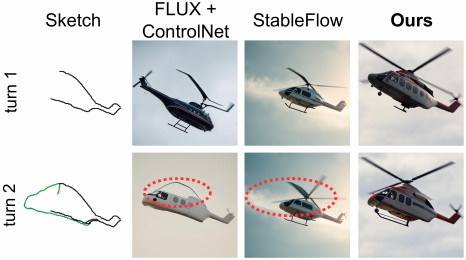}
  \vspace{-16pt}
% \caption{\textit{Baseline} and \textit{StableFlow} exhibit temporal instability and stroke artifacts on partial \textcolor{black}{strokes} (\textcolor{red}{red} borders) under progressive sketching. Ours consistently remains visually stable across turns.}
\caption{Progressive sketching.}
  \label{fig:pilotstudy_temporal}
\end{subfigure}
% \hfill
\hspace{15pt}
\begin{subfigure}{0.42\linewidth}
  \centering
  \includegraphics[width=\linewidth]{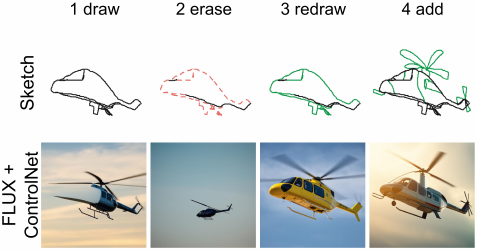}
  \vspace{-16pt}
  % \caption{Removal and re-addition of strokes (Sequence 1, 3). \textit{Baseline} fails to preserve previously established concepts, producing inconsistency that shift identity across editing turns.}
  \caption{Stroke removal and re-addition.}
  \label{fig:pilotstudy_removal}
\end{subfigure}
\vspace{-4pt}
\caption{Pilot studies illustrating failure modes of existing methods: temporal drift, prompt-sketch conflict under partial strokes, and contradiction upon stroke removal. \textbf{(a)} Under progressive sketching, \textit{FLUX+ControlNet} and \textit{StableFlow} show temporal instability and artifacts on the partial strokes (\textcolor{red}{red} borders), while ours remains stable across turns. \textbf{(b)} When strokes are removed and re-added (Seq.~1-3), \textit{FLUX+ControlNet} fails to preserve established concepts, shifting identity across turns.}
\label{fig:pilot_studies}
\vspace{-1em}
\end{figure}

Our pilot studies (\cref{fig:pilotstudy_temporal,fig:pilotstudy_removal}) show
that existing formulations struggle to maintain a coherent progressive state:
appearance can drift between turns, a fixed control strength can either imprint
partial strokes too strongly or under-enforce them, and erased content may persist
when the sketch is revised. ProgressNet addresses these coupled failures through
explicit concept memory, selective temporal feature reuse, and adaptive control
modulation.

\section{Proposed Methodology}
\label{sec:methodology}
\vspace{-0.25em}
\textbf{Overview.}
Given a sketch $\mathbf{S}_t$, prompt $P_t$, and previous output $I_{t-1}$, we generate
$I_t=\mathcal{G}(\mathbf{S}_t,P_t,I_{t-1};\theta)$ while keeping pretrained weights
$\theta$ frozen and supporting stroke addition, removal, and prompt updates.
ProgressNet combines three mechanisms: (i) \emph{PCM} uses
operation-aware tile conditioning to preserve prior appearance (\cref{sec:pcm});
(ii) \emph{LS-KVI} reuses previous-turn features only at
selected layers to maintain context while permitting structural change
(\cref{sec:lskvi}); and 
%%%%%%%%%% --------------FINAL
% (iii) \emph{BAC} replaces fixed
% ControlNet scaling with step and band specific 
% % modulation 
% adaptation driven by inter-turn control drift (\cref{sec:bac}).
%%%%%%%%%% --------------FINAL
(iii) BAC applies retrospective, step- and band-specific modulation to the
memory-conditioning ControlNet stream using cached inter-turn comparisons.
\vspace{-0.2cm}
\subsection{Previous-Concept Memory (PCM)}
\label{sec:pcm}
\vspace{-0.1cm}

% Progressive sketching is inherently underdetermined: the current strokes constrain
% geometry, but provide little evidence for appearance, illumination, or scene context.
% Consequently, independently regenerating each partial sketch can substantially alter
% already established content~\citep{koley2024itsallaboutyoursketch,
% liu2025sketchvideo,luo2023latent}. Existing sketch-control methods
% ~\citep{zhang2023controlnet,zhao2023unicontrolnet} condition only on the current
% structure and therefore carry no explicit notion of what the image looked like at the
% previous interaction turn.
Progressive sketching is underdetermined: partial strokes constrain geometry but say little about appearance or context, so regenerating each turn independently can alter established content~\citep{koley2024itsallaboutyoursketch,liu2025sketchvideo,luo2023latent}. Existing sketch-control methods~\citep{zhang2023controlnet,zhao2023unicontrolnet} condition only on current strokes, with no memory of previous turns.

\begin{figure}[!t]
    \centering
    % \vspace{-0.5em}
    % \includegraphics[width=\textwidth]{images/pcm_v6_cr_iclr.pdf}
    % \includegraphics[width=\textwidth]{images/methodology_iclr_v2.pdf}
    \includegraphics[width=\textwidth]{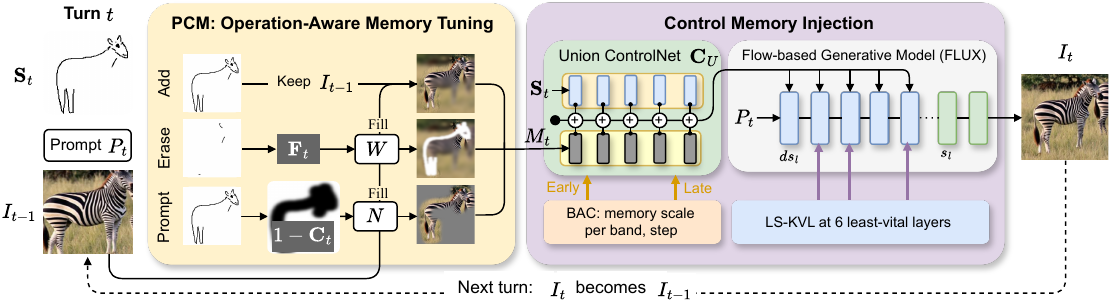}
    \vspace{-1.5em}
% \caption{Overview of ProgressNet. \textbf{Left}: the previous output $I_{t-1}$ is prepared according to the current operation: addition keeps $I_{t-1}$, removal white-fills ($W$) the erased region $\mathbf{F}_t$, and a prompt update gray-fills ($N$) the uncommitted region $1-\mathbf{C}_t$. \textbf{Middle}: the memory $M_t$ and current sketch $\mathbf{S}_t$ condition Union ControlNet $\mathbf{C}_U$, while BAC scales the memory stream per layer band and solver step. \textbf{Right}: LS-KVI reuses previous-turn K/V features at six low-vitality FLUX layers to preserve temporal context while the current structure adapts.}
\caption{Overview of ProgressNet. \textbf{Left}: the previous output $I_{t-1}$ is prepared according to the current operation: addition keeps $I_{t-1}$, removal white-fills ($W$) the erased region $\mathbf{F}_t$, and a prompt update gray-fills ($N$) the uncommitted region $1-\mathbf{C}_t$. \textbf{Middle}: the memory $M_t$ and current sketch $\mathbf{S}_t$ condition Union ControlNet $\mathbf{C}_U$, while BAC scales the memory stream per layer band and solver step. \textbf{Right}: LS-KVI reuses previous-turn K/V features at six low-vitality FLUX layers.}
    \label{fig:pcm}
    \vspace{-6pt}
\end{figure}
\textbf{Memory-induced correction.}
We retain the previous generation $I_{t-1}$ as appearance memory, adapted per interaction: adding strokes preserves appearance, removing strokes withdraws obsolete evidence, and prompt updates retain only committed content. Let $\mathbf{F}_t$ be the erased region, $\mathbf{C}_t$ the soft commitment mask, and $W$, $N$ white and gray images. The memory condition is:
% \vspace{-2pt}
\begin{equation}
M_t:\quad
s_A \mapsto I_{t-1},\qquad
s_R \mapsto (1-\mathbf{F}_t)\odot I_{t-1}+\mathbf{F}_t\odot W,\qquad
p_U \mapsto \mathbf{C}_t\odot I_{t-1}+(1-\mathbf{C}_t)\odot N .
\label{eq:pcm_memory}
% \vspace{-0.25em}
\end{equation}
Where, $s_A$, $s_R$, and $p_U$ denote stroke addition, removal, and
prompt update, respectively; $\mathbf{C}_t$ marks committed content derived
from the current processed sketch (\cref{sec:supp_pcm_impl}). Neutral filling
withdraws obsolete evidence, while $\mathbf{C}_t$ permits uncommitted regions
to adapt to the updated prompt.

% \vspace{-0.4em}
\textbf{Memory-structure fusion.}
The prepared RGB memory $M_t$ is passed through the tile branch of 
Union ControlNet ($\mathbf{C}_U$), while the current sketch $\mathbf{S}_t$ supplies the binary structural sketch condition. Their residuals are fused at each controlled layer:
% \vspace{-0.5em}
\begin{equation}
\mathbf{h}'_{\ell}
=
\mathbf{h}_{\ell}
+
\lambda_s\,\mathbf{r}^{\ell}_{\mathrm{str}}(\mathbf{S}_t)
+
\lambda_m\,\mathbf{r}^{\ell}_{\mathrm{mem}}(M_t),
\label{eq:pcm_fused_residual}
% \vspace{-0.6em}
\end{equation}
where $\lambda_s$ and $\lambda_m$ weight structural and appearance evidence, respectively. The current sketch thus determines \emph{what should change}, while the PCM determines \emph{what should remain}. Since the memory is an RGB image, it can be edited directly according to the user operation before conditioning, with no extra learned module or additional learned preprocessing. PCM thus turns a stateless sketch-conditioned generator into an operation-aware progressive one, entirely at inference time.
\vspace{-0.1cm}
\subsection{Layer-Selective K/V Injection (LS-KVI)}
\label{sec:lskvi}
\vspace{-0.1cm}
PCM anchors successive generations via memory conditioning, but internal representations can still drift as attention is recomputed at each step. Training-free editors reduce such drift by reusing attention features, either injecting source features into key DiT layers~\citep{avrahami2025stableflow} or reusing layer-specific K/V features~\citep{wei2025freeflux,dalva2025fluxspace,kim2025reflex}. However, they target prompt-only editing, not the control-driven edits central to progressive drawing.

\begin{wrapfigure}{r}{0.4\linewidth}
    \hspace*{-4pt}
    % \par\vspace{-1em}
    \centering
    \includegraphics[width=\linewidth]{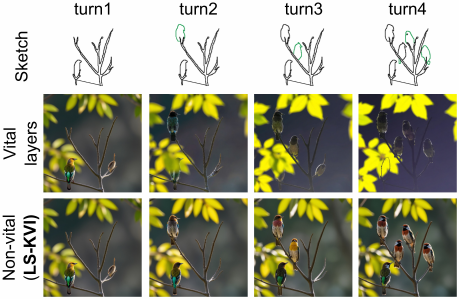}
    \vspace{-1.5em}
    % \caption{Naively transferring vital-layer feature injection from text-guided
    % editing to progressive ControlNet generation over-preserves the initial
    % composition as new strokes arrive. LS-KVI instead propagates previous-turn
    % information through low-vitality layers, retaining scene context while allowing
    % structure and semantics to evolve.}
% \caption{Vital-layer injection increasingly locks sharper initial information while losing adaptability. Non-vital LS-KVI preserves appearance and sharpness while allowing the structure to evolve with the sketch.}
\caption{Vital vs.\ non-vital layer. Vital-layer injection locks in early details, losing adaptability, while non-vital LS-KVI preserves appearance and sharpness as structure follows sketch.}
    \label{fig:lskvi_exp}
    \par\vspace{-0.5em}
\end{wrapfigure}

\textbf{From image editing to progressive control.}
Progressive generation differs from text-guided editing. In editing, feature injection preserves a source image while text updates selected semantics. In progressive control, each turn adds a new spatial signal via ControlNet, and new strokes may change geometry, pose, or even the concept. Carrying strong previous-turn features through influential layers can thus conflict with the current structure and over-preserve content that should evolve, as we observe when adapting vital-layer injection to progressive sketching (\cref{fig:lskvi_exp}). This suggests a different principle: progressive generation should not freeze the layers most responsible for forming the current concept. We therefore leave vital layers without direct temporal K/V injection and reuse features at selected lower-vitality layers, assessing the resulting preservation and adaptation empirically.

% \vspace{-0.4em}
\textbf{Progressive layer selection.}
Following the residual-bypass analysis used to study layer importance
~\citep{avrahami2025stableflow}, we measure layer vitality directly under progressive
controlled generation rather than prompt-only editing. For each transformer layer
$\ell$, we bypass that layer through its residual connection and measure the output
change across progressive sketch stages. Let $\hat I^{\mathrm{ref}}_{x,k}$ denote the reference generation for sketch $x$ at progression stage $k$, and let $\hat I^{(-\ell)}_{x,k}$ denote the corresponding generation with layer $\ell$ bypassed. We define:
\vspace{-0.25em}
\begin{equation}
\mathcal{V}(\ell)
=
\frac{1}{|\mathcal{D}||\mathcal{S}|}
\sum_{x\in\mathcal{D}}
\sum_{k\in\mathcal{S}}
\left\|
\hat I^{\mathrm{ref}}_{x,k}
-
\hat I^{(-\ell)}_{x,k}
\right\|_2^2 .
\label{eq:vitality}
\end{equation}
% \par\vspace{-0.25em}
% \par\vspace{-1em}
% where, $\mathcal{D}$ denotes progressive sequences and $\mathcal{S}$ their
% completion stages. High $\mathcal{I}(\ell)$ identifies layers that strongly influence the evolving
% concept and are therefore excluded from temporal K/V injection. LS-KVI selects $\mathcal{L}_{\mathrm{inj}}$ from low-vitality layers, propagating temporal context without constraining adaptation.
where $\mathcal{D}$ denotes the set of progressive sequences and
$\mathcal{S}$ the evaluated completion stages. We use
$\mathcal{V}(\ell)$ as a layer-selection criterion: lower scores
identify layers whose bypass causes smaller output deviations in the evaluated
progressive setting.
We select $L_{\mathrm{inj}}$ from these layers and assess the
injection choice separately through the vital vs. non-vital
comparisons in \cref{sec:supp_components,fig:supp_lskvi_qual}.

% \vspace{-0.3em}
\textbf{Selective temporal injection.}
For each turn $t>0$, we cache the preceding K/V projections and inject them only
for $\ell\in\mathcal{L}_{\mathrm{inj}}$ as
$\mathbf{K}^{t\prime}_{\ell}=\mathbf{K}^{t}_{\ell}
+\alpha_{\mathrm{kv}}\mathbf{K}^{t-1}_{\ell}$ and
$\mathbf{V}^{t\prime}_{\ell}=\mathbf{V}^{t}_{\ell}
+\alpha_{\mathrm{kv}}\mathbf{V}^{t-1}_{\ell}$, while layers outside $L_{\mathrm{inj}}$ receive no direct
temporal K/V injection.
% LS-KVI therefore preserves temporal context without rigidly copying the
% previous concept: low-vitality layers carry what should remain stable, while
% high-vitality layers remain available for progressive structural and semantic
% adaptation. 
% The cache is refreshed after every turn and cleared upon stroke removal
% to prevent erased content from being reintroduced through stale correspondence
% pathways~\citep{qi2023fatezero,geyer2023tokenflow}.
% The cache is refreshed after every turn and cleared upon stroke removal
% to prevent erased content from being reintroduced through stale correspondence
% pathways~\citep{qi2023fatezero,geyer2023tokenflow}.
The cache is refreshed after every turn and cleared upon stroke removal;
this prevents stale K/V correspondences from resisting erasure, as quantified
in \cref{tab:supp_erasure}.

% PCM and LS-KVI consequently operate at complementary levels: 
Putting it all together, PCM carries the previous concept through external conditioning, whereas LS-KVI reuses internal attention features at selected
layers. Neither, however, determines how strongly the preceding memory should constrain generation with progressive drawing. We address this
through adaptive memory control in \cref{sec:bac}.
\subsection{Banded Adaptive Control (BAC)}
\label{sec:bac}
% \vspace{-0.5em}
% PCM and LS-KVI preserve information across turns, but neither determines how
% strongly the current spatial condition should influence generation. This becomes
% critical during sketching: a partial stroke can be ambiguous or conflict
% with the prompt, while the same structure may become reliable only a few turns
% later. Existing adaptive-control methods learn a mixing policy or selectively modify
% conditioning locations~\citep{liu2024smartcontrol,cohen2025conditional}; instead,
% we ask whether the pretrained model itself exposes how strongly its spatial-control response changes under progressive updates.
PCM and LS-KVI preserve information across turns, but PCM retains a fixed
memory-control strength. During progressive sketching, however, how strongly
the preceding appearance should constrain the current generation changes as
new structural evidence arrives. Rather than learning a mixing policy or modifying conditioning flow~\citep{liu2024smartcontrol,cohen2025conditional}, we ask whether the pretrained model itself reveals how its control response changes under progressive updates.

\begin{figure*}[!t]
\vspace{-0.7em}
  \centering
  \includegraphics[width=\linewidth]{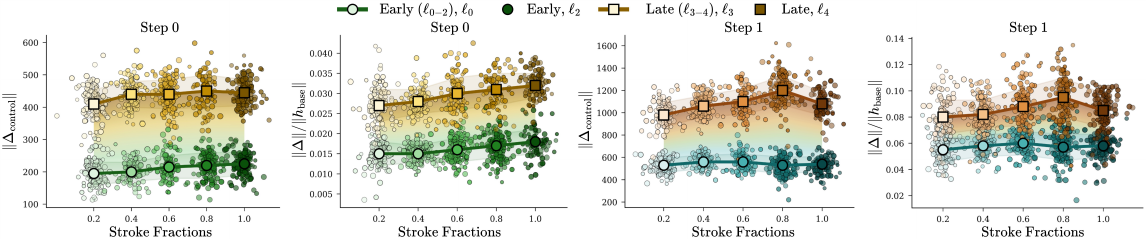}
  \vspace{-18pt}
  \caption{Layer- and solver-step variation in control
responses. Control-induced perturbations and their normalized
magnitudes vary across the evaluated layer groups and solver
steps during progression. Appendix \cref{sec:supp_sd35,sec:supp_crossdataset} report complementary
feature-redistribution diagnostics on SD3.5 and across four
sketch domains.}
  \label{fig:bac_control_dist}
\vspace{-1em}
\end{figure*}

\textbf{Control is band- and step-asymmetric.}
We first study how ControlNet changes the backbone during progressive generation.
Let $\mathbf{h}^{\mathrm{ctrl}}_{\ell}(t,s)$ and
$\mathbf{h}^{\mathrm{base}}_{\ell}(t,s)$ denote hidden states at transformer
layer $\ell$, progressive stage $t$, and solver step $s$, with and without spatial
control. We measure the induced perturbation as
$\Delta_{\ell}^{(t,s)}
=
\mathbf{h}^{\mathrm{ctrl}}_{\ell}(t,s)
-
\mathbf{h}^{\mathrm{base}}_{\ell}(t,s)$.
The perturbation reflects the backbone response across depth
(\cref{fig:supp_perlayer}) and varies across layer groups and solver steps
(\cref{fig:bac_control_dist}). Complementary diagnostics show related asymmetries
on SD3.5 and four sketch domains
(\cref{sec:supp_sd35,sec:supp_crossdataset}). These observations motivate
band- and step-specific memory modulation rather than a single global factor.

\vspace{-0.4em}
\textbf{Retrospective ControlNet feature comparison.}
BAC must determine its factor before sampling. A current-turn band statistic is
available only after that band is traversed, requiring replay to apply it
consistently. We therefore use the latest complete transition,
$t\!-\!2\rightarrow t\!-\!1$, as a causal proxy for recent memory redistribution.
Let $\bar{\mathbf{c}}^{k}_{u,s,\mathrm{mem}}$ denote the raw memory-conditioning
feature cached after turn $u$ from ControlNet block $k$ at step $s$.
For each block $k$ and step $s$, we compute channel-wise JSD,
\begin{equation}
J^{k}_{t,s}
=
\operatorname{JSD}_{\mathrm{cw}}\!\left(
\bar{\mathbf{c}}^{k}_{t-2,s,\mathrm{mem}},
\bar{\mathbf{c}}^{k}_{t-1,s,\mathrm{mem}}
\right),
\qquad
f^{b}_{t,s}
=
\operatorname{clip}\!\left(
\frac{1}{|\mathcal{K}_b|}
\sum_{k\in\mathcal{K}_b}J^{k}_{t,s},
f_{\min},f_{\max}
\right).
\label{eq:bac_cached}
\end{equation}
Here, $\operatorname{JSD}_{\mathrm{cw}}$ averages divergence across feature channels, $\mathcal{K}_b$ denotes blocks in $b \in \{\mathrm{early}, \mathrm{late}\}$ and $f_{\min},f_{\max}$ bound modulation. Larger recent redistribution indicates a less stable appearance state, retaining stronger memory conditioning to counter drift, while stable representations permit attenuation. BAC modulates only the current memory-conditioning stream, $\tilde{\mathbf{c}}^{k}_{t,s,\mathrm{mem}} = f^{b(k)}_{t,s}\mathbf{c}^{k}_{t,s,\mathrm{mem}}$, where $b(k)$ maps block $k$ to its band. Accordingly, \cref{eq:pcm_fused_residual} becomes
\begin{equation}
\mathbf{h}'_{\ell,s}
=
\mathbf{h}_{\ell,s}
+
\lambda_s\,\mathbf{r}^{\ell}_{\mathrm{str}}(\mathbf{S}_t)
+
\lambda_m\,\tilde{\mathbf{r}}^{\ell}_{\mathrm{mem},s}(M_t);
\qquad
\{\tilde{\mathbf c}^{k}_{t,s,\mathrm{mem}}\}_{k}
\longmapsto
\tilde{\mathbf r}^{\ell}_{\mathrm{mem},s}.
\label{eq:bac_factor}
\end{equation}
BAC produces four factors under two-step inference:
${f^{\mathrm{early}}_{t,0},f^{\mathrm{late}}_{t,0},
f^{\mathrm{early}}_{t,1},f^{\mathrm{late}}_{t,1}}$.

\vspace{-0.2em}
\section{Experiments}
\label{sec:experiments}
\vspace{-0.3em}
\textbf{Datasets.}
% We evaluate on three complementary sketch domains: FS-COCO~\citep{chowdhury2022fs}
% ($N{=}10{,}000$), comprising temporally ordered freehand scene sketches paired with
% images and captions; Photo-Sketching ($N{=}500$), containing edge-aligned contour
% sketches; and Sketchy ($N{=}1{,}250$), stratified across 125 categories of
% single-object freehand sketches. Together, they cover scene/object-level and
% freehand/contour-based sketch representations.
We evaluate on three sketch domains:
FS-COCO~\citep{chowdhury2022fs}
(freehand scene sketches),
Photo-Sketching~\citep{li2019photo}
(edge-aligned contour sketches), and
Sketchy~\citep{sangkloy2016sketchy}
(single-object freehand sketches).
Together, they span scene- and object-level generation as well as freehand
and contour-based representations.

\vspace{-0.15em}
\textbf{Implementation Details.}
We use FLUX.1-schnell~\citep{flux2024} with Union ControlNet at
$1024{\times}1024$ resolution and two denoising steps. We set
$\alpha_{kv}{=}0.3$, $f_{\min}{=}0.05$, and $f_{\max}{=}1.0$.
Experiments are conducted on a single NVIDIA H200 GPU.
% \vspace{-3pt}
\begin{table*}[!ht]
\centering
\caption{Comparison with SOTA methods across sketch completion levels and domains. Latency is measured per turn, independent of completion level. \textbf{Bold}: best; \ul{underline}: second best.}
\label{tab:quantitative}
\vspace{-6pt}
\definecolor{addblack}{RGB}{30,80,160}
\definecolor{oursblue}{RGB}{230,245,255}
\setlength{\tabcolsep}{1.5pt}
\renewcommand{\arraystretch}{1}

\resizebox{\textwidth}{!}{%
\scriptsize
\begin{tabular}{l cccccc cccccc cccccc}
\toprule
& \multicolumn{6}{c}{\textbf{10\% Completion}}
& \multicolumn{6}{c}{\textbf{50\% Completion}}
& \multicolumn{6}{c}{\textbf{100\% Completion}} \\
\cmidrule(lr){2-7}
\cmidrule(lr){8-13}
\cmidrule(lr){14-19}

Metric
& \rotatebox{90}{\scriptsize S-a-S}
& \rotatebox{90}{\scriptsize SDXS}
& \rotatebox{90}{\scriptsize CB}
& \rotatebox{90}{\scriptsize Flux}
& \rotatebox{90}{\scriptsize SF}
& \cellcolor{oursblue}\rotatebox{90}{\scriptsize \textbf{Ours}}
& \rotatebox{90}{\scriptsize S-a-S}
& \rotatebox{90}{\scriptsize SDXS}
& \rotatebox{90}{\scriptsize CB}
& \rotatebox{90}{\scriptsize Flux}
& \rotatebox{90}{\scriptsize SF}
& \cellcolor{oursblue}\rotatebox{90}{\scriptsize \textbf{Ours}}
& \rotatebox{90}{\scriptsize S-a-S}
& \rotatebox{90}{\scriptsize SDXS}
& \rotatebox{90}{\scriptsize CB}
& \rotatebox{90}{\scriptsize Flux}
& \rotatebox{90}{\scriptsize SF}
& \cellcolor{oursblue}\rotatebox{90}{\scriptsize \textbf{Ours}} \\
\midrule

% ============================================================
% FS-COCO
% ============================================================
\multicolumn{19}{l}{\textbf{FS-COCO}} \\[-2pt]

FID-I $\downarrow$
& 192.3 & 40.3 & 195.9 & \textbf{32.9}
& 36.8 & \cellcolor{oursblue}\ul{36.5}
& 200.1 & 41.4 & 209.9 & 44.7
& \ul{40.7} & \cellcolor{oursblue}\textbf{36.0}
& 219.2 & \ul{49.6} & 208.3 & 67.9
& 52.8 & \cellcolor{oursblue}\textbf{37.3} \\

FID-C $\downarrow$
& 0.432 & 0.161 & 0.400 & \textbf{0.149}
& 0.185 & \cellcolor{oursblue}\ul{0.152}
& 0.434 & \ul{0.170} & 0.396 & 0.188
& 0.201 & \cellcolor{oursblue}\textbf{0.141}
& 0.475 & \ul{0.188} & 0.407 & 0.264
& 0.217 & \cellcolor{oursblue}\textbf{0.143} \\

DINOv2 $\uparrow$
& 0.286 & 0.331 & 0.328 & 0.375
& \ul{0.397} & \cellcolor{oursblue}\textbf{0.400}
& 0.305 & 0.337 & 0.340 & \ul{0.386}
& 0.384 & \cellcolor{oursblue}\textbf{0.415}
& 0.276 & 0.328 & 0.317 & \ul{0.361}
& 0.359 & \cellcolor{oursblue}\textbf{0.413} \\

CLIPScore $\uparrow$
& 0.267 & 0.299 & 0.293 & 0.298
& \ul{0.301} & \cellcolor{oursblue}\textbf{0.306}
& 0.268 & \ul{0.297} & 0.291 & 0.285
& 0.296 & \cellcolor{oursblue}\textbf{0.305}
& 0.256 & \ul{0.291} & 0.289 & 0.265
& 0.289 & \cellcolor{oursblue}\textbf{0.303} \\

T5-Cos $\uparrow$
& 0.637 & 0.649 & 0.645 & 0.651
& \textbf{0.659} & \cellcolor{oursblue}\ul{0.657}
& 0.631 & \ul{0.647} & 0.637 & 0.642
& \textbf{0.656} & \cellcolor{oursblue}\textbf{0.656}
& 0.630 & 0.645 & 0.640 & 0.641
& \ul{0.654} & \cellcolor{oursblue}\textbf{0.655} \\

\addlinespace[2pt]
\midrule

% ============================================================
% PHOTO-SKETCHING
% ============================================================
\multicolumn{19}{l}{\textbf{Photo-Sketching}} \\[-2pt]

FID-I $\downarrow$
& 183.9 & 194.9 & 173.4 & \textbf{66.0}
& 81.0 & \cellcolor{oursblue}\ul{76.2}
& 188.5 & 188.7 & 176.7 & \textbf{70.6}
& \textbf{70.6} & \cellcolor{oursblue}\ul{73.3}
& 193.4 & 195.0 & 175.5 & 77.8
& \ul{77.5} & \cellcolor{oursblue}\textbf{72.5} \\

FID-C $\downarrow$
& 0.358 & 0.294 & 0.335 & \textbf{0.116}
& 0.171 & \cellcolor{oursblue}\ul{0.148}
& 0.364 & 0.287 & 0.333 & \textbf{0.121}
& 0.156 & \cellcolor{oursblue}\ul{0.137}
& 0.374 & 0.303 & 0.334 & \ul{0.136}
& 0.139 & \cellcolor{oursblue}\textbf{0.135} \\

DINOv2 $\uparrow$
& 0.259 & 0.259 & 0.341 & 0.325
& \textbf{0.381} & \cellcolor{oursblue}\textbf{0.381}
& 0.275 & 0.283 & 0.326 & \ul{0.355}
& \ul{0.355} & \cellcolor{oursblue}\textbf{0.382}
& 0.274 & 0.268 & 0.326 & \ul{0.381}
& 0.354 & \cellcolor{oursblue}\textbf{0.382} \\

CLIPScore $\uparrow$
& 0.285 & 0.310 & 0.321 & 0.319
& \ul{0.325} & \cellcolor{oursblue}\textbf{0.334}
& 0.281 & 0.311 & 0.317 & 0.310
& \ul{0.323} & \cellcolor{oursblue}\textbf{0.332}
& 0.278 & 0.308 & 0.315 & 0.309
& \ul{0.323} & \cellcolor{oursblue}\textbf{0.331} \\

T5-Cos $\uparrow$
& 0.721 & 0.725 & 0.777 & 0.760
& \textbf{0.781} & \cellcolor{oursblue}\ul{0.780}
& 0.714 & 0.726 & 0.770 & 0.760
& \ul{0.777} & \cellcolor{oursblue}\textbf{0.784}
& 0.710 & 0.728 & \ul{0.767} & 0.766
& \textbf{0.783} & \cellcolor{oursblue}\textbf{0.783} \\

\addlinespace[2pt]
\midrule

% ============================================================
% SKETCHY
% ============================================================
\multicolumn{19}{l}{
\textbf{Sketchy}
} \\[-1.3pt]

FID-I $\downarrow$
& 123.8 & 119.7 & 117.1 & 104.7
& \textbf{100.5} & \cellcolor{oursblue}\ul{101.6}
& 125.3 & 121.1 & 116.4 & \ul{111.8}
& 115.2 & \cellcolor{oursblue}\textbf{100.2}
& 124.1 & 122.0 & 116.4 & \ul{115.5}
& 137.1 & \cellcolor{oursblue}\textbf{99.6} \\

FID-C $\downarrow$
& \textbf{0.264} & 0.282 & 0.378 & \ul{0.267}
& 0.299 & \cellcolor{oursblue}0.292
& 0.281 & 0.280 & 0.378 & \ul{0.277}
& 0.308 & \cellcolor{oursblue}\textbf{0.270}
& \ul{0.281} & \ul{0.281} & 0.382 & 0.292
& 0.334 & \cellcolor{oursblue}\textbf{0.271} \\

DINOv2 $\uparrow$
& 0.339 & 0.360 & 0.425 & 0.446
& \ul{0.474} & \cellcolor{oursblue}\textbf{0.479}
& 0.345 & 0.361 & 0.433 & \ul{0.447}
& 0.430 & \cellcolor{oursblue}\textbf{0.487}
& 0.358 & 0.365 & 0.434 & \ul{0.451}
& 0.395 & \cellcolor{oursblue}\textbf{0.496} \\

CLIPScore $\uparrow$
& 0.269 & 0.291 & \textbf{0.307} & 0.297
& 0.301 & \cellcolor{oursblue}\ul{0.302}
& 0.268 & 0.289 & \textbf{0.306} & 0.293
& 0.291 & \cellcolor{oursblue}\ul{0.300}
& 0.268 & 0.289 & \ul{0.299} & 0.290
& 0.281 & \cellcolor{oursblue}\textbf{0.316} \\

T5-Cos $\uparrow$
& 0.526 & 0.520 & \ul{0.577} & 0.522
& 0.524 & \cellcolor{oursblue}\textbf{0.627}
& 0.523 & 0.520 & \ul{0.529} & 0.517
& 0.525 & \cellcolor{oursblue}\textbf{0.626}
& \ul{0.526} & 0.519 & 0.523 & 0.517
& 0.524 & \cellcolor{oursblue}\textbf{0.623} \\

\midrule

% ============================================================
% LATENCY
% ============================================================
Latency (s) $\downarrow$
& 2.525
& \textbf{0.049}
& 9.483
& \ul{0.761}
& 0.762
& \cellcolor{oursblue}0.829
& \multicolumn{12}{c}{
    \textit{completion-invariant}
    \qquad
    {\scriptsize\textcolor{addblack}{\textbf{Ours: +0.068\,s vs.\ Flux}}}
} \\

\bottomrule
\end{tabular}
}
\vspace{-6pt}

\end{table*}
\begin{table*}[!ht]
\centering
\caption{
Progressive stability across sketch domains.
PFC-DINO and PFC-LPIPS evaluate temporal coherence and generation quality at the
three reported FID-I anchors. Cross-Comp. reports cross-completion robustness,
normalized to the Flux baseline ($100$).
\textbf{Bold}: best; \ul{underline}: second best.
}
\label{tab:stability}
\vspace{-6pt}

\definecolor{oursblue}{RGB}{230,245,255}
\setlength{\tabcolsep}{2.5pt}
\renewcommand{\arraystretch}{0.92}

\resizebox{\textwidth}{!}{%
\small
\begin{tabular}{ll|cccc|cccc|cccc}
\toprule
& &
\multicolumn{4}{c|}{\textsc{PFC-DINO} $\uparrow$}
&
\multicolumn{4}{c|}{\textsc{PFC-LPIPS} $\uparrow$}
&
\multicolumn{4}{c}{\textsc{Cross-Comp. Robustness} $\uparrow$}
\\
\cmidrule(lr){3-6}
\cmidrule(lr){7-10}
\cmidrule(lr){11-14}

Dataset & Method
& $10\prog20$
& $30\prog50$
& $80\prog100$
& Mean
& $10\prog20$
& $30\prog50$
& $80\prog100$
& Mean
& Dist.
& Vis.
& Text
& Overall
\\
\midrule

% ============================================================
% FS-COCO
% ============================================================
\multirow{6}{*}{\textbf{FS-COCO}}
& Sketch-a-Sketch
& 0.237 & 0.228 & 0.222 & 0.229
& 0.331 & 0.322 & 0.299 & 0.317
& 39.51 & 76.22 & 97.50 & 66.47 \\

& SDXS
& 1.883 & 1.824 & 1.579 & 1.762
& 1.864 & 1.820 & 1.546 & 1.744
& \ul{133.14} & 90.18 & 105.10 & 108.06 \\

& Cond.\ Balance
& 0.261 & 0.236 & 0.229 & 0.242
& 0.318 & 0.298 & 0.299 & 0.305
& 42.74 & 87.68 & 104.06 & 73.06 \\

& Flux Baseline
& 1.836 & 1.387 & 0.999 & 1.407
& 1.896 & 1.425 & 0.982 & 1.435
& 100.00 & 100.00 & 100.00 & 100.00 \\

& StableFlow
& \ul{2.198} & \ul{2.120} & \ul{1.682} & \ul{2.000}
& \ul{2.174} & \ul{2.151} & \ul{1.726} & \ul{2.017}
& 119.70 & \ul{100.23} & \ul{105.41} & \ul{108.14} \\

\rowcolor{oursblue}
& \textbf{Ours}
& \textbf{2.244} & \textbf{2.333} & \textbf{2.343} & \textbf{2.307}
& \textbf{2.190} & \textbf{2.257} & \textbf{2.223} & \textbf{2.223}
& \textbf{167.49} & \textbf{110.64} & \textbf{107.99} & \textbf{126.02}
\\

\midrule

% ============================================================
% PHOTO-SKETCHING
% ============================================================
\multirow{6}{*}{\textbf{Photo-Sketch}}
& Sketch-a-Sketch
& 0.202 & 0.202 & 0.207 & 0.204
& 0.343 & 0.337 & 0.334 & 0.338
& 37.63 & 78.94 & 92.07 & 64.91 \\

& SDXS
& 0.325 & 0.345 & 0.367 & 0.345
& 0.384 & 0.403 & 0.403 & 0.397
& 41.45 & 78.41 & 97.88 & 68.26 \\

& Cond.\ Balance
& 0.430 & 0.406 & 0.440 & 0.425
& 0.501 & 0.490 & 0.512 & 0.501
& 41.39 & 97.56 & 101.79 & 74.35 \\

& Flux Baseline
& 0.735 & 0.732 & 0.749 & 0.739
& 0.932 & 0.887 & 0.838 & 0.886
& \textbf{100.00} & 100.00 & 100.00 & \ul{100.00} \\

& StableFlow
& \ul{1.001} & \textbf{1.229} & \ul{1.163} & \textbf{1.131}
& \ul{1.029} & \textbf{1.285} & \ul{1.201} & \textbf{1.172}
& 86.26 & \ul{105.74} & \ul{103.73} & 98.17 \\

\rowcolor{oursblue}
& \textbf{Ours}
& \textbf{1.024} & \ul{1.095} & \textbf{1.183} & \ul{1.101}
& \textbf{1.085} & \ul{1.154} & \textbf{1.206} & \ul{1.148}
& \ul{95.80} & \textbf{114.76} & \textbf{105.24} & \textbf{104.98}
\\

\midrule

% ============================================================
% SKETCHY
% ============================================================
\multirow{6}{*}{\textbf{Sketchy}}
& Sketch-a-Sketch
& 0.384 & 0.377 & 0.407 & 0.389
& 0.509 & 0.505 & 0.516 & 0.510
& 96.54 & 76.17 & 96.74 & 89.27 \\

& SDXS
& 0.675 & 0.644 & 0.674 & 0.664
& 0.646 & 0.634 & 0.657 & 0.646
& 98.37 & 80.71 & 99.99 & 92.59 \\

& Cond.\ Balance
& 0.759 & 0.728 & \ul{0.752} & 0.746
& 0.773 & 0.758 & \ul{0.780} & \ul{0.770}
& 86.37 & 95.70 & \ul{101.65} & 94.36 \\

& Flux Baseline
& 0.634 & 0.591 & 0.613 & 0.613
& 0.587 & 0.556 & 0.556 & 0.566
& \ul{100.00} & \ul{100.00} & 100.00 & \ul{100.00} \\

& StableFlow
& \ul{0.843} & \ul{0.764} & 0.657 & \ul{0.755}
& \textbf{0.813} & \ul{0.772} & 0.674 & 0.753
& 86.62 & 90.37 & 98.94 & 91.84 \\

\rowcolor{oursblue}
& \textbf{Ours}
& \textbf{0.868} & \textbf{0.886} & \textbf{0.928} & \textbf{0.894}
& \ul{0.795} & \textbf{0.830} & \textbf{0.881} & \textbf{0.835}
& \textbf{106.40} & \textbf{107.77} & \textbf{111.16} & \textbf{108.42}
\\

\bottomrule
\end{tabular}
}
\vspace{-8pt}
\end{table*}
\vspace{-0.55em}

% We evaluate generation quality at $\{10,50,100\}\%$ sketch completion (\cref{tab:quantitative}) and progressive behaviour over $\{10,20,30,50,70,80,100\}\%$ (\cref{tab:stability}), preserving stroke order and captions. Distributional quality is measured with FID-I and FID-C~\citep{szegedy2015going,radford2021learning}; paired semantic fidelity with DINOv2~\citep{oquab2023dinov2}; and text alignment with CLIPScore~\citep{radford2021learning} and T5-Cos (BLIP-2 captions~\citep{li2023blip} + T5 embeddings~\citep{raffel2020exploring}). For progressive evaluation, we use Progressive Fidelity Coherence (PFC) to summarize semantic fidelity and turn-to-turn coherence. We also report cross-completion robustness in distributional (Dist.), visual (Vis.), and textual (Text) dimensions; full definitions and raw consecutive-frame DINOv2/LPIPS results are in the supplementary.\\
\noindent\textbf{Evaluation Protocol and Metrics.}
\label{sec:eval_protocol}
We evaluate generation quality at specific sketch-completion stages (Table~\ref{tab:quantitative}) and progressive stability across representative transitions (Table~\ref{tab:stability}), while preserving stroke order and captions. We measure distributional quality with FID-I \citep{szegedy2015going}
and FID-C \citep{kynkaanniemirole}, semantic fidelity with DINOv2
\citep{oquab2023dinov2}, and text alignment with CLIPScore
\citep{hessel2021clipscore,radford2021learning} and T5-Cos
(BLIP-2 captions \citep{li2023blip} + T5 embeddings
\citep{raffel2020exploring}). For progression, PFC-DINO and PFC-LPIPS measure temporal stability. We additionally report quality-adjusted cross-completion robustness in distributional (Dist.), visual (Vis.), textual (Text), and overall dimensions, normalized to the FLUX baseline ($100$). Full definitions are provided in Appendix~\ref{sec:supp_progressive_metrics}.\\
% For our method, each completion stage feeds the previous output as tile memory and warm-start latent (Sec.~\cref{sec:pcm}), mirroring the interactive stroke-by-stroke regime; baselines regenerate independently at each stage.
% We quantify generation along three complementary axes. \\
\vspace{-1.5em}
\subsection{Quantitative Results}
\vspace{-0.5em}
We compare against Sketch-a-Sketch (S-a-S)~\citep{sarukkai2024block},
SDXS~\citep{song2024sdxs},
Conditional Balance (CB)~\citep{cohen2025conditional},
FLUX+ControlNet~\citep{flux2024,zhang2023controlnet},
and StableFlow (SF)~\citep{avrahami2025stableflow}. \cref{tab:quantitative} evaluates generation quality at {10\%, 50\%, 100\%} sketch completion across three sketch datasets. ProgressNet maintains strong semantic fidelity: DINOv2 is highest or tied-highest at nearly all completion levels, with competitive CLIP/T5 alignment. This suggests the method stays faithful to the target image and prompt throughout the drawing, rather than only at the final drawing state. \cref{tab:stability} presents progressive fidelity and inter-turn coherence using PFC-DINO and PFC-LPIPS, which assess coherence while also accounting for quality. ProgressNet obtains the highest mean PFC-DINO/PFC-LPIPS on FS-COCO and Sketchy, and the highest final-transition scores across all three sketch domains. The cross-completion indices additionally summarize quality and its variation across completion levels.
\begin{figure*}[!t]
\centering
% =========================================================
% Figure 7
% =========================================================
% \includegraphics[width=\linewidth]{images/qualitative_part1_v3_iclr.pdf}
\includegraphics[width=\linewidth]{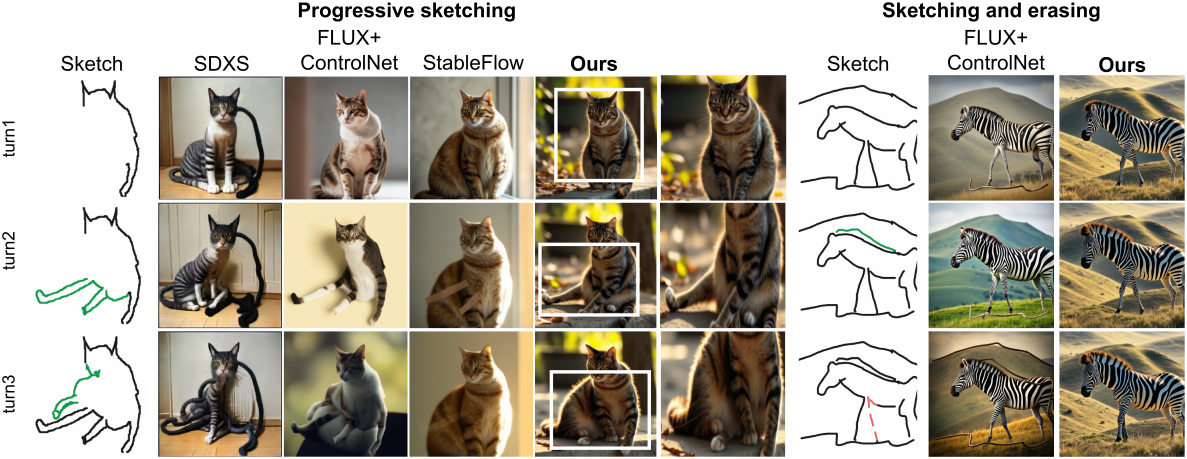}
% \vspace{-2pt}
\caption{
Progressive sketching (left) and sketch removal (right).
Competing methods frequently drift or introduce structural artifacts as
strokes evolve, while ProgressNet preserves appearance and scene consistency
while adapting to added or removed strokes.
}
\label{fig:qualitative_main_part1}
% \vspace{1em}
% =========================================================
% Figure 8
% =========================================================
% \includegraphics[width=\linewidth]{images/fig8_final_v1_cropped_iclr.pdf}
% \includegraphics[width=\linewidth]{images/fig8_final_v2_cropped_iclr_v3.pdf}
\includegraphics[width=\linewidth]{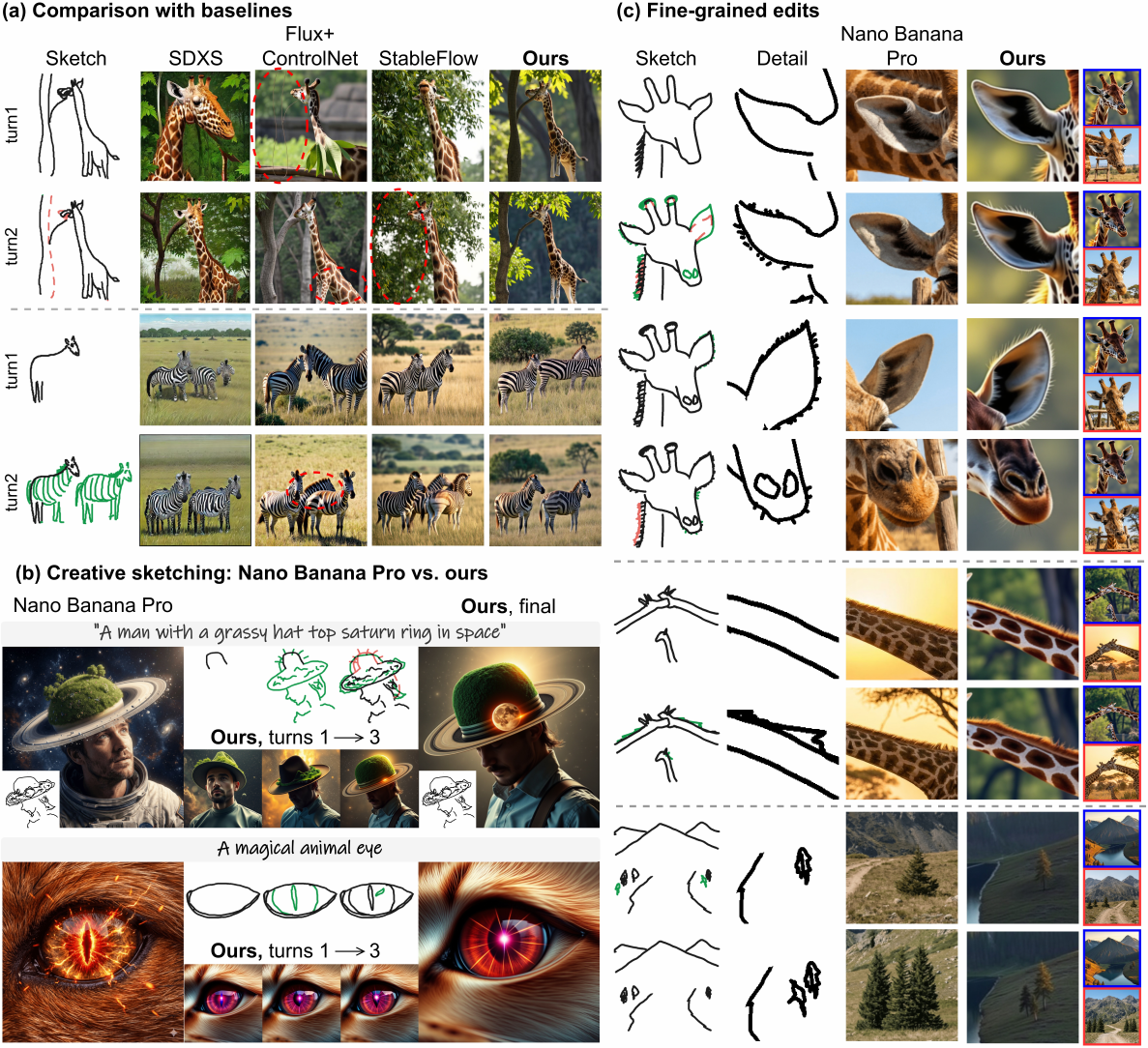}
% \vspace{-18pt}
\caption{
Qualitative results.
\textbf{Top-left}: Cross-method comparison on progressive giraffe and zebra
sketching; red circles indicate temporal drift and artifacts, while ours
remains coherent.
\textbf{Bottom-left}: Creative comparison with Nano Banana Pro, where ours
better preserves the evolving sketch structure and intended concept.
\textbf{Right}: Fine-grained progressive edits to the giraffe ear, mane, and
scenery; ours performs localized adaptation while maintaining global
consistency.
}
\label{fig:qualitative_main_part2}
\par\vspace{-2em}
\end{figure*}
% \textbf{Qualitative Results.}
% \label{sec:qualitative}
% \Cref{fig:qualitative_main_part1,fig:qualitative_main_part2} examine progressive sketching, removal, local editing, and open-ended freehand creation. Across the cat, zebra, and giraffe sequences, competing methods
% % \par\vspace{-3.5em}
% frequently alter pose, appearance, or background as the sketch evolves, whereas ProgressNet preserves a recognizable visual state while incorporating newly introduced structure. This behaviour 
% also extends to stroke removal and fine-grained edits \eg corrected zebra, giraffe ear and mane
% \vspace{-1em}
% \vspace{-1.5em}
\subsection{Qualitative Results}
\vspace{-0.5em}
% \vspace{-0.2em}
\label{sec:qualitative}
\cref{fig:qualitative_main_part1,fig:qualitative_main_part2} examine how progressive
memory supports generation as the sketch is continuously extended, revised, or
locally edited. In \cref{fig:qualitative_main_part1}, comparisons against standard
baselines show that ProgressNet better preserves appearance and scene context
during progressive sketching and remains coherent when previously drawn strokes
are removed. \cref{fig:qualitative_main_part2} extends this evaluation to
cross-method comparisons and more demanding interactions against Nano Banana Pro,
including fine-grained giraffe and scene-level edits as well as open-ended creative
generation. These examples also show reduced temporal drift and stroke-induced
artifacts under progressive updates. Across these cases, the progressive sketching
process allows local structure to evolve without repeatedly reinterpreting the
established image, while the retained visual state provides an important basis
for subsequent edits. Together, these illustrate the practical benefit of progressive
memory: users can incrementally develop, revise, and refine an image while preserving
the context established earlier.\\
\vspace{-1em}
\subsection{User study}
\vspace{-0.5em}
\begin{wraptable}{r}{0.4\columnwidth}
\vspace{-2.5em}
\centering
\caption{User study preferences.}
\par\vspace{-0.6em}
\label{tab:user_study}
\begin{threeparttable}
\definecolor{oursblue}{RGB}{230,245,255}
\setlength{\tabcolsep}{1pt}
\renewcommand{\arraystretch}{0.8}
\resizebox{0.4\columnwidth}{!}{%
\scriptsize
\begin{tabular}{l ccc cccccc}
\toprule
& \multicolumn{3}{c}{\textbf{Study 1: Overall}} & \multicolumn{6}{c}{\textbf{Study 2: Operation-based}} \\
\cmidrule(lr){2-4}\cmidrule(lr){5-10}
  & \rotatebox{90}{\scriptsize Pref.}
  & \rotatebox{90}{\scriptsize Coher.}
  & \rotatebox{90}{\scriptsize Qual.}
  & \rotatebox{90}{\scriptsize Erase}
  & \rotatebox{90}{\scriptsize Edit}
  & \rotatebox{90}{\scriptsize Prog.}
  & \rotatebox{90}{\scriptsize F-Draw}
  & \rotatebox{90}{\scriptsize F-Erase}
  & \rotatebox{90}{\scriptsize Preserv.} \\
\midrule
% S-a-S~\citep{sarukkai2024block}  & 2.9 & 4.9 & 3.4 & 1.4 & 1.9 & 3.0 & 5.1 & 4.2 & 3.8 \\
% SDXS~\citep{song2024sdxs}   & 4.1 & 13.2 & 4.3 & 1.8 & 2.3 & 4.2 & 7.1 & 5.8 & 5.1 \\
% CB~\citep{cohen2025conditional}     & 5.1 & 10.1 & 7.2 & 2.1 & 4.1 & 12.1 & 7.6 & 7.2 & 7.4 \\
% Flux~\citep{flux2024}   & 5.9 & 6.8 & 10.2 & 2.8 & 5.3 & 7.6 & 8.1 & 7.8 & 8.2 \\
% SF~\citep{avrahami2025stableflow}     & 10.2 & 20.3 & 15.1 & 6.7 & 8.8 & 18.3 & 22.4 & 19.7 & 17.6 \\
S-a-S  & 2.9 & 4.9 & 3.4 & 1.4 & 1.9 & 3.0 & 5.1 & 4.2 & 3.8 \\
SDXS   & 4.1 & 13.2 & 4.3 & 1.8 & 2.3 & 4.2 & 7.1 & 5.8 & 5.1 \\
CB     & 5.1 & 10.1 & 7.2 & 2.1 & 4.1 & 12.1 & 7.6 & 7.2 & 7.4 \\
Flux   & 5.9 & 6.8 & 10.2 & 2.8 & 5.3 & 7.6 & 8.1 & 7.8 & 8.2 \\
SF     & 10.2 & 20.3 & 15.1 & 6.7 & 8.8 & 18.3 & 22.4 & 19.7 & 17.6 \\
\rowcolor{oursblue}
\textbf{Ours} & \textbf{71.8} & \textbf{44.7} & \textbf{59.8} & \textbf{85.2} & \textbf{77.6} & \textbf{54.8} & \textbf{49.7} & \textbf{55.3} & \textbf{57.9} \\
\bottomrule
\end{tabular}}
\begin{tablenotes}
\scriptsize
\item \hspace{-0.4cm}\textbf{Bold}: best, \ul{underline}: second best.
\vspace{-4em}
\end{tablenotes}
% \vspace{-2pt}
\end{threeparttable}
\end{wraptable}
We perform two user studies to compare outputs across preference, temporal coherency, generation quality, erasure, and prompt editing where ProgressNet is consistently preferred over all baselines (\cref{tab:user_study}). StableFlow is the strongest competitor on coherency and fine-drawing, while ProgressNet leads every reported criterion, with its largest share on erasure.
\subsection{Ablation Studies}
\begin{wrapfigure}{r}{0.4\textwidth}
\vspace{-0.3em}
\hspace*{-4pt}
\centering
\includegraphics[width=\linewidth]{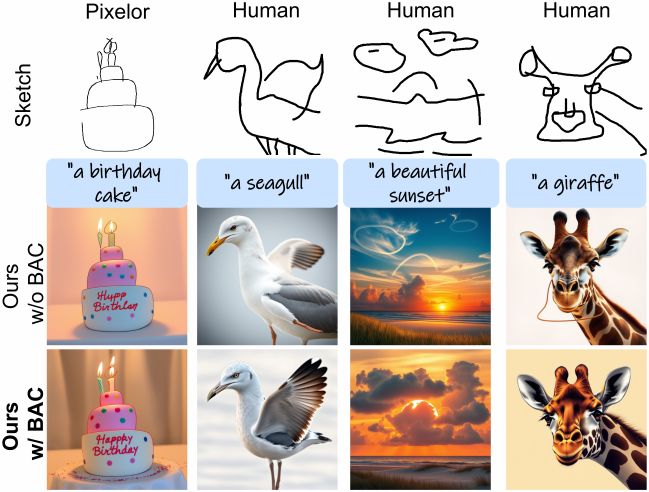}
\vspace{-10pt}
\caption{Without BAC, sketch-edge artifacts leak under prompt and sketch conflict or imperfect sketches from Pixelor~\citep{bhunia2020pixelor} and human drawn sketches. BAC reduces visible artifacts associated with these difficult sketch-prompt configurations.}
\label{fig:bac_ablation}
\vspace{-2em}
\end{wrapfigure}

\vspace{-0.5em}
We evaluate cumulative additions to the baseline
(\cref{tab:ablation}), interpreting aggregate metrics together with
layer-selection comparisons (\cref{fig:lskvi_exp,fig:supp_lskvi_qual}) and the
with/without-BAC examples (\cref{fig:bac_ablation}).
% We enable PCM, LS-KVI, and BAC over the FLUX baseline
% (\cref{tab:ablation}), exposing their complementary roles rather than expecting
% monotonic gains on every individual metric.
\vspace{-0.6em}
\begin{table*}[!ht]
\centering
\caption{
Ablation on FS-COCO across sketch completion levels.
Each row cumulatively enables PCM, LS-KVI, and BAC.
\textbf{Bold}: best; \ul{underline}: second best.
}
\label{tab:ablation}
\vspace{-6pt}

\begin{threeparttable}

\definecolor{addblack}{RGB}{30,80,160}
\definecolor{addpill}{RGB}{220,235,255}
\definecolor{oursblue}{RGB}{230,245,255}

\setlength{\tabcolsep}{1.2pt}
\renewcommand{\arraystretch}{0.3}

\resizebox{\textwidth}{!}{%
\scriptsize
\begin{tabular}{l ccc | ccccc | ccccc | c}
\toprule
& & & &
\multicolumn{5}{c|}{\textsc{50\% Completion}} &
\multicolumn{5}{c|}{\textsc{100\% Completion}} &
\multicolumn{1}{c}{} \\
\cmidrule(lr){5-9}
\cmidrule(lr){10-14}

Variant
& {\scriptsize PCM}
& {\scriptsize KV}
& {\scriptsize BAC}

& {\scriptsize FID-I $\downarrow$}
& {\scriptsize FID-C $\downarrow$}
& {\scriptsize DINOv2 $\uparrow$}
& {\scriptsize CLIP $\uparrow$}
& {\scriptsize T5 $\uparrow$}

& {\scriptsize FID-I $\downarrow$}
& {\scriptsize FID-C $\downarrow$}
& {\scriptsize DINOv2 $\uparrow$}
& {\scriptsize CLIP $\uparrow$}
& {\scriptsize T5 $\uparrow$}

& {\scriptsize Latency $\downarrow$} \\
\midrule

% ============================================================
% BASELINE
% ============================================================
Baseline
& \ding{55} & \ding{55} & \ding{55}

& 44.7
& 0.188
& 0.386
& 0.285
& 0.642

& 67.9
& 0.264
& 0.361
& 0.265
& 0.641

& \textbf{0.761} \\

% ============================================================
% + PCM
% ============================================================
+ PCM
& \ding{51} & \ding{55} & \ding{55}

& \textbf{33.9}
& \ul{0.145}
& \textbf{0.420}
& \ul{0.304}
& \ul{0.654}

& \textbf{36.6}
& \ul{0.150}
& \ul{0.414}
& \ul{0.301}
& \ul{0.653}

& \ul{0.816}\;
\fcolorbox{addblack}{addpill}
{\scriptsize\textcolor{addblack}{\textbf{+0.055}}} \\

% ============================================================
% + LS-KVI
% ============================================================
+ LS-KVI
& \ding{51} & \ding{51} & \ding{55}

& \ul{35.7}
& 0.161
& \ul{0.415}
& 0.303
& \ul{0.654}

& 37.9
& 0.162
& \textbf{0.415}
& 0.300
& \ul{0.653}

& \ul{0.816}\;
\fcolorbox{addblack}{addpill}
{\scriptsize\textcolor{addblack}{\textbf{+0.055}}} \\

% ============================================================
% OURS
% ============================================================
\rowcolor{oursblue}
\textbf{Ours}
& \ding{51} & \ding{51} & \ding{51}

& 36.0
& \textbf{0.141}
& \ul{0.415}
& \textbf{0.305}
& \textbf{0.656}

& \ul{37.3}
& \textbf{0.143}
& 0.413
& \textbf{0.303}
& \textbf{0.655}

& 0.829\;
\fcolorbox{addblack}{addpill}
{\scriptsize\textcolor{addblack}{\textbf{+0.068}}} \\

\bottomrule
\end{tabular}
}

\end{threeparttable}
\par\vspace{-0.5em}
\end{table*}

\vspace{-0.6em}
\noindent\textbf{(i) Importance of PCM.}
PCM supplies the dominant improvement in reference and joint fidelity.
At full completion, FID-I decreases substantially from 67.9 to 36.6, while DINOv2 increases from 0.361 to 0.414. It establishes the appearance
memory on top of which the remaining components are evaluated.

\noindent \textbf{(ii) Impact of Layer-Selective K/V Injection.}
Adding LS-KVI to PCM maintains comparable DINOv2 and text alignment, with a
trade-off in FID-I and FID-C. Separately, \cref{fig:lskvi_exp,fig:supp_lskvi_qual}
show that the selected non-vital-layer configuration better
preserves appearance than the vital-layer
configuration in the illustrated sequences.
% The table
% evaluates the marginal effect of feature reuse, while
% the figures support it through qualitative comparison.
% Beyond feature reuse,  shows that
% invalidating stale LS-KVI state enables more reliable recovery
% from stroke removal. 
For stroke removal, \cref{tab:supp_erasure} validates LS-KVI state management: clearing stale K/V enables substantially stronger recovery toward the pre-edit state.

\noindent\textbf{(iii) Impact of Banded Adaptive Control.}
Adding BAC yields modest quantitative gains in FID-C and CLIP/T5 across later completion stages. At full completion, FID-C improves from 0.162 to 0.143, while CLIP/T5 increase from 0.300/0.653 to 0.303/0.655. These metrics may not fully capture its main effect: reducing visible sketch-edge artifacts under conflicting or imperfect strokes by adapting the
strength of appearance-memory conditioning. Accordingly, \cref{fig:bac_ablation} shows this correction directly, supporting an essential role for adaptive control beyond uniform metric improvement.

\vspace{-0.2em}
\section{Conclusion and Limitations}
\label{sec:conclusion}
\vspace{-8pt}
We proposed ProgressNet, a training-free framework for progressive generation that stabilizes interactions between evolving strokes and prompts. PCM and LS-KVI preserve temporal consistency, while BAC uses retrospective memory-feature divergence to adapt memory. ProgressNet enables progressive drawing by improving fidelity and alignment across completion levels with near-real-time performance. Our work opens avenues for more modality-agnostic progressive generation.

\vspace{-0.3em}
\noindent\textbf{Limitations.}
\textit{(i)} ProgressNet is primarily evaluated with sketch-based control; extending its memory and adaptive-control mechanisms to other modalities, such as depth and pose, remains unexplored.
\textit{(ii)} BAC uses predefined layer bands and modulation bounds; automatically identifying control-sensitive regions on different conditioning modalities is a promising future direction.
\section*{Acknowledgements}
This work was supported by the Arts and Humanities Research Council (AHRC) through the CoSTAR National Lab (Grant Ref. AH/Y001060/1). We gratefully acknowledge the support and resources provided by the CoSTAR National Lab that contributed to this research.

\newpage
\appendix
% \section{Appendix}
% Appendix: letter-numbered sections (A, B, ...) with A.1, A.2, ... subsections
\appendix
\section{Quantitative Evaluations}
\label{sec:supp_quant}
\vspace{-8pt}
% Figures and tables continue the main paper's numbering (no counter reset, no S prefix).

\subsection{Dataset and Progressive Evaluation Protocol}
\label{sec:supp_protocol}
We evaluate three complementary sketch domains using fixed evaluation subsets and identical inputs for every compared method. FS-COCO~\citep{chowdhury2022fs} contributes 10,000 temporally ordered freehand scene sketches with paired photographs and captions. Photo-Sketching~\citep{li2019photo} contributes 500 contour-sketch/image pairs from its 5,000 available pairs. Sketchy~\citep{sangkloy2016sketchy} contributes 1,250 single-object freehand sketches, stratified uniformly across all 125 categories with 10 samples per category. The resulting generation-quality evaluation contains 11,750 samples. For each sample, the ordered sketch sequence and associated text condition are fixed once and shared across all methods; progressive states are evaluated at $\{10,20,30,50,70,80,100\}\%$ completion.

\begin{table}[!htb]
\centering
\caption{Datasets used for generation-quality and progressive evaluation. The same sample identities and progressive states are supplied to every method.}
\label{tab:supp_dataset_protocol}
\vspace{-5pt}
\setlength{\tabcolsep}{4pt}
\small
\begin{tabular}{lcll}
\toprule
Dataset & Eval. $N$ & Sketch domain & Sampling / pairing \\
\midrule
FS-COCO & 10,000 & Freehand scenes & Full evaluation set; paired image/text \\
Photo-Sketching & 500 & Contour tracing & Evaluation subset of paired sketch/image data \\
Sketchy & 1,250 & Freehand objects & 10/category across 125 categories \\
\bottomrule
\end{tabular}
\end{table}
% IMPORTANT: if Photo-Sketching or Sketchy use a reconstructed rather than native stroke order,
% add the exact ordering rule here before submission. Likewise state the exact source of their
% text condition if it is not already documented in the main paper/code release.

\subsection{Progressive Evaluation Metrics}
\label{sec:supp_progressive_metrics}

Let $I_{i,k}$ denote the generated image for sample $i$ at sketch-completion level $k$. For a progressive transition $a\rightarrow b$, we measure consecutive-frame semantic coherence using DINOv2 cosine similarity,
\begin{equation}
D_{a\rightarrow b}
=
\frac{1}{N}
\sum_{i=1}^{N}
\cos\!\left(
\phi_{\mathrm{DINO}}(I_{i,a}),
\phi_{\mathrm{DINO}}(I_{i,b})
\right),
\label{eq:pfc_dino_temporal}
\end{equation}
and perceptual change using LPIPS,
\begin{equation}
L_{a\rightarrow b}
=
\frac{1}{N}
\sum_{i=1}^{N}
\mathrm{LPIPS}(I_{i,a},I_{i,b}).
\label{eq:pfc_lpips_temporal}
\end{equation}

\paragraph{Quality-adjusted progressive fidelity--coherence.}
We evaluate temporal coherence jointly with the distributional quality of the corresponding sketch-completion regime. To use the same reported quality anchors for all three datasets, we associate the three representative progressive transitions
\begin{equation}
\tau_{10}=10\!\rightarrow\!20,\qquad
\tau_{50}=30\!\rightarrow\!50,\qquad
\tau_{100}=80\!\rightarrow\!100
\label{eq:pfc_transitions}
\end{equation}
with FID-I measured at the corresponding reported anchors
$q\in\mathcal{A}=\{10,50,100\}$.
Let $\mathrm{FID}_{I,q}$ denote FID-I at anchor $q$.

We define the DINO-based progressive fidelity--coherence score as
\begin{equation}
\mathrm{PFC\text{-}DINO}_{q}
=
\frac{100\,D_{\tau_q}}{\mathrm{FID}_{I,q}},
\qquad q\in\mathcal{A},
\label{eq:pfc_dino}
\end{equation}
and its LPIPS counterpart as
\begin{equation}
\mathrm{PFC\text{-}LPIPS}_{q}
=
\frac{100}
{\mathrm{FID}_{I,q}\left(1+L_{\tau_q}\right)}.
\label{eq:pfc_lpips}
\end{equation}
Both scores are higher-is-better. The factor $100$ only sets a convenient numerical scale and does not affect rankings. These quantities are quality-adjusted composite summaries rather than pure temporal-coherence measures. PFC-DINO combines high semantic inter-turn similarity with strong distributional quality, while PFC-LPIPS analogously combines low perceptual change with the same quality adjustment. Their reported means are
\begin{equation}
\overline{P}_{D}
=
\frac{1}{|\mathcal{A}|}
\sum_{q\in\mathcal{A}}
\mathrm{PFC\text{-}DINO}_{q},
\qquad
\overline{P}_{L}
=
\frac{1}{|\mathcal{A}|}
\sum_{q\in\mathcal{A}}
\mathrm{PFC\text{-}LPIPS}_{q}.
\label{eq:pfc_means}
\end{equation}

\paragraph{Cross-completion robustness.}
PFC evaluates selected consecutive transitions, whereas cross-completion robustness measures whether standard generation metrics remain both strong and stable over the three common completion anchors $\mathcal{A}=\{10,50,100\}$. For each metric $x$, let
\begin{equation}
\mu_x
=
\frac{1}{|\mathcal{A}|}
\sum_{q\in\mathcal{A}}x_q,
\qquad
\mathrm{CV}_x
=
\frac{\sigma_x}{\mu_x},
\label{eq:cross_cv}
\end{equation}
where $\sigma_x$ is the population standard deviation across the three anchors.

For the two distributional metrics, we first define their relative variation
\begin{equation}
V_{\mathrm{Dist}}
=
\sqrt{
\frac{
\mathrm{CV}_{\text{FID-I}}^2+
\mathrm{CV}_{\text{FID-C}}^2
}{2}},
\label{eq:cross_dist_var}
\end{equation}
and combine this with their average quality:
\begin{equation}
Q_{\mathrm{Dist}}
=
\frac{1}{
\sqrt{
\mu_{\text{FID-I}}
\mu_{\text{FID-C}}
}
\left(1+V_{\mathrm{Dist}}\right)
}.
\label{eq:cross_dist}
\end{equation}

Visual robustness uses paired DINOv2 fidelity,
\begin{equation}
Q_{\mathrm{Vis}}
=
\frac{
\mu_{\mathrm{DINO}}
}{
1+\mathrm{CV}_{\mathrm{DINO}}
},
\label{eq:cross_vis}
\end{equation}
while textual robustness jointly considers CLIPScore and T5-Cos:
\begin{equation}
V_{\mathrm{Text}}
=
\sqrt{
\frac{
\mathrm{CV}_{\mathrm{CLIP}}^2+
\mathrm{CV}_{\mathrm{T5}}^2
}{2}},
\qquad
Q_{\mathrm{Text}}
=
\frac{
\sqrt{
\mu_{\mathrm{CLIP}}
\mu_{\mathrm{T5}}
}
}{
1+V_{\mathrm{Text}}
}.
\label{eq:cross_text}
\end{equation}

To provide a common and interpretable scale across sketch domains, each robustness axis is normalized to the FLUX baseline evaluated on the same dataset:
\begin{equation}
R_x
=
100
\frac{Q_x}{Q_x^{\mathrm{Flux}}},
\qquad
x\in
\{\mathrm{Dist},\mathrm{Vis},\mathrm{Text}\}.
\label{eq:cross_normalized}
\end{equation}
Thus, $100$ denotes the FLUX baseline and values above $100$ indicate stronger quality-adjusted cross-completion robustness. Finally, we summarize the three complementary axes by their geometric mean,
\begin{equation}
R_{\mathrm{Overall}}
=
\left(
R_{\mathrm{Dist}}
R_{\mathrm{Vis}}
R_{\mathrm{Text}}
\right)^{1/3}.
\label{eq:cross_overall}
\end{equation}
All four cross-completion scores are higher-is-better. Unlike PFC, these quantities measure how well standard generation quality and alignment are maintained across completion levels.

\subsection{Extended Quantitative Results}
\label{sec:supp_extended}
% \vspace{-4pt}
Along with the main paper, we additionally record results for quantitative comparison and ablation at intermediate sketch completion levels.~\cref{tab:supp_intermediate} extends the six-method comparison of main paper \cref{tab:quantitative} to the full seven-level sweep $\{10, 20, 30, 50, 70, 80, 100\}\%$, revealing how each method's distributional quality and semantic fidelity evolve as strokes progressively accumulate.~\cref{tab:supp_ablation} correspondingly extends the component ablation of main paper \cref{tab:ablation} to the intermediate levels, isolating the individual contribution of each proposed component, PCM, LS-KVI, and BAC, at every stage of sketch completion. Together, these tables show that the gains reported at the three anchor levels $\{10, 50, 100\}\%$ are not an artifact of those selected levels. 
% On FS-COCO, ProgressNet leads or matches the strongest reported value on each metric from 20\% completion onward, while the main-paper cross-dataset results provide the corresponding evaluation on Photo-Sketching and Sketchy.

\begin{table*}[!h]
\centering
\vspace{10pt}
\caption{Extended comparison at intermediate sketch completion levels on FS-COCO. Results at $\{10,50,100\}\%$ are in the main paper (\cref{tab:quantitative}).}
\label{tab:supp_intermediate}
\vspace{-8pt}
\begin{threeparttable}
\definecolor{oursblue}{RGB}{230,245,255}
\setlength{\tabcolsep}{1.5pt}
\renewcommand{\arraystretch}{1.3}
\resizebox{\textwidth}{!}{%
\scriptsize
\begin{tabular}{l cccccc cccccc cccccc cccccc}
\toprule
 & \multicolumn{6}{c}{\textbf{20\% Completion}} & \multicolumn{6}{c}{\textbf{30\% Completion}} & \multicolumn{6}{c}{\textbf{70\% Completion}} & \multicolumn{6}{c}{\textbf{80\% Completion}} \\
\cmidrule(lr){2-7}\cmidrule(lr){8-13}\cmidrule(lr){14-19}\cmidrule(lr){20-25}
Metric & \rotatebox{90}{\scriptsize S-a-S~\citep{sarukkai2024block}} & \rotatebox{90}{\scriptsize SDXS~\citep{song2024sdxs}} & \rotatebox{90}{\scriptsize CB~\citep{cohen2025conditional}} & \rotatebox{90}{\scriptsize Flux~\citep{flux2024}} & \rotatebox{90}{\scriptsize SF~\citep{avrahami2025stableflow}} & \cellcolor{oursblue}\rotatebox{90}{\scriptsize \textbf{Ours}} & \rotatebox{90}{\scriptsize S-a-S~\citep{sarukkai2024block}} & \rotatebox{90}{\scriptsize SDXS~\citep{song2024sdxs}} & \rotatebox{90}{\scriptsize CB~\citep{cohen2025conditional}} & \rotatebox{90}{\scriptsize Flux~\citep{flux2024}} & \rotatebox{90}{\scriptsize SF~\citep{avrahami2025stableflow}} & \cellcolor{oursblue}\rotatebox{90}{\scriptsize \textbf{Ours}} & \rotatebox{90}{\scriptsize S-a-S~\citep{sarukkai2024block}} & \rotatebox{90}{\scriptsize SDXS~\citep{song2024sdxs}} & \rotatebox{90}{\scriptsize CB~\citep{cohen2025conditional}} & \rotatebox{90}{\scriptsize Flux~\citep{flux2024}} & \rotatebox{90}{\scriptsize SF~\citep{avrahami2025stableflow}} & \cellcolor{oursblue}\rotatebox{90}{\scriptsize \textbf{Ours}} & \rotatebox{90}{\scriptsize S-a-S~\citep{sarukkai2024block}} & \rotatebox{90}{\scriptsize SDXS~\citep{song2024sdxs}} & \rotatebox{90}{\scriptsize CB~\citep{cohen2025conditional}} & \rotatebox{90}{\scriptsize Flux~\citep{flux2024}} & \rotatebox{90}{\scriptsize SF~\citep{avrahami2025stableflow}} & \cellcolor{oursblue}\rotatebox{90}{\scriptsize \textbf{Ours}} \\
\midrule
FID-I $\downarrow$ & 195.1 & 39.8 & 198.4 & 37.0 & \ul{36.5} & \cellcolor{oursblue}\textbf{35.8} & 202.9 & 40.2 & 193.4 & 38.5 & \ul{38.3} & \cellcolor{oursblue}\textbf{36.4} & 210.7 & 44.0 & 203.4 & 52.5 & \ul{43.6} & \cellcolor{oursblue}\textbf{36.1} & 212.6 & \ul{45.7} & 208.5 & 58.0 & 47.6 & \cellcolor{oursblue}\textbf{36.7} \\
FID-C $\downarrow$ & 0.442 & \ul{0.163} & 0.406 & \ul{0.163} & 0.190 & \cellcolor{oursblue}\textbf{0.158} & 0.433 & \ul{0.165} & 0.396 & 0.166 & 0.198 & \cellcolor{oursblue}\textbf{0.144} & 0.465 & \ul{0.177} & 0.398 & 0.212 & 0.203 & \cellcolor{oursblue}\textbf{0.142} & 0.470 & \ul{0.180} & 0.421 & 0.229 & 0.209 & \cellcolor{oursblue}\textbf{0.141} \\
DINOv2 $\uparrow$ & 0.288 & 0.333 & 0.331 & 0.379 & \ul{0.392} & \cellcolor{oursblue}\textbf{0.402} & 0.281 & 0.336 & 0.349 & 0.383 & \ul{0.387} & \cellcolor{oursblue}\textbf{0.409} & 0.290 & 0.335 & 0.319 & \ul{0.381} & 0.378 & \cellcolor{oursblue}\textbf{0.415} & 0.289 & 0.334 & 0.313 & \ul{0.375} & 0.370 & \cellcolor{oursblue}\textbf{0.415} \\
CLIPScore $\uparrow$ & 0.256 & 0.299 & 0.292 & 0.294 & \ul{0.300} & \cellcolor{oursblue}\textbf{0.304} & 0.266 & \ul{0.298} & 0.291 & 0.291 & 0.297 & \cellcolor{oursblue}\textbf{0.305} & 0.266 & \ul{0.295} & 0.291 & 0.278 & 0.294 & \cellcolor{oursblue}\textbf{0.304} & 0.252 & \ul{0.294} & 0.289 & 0.274 & 0.291 & \cellcolor{oursblue}\textbf{0.304} \\
T5-Cos $\uparrow$ & 0.630 & 0.648 & 0.644 & 0.648 & \ul{0.656} & \cellcolor{oursblue}\textbf{0.658} & 0.623 & 0.648 & 0.639 & 0.646 & \ul{0.656} & \cellcolor{oursblue}\textbf{0.657} & 0.629 & 0.646 & 0.644 & 0.642 & \ul{0.654} & \cellcolor{oursblue}\textbf{0.656} & 0.629 & 0.646 & 0.630 & 0.641 & \ul{0.654} & \cellcolor{oursblue}\textbf{0.655} \\
\bottomrule
\end{tabular}}
\begin{tablenotes}
\scriptsize
\item \hspace{-0.4cm}\textbf{Bold}: best, \ul{underline}: second best.
\end{tablenotes}
\end{threeparttable}
\end{table*}
% \vspace{-6pt}
\paragraph{On FID-I at 10\% completion.} At 10\% sketch completion, sketches are highly sparse and provide limited structural constraint. ProgressNet ($36.5$) is therefore marginally above the Flux Baseline ($32.9$) at this stage. From $20\%$ completion onward, ProgressNet matches or leads on FID-I and FID-C across all completion levels, consistent with effective appearance anchoring as the sketch becomes more informative. 
% \vspace{-4pt}
\paragraph{On T5 alignment at early completion.} At 10\% completion, StableFlow marginally leads on T5-Cos ($0.659$ vs.\ $0.657$); the two methods are tied at $50\%$ ($0.656$ each), with ProgressNet leading at full completion ($0.655$ vs.\ $0.654$). Moreover, ProgressNet leads on every intermediate completion level as presented in~\cref{tab:supp_intermediate}. At the very sparse 10\% stage, text-alignment measurements are more sensitive to the under-specified visual condition. From $20\%$ completion onward, where the sketch provides stronger structural evidence, ProgressNet consistently leads or matches on the reported text-alignment metrics.

\begin{table*}[!h]
\centering
\vspace{10pt}
\caption{Extended ablation at intermediate completion levels on FS-COCO. Results at $\{50,100\}\%$ are in the main paper (\cref{tab:ablation}).}
\label{tab:supp_ablation}
\vspace{-8pt}
\begin{threeparttable}
\definecolor{oursblue}{RGB}{230,245,255}
\setlength{\tabcolsep}{1.5pt}
\renewcommand{\arraystretch}{1.3}
\resizebox{\textwidth}{!}{%
\scriptsize
\begin{tabular}{lccc ccccc ccccc ccccc ccccc ccccc}
\toprule
 & & & & \multicolumn{5}{c}{\textbf{10\% Completion}} & \multicolumn{5}{c}{\textbf{20\% Completion}} & \multicolumn{5}{c}{\textbf{30\% Completion}} & \multicolumn{5}{c}{\textbf{70\% Completion}} & \multicolumn{5}{c}{\textbf{80\% Completion}} \\
\cmidrule(lr){5-9}\cmidrule(lr){10-14}\cmidrule(lr){15-19}\cmidrule(lr){20-24}\cmidrule(lr){25-29}
Variant & \rotatebox{90}{PCM} & \rotatebox{90}{KV} & \rotatebox{90}{BAC} & \rotatebox{90}{FID-I$\downarrow$} & \rotatebox{90}{FID-C$\downarrow$} & \rotatebox{90}{DINO$\uparrow$} & \rotatebox{90}{CLIP$\uparrow$} & \rotatebox{90}{T5$\uparrow$} & \rotatebox{90}{FID-I$\downarrow$} & \rotatebox{90}{FID-C$\downarrow$} & \rotatebox{90}{DINO$\uparrow$} & \rotatebox{90}{CLIP$\uparrow$} & \rotatebox{90}{T5$\uparrow$} & \rotatebox{90}{FID-I$\downarrow$} & \rotatebox{90}{FID-C$\downarrow$} & \rotatebox{90}{DINO$\uparrow$} & \rotatebox{90}{CLIP$\uparrow$} & \rotatebox{90}{T5$\uparrow$} & \rotatebox{90}{FID-I$\downarrow$} & \rotatebox{90}{FID-C$\downarrow$} & \rotatebox{90}{DINO$\uparrow$} & \rotatebox{90}{CLIP$\uparrow$} & \rotatebox{90}{T5$\uparrow$} & \rotatebox{90}{FID-I$\downarrow$} & \rotatebox{90}{FID-C$\downarrow$} & \rotatebox{90}{DINO$\uparrow$} & \rotatebox{90}{CLIP$\uparrow$} & \rotatebox{90}{T5$\uparrow$} \\
\midrule
Baseline & \ding{55} & \ding{55} & \ding{55} & \textbf{32.9} & \textbf{0.149} & \ul{0.375} & \ul{0.298} & \ul{0.651} & 37.0 & 0.163 & 0.379 & 0.294 & 0.648 & 38.5 & 0.166 & 0.383 & 0.291 & 0.646 & 52.5 & 0.212 & 0.381 & 0.278 & 0.642 & 58.0 & 0.229 & 0.375 & 0.274 & 0.641 \\
+ PCM & \ding{51} & \ding{55} & \ding{55} & \ul{36.5} & \ul{0.152} & \textbf{0.400} & \textbf{0.306} & \textbf{0.657} & \textbf{35.0} & \textbf{0.148} & \textbf{0.407} & \textbf{0.306} & \ul{0.656} & \textbf{34.3} & \ul{0.147} & \textbf{0.413} & \textbf{0.305} & \ul{0.656} & \textbf{34.5} & \ul{0.146} & \textbf{0.422} & \ul{0.303} & \ul{0.654} & \textbf{34.7} & \ul{0.147} & \textbf{0.421} & \ul{0.303} & \ul{0.653} \\
+ LS-KVI & \ding{51} & \ding{51} & \ding{55} & \ul{36.5} & \ul{0.152} & \textbf{0.400} & \textbf{0.306} & \textbf{0.657} & 37.0 & 0.163 & \ul{0.402} & \ul{0.304} & \ul{0.656} & \ul{36.4} & 0.163 & 0.408 & \ul{0.303} & 0.655 & 36.2 & 0.159 & \ul{0.419} & 0.302 & 0.653 & \ul{36.5} & 0.160 & \ul{0.418} & 0.301 & \ul{0.653} \\
\cellcolor{oursblue}Full (+ BAC) & \ding{51} & \ding{51} & \ding{51} & \cellcolor{oursblue}\ul{36.5} & \cellcolor{oursblue}\ul{0.152} & \cellcolor{oursblue}\textbf{0.400} & \cellcolor{oursblue}\textbf{0.306} & \cellcolor{oursblue}\textbf{0.657} & \cellcolor{oursblue}\ul{35.8} & \cellcolor{oursblue}\ul{0.158} & \cellcolor{oursblue}\ul{0.402} & \cellcolor{oursblue}\ul{0.304} & \cellcolor{oursblue}\textbf{0.658} & \cellcolor{oursblue}\ul{36.4} & \cellcolor{oursblue}\textbf{0.144} & \cellcolor{oursblue}\ul{0.409} & \cellcolor{oursblue}\textbf{0.305} & \cellcolor{oursblue}\textbf{0.657} & \cellcolor{oursblue}\ul{36.1} & \cellcolor{oursblue}\textbf{0.142} & \cellcolor{oursblue}0.415 & \cellcolor{oursblue}\textbf{0.304} & \cellcolor{oursblue}\textbf{0.656} & \cellcolor{oursblue}36.7 & \cellcolor{oursblue}\textbf{0.141} & \cellcolor{oursblue}0.415 & \cellcolor{oursblue}\textbf{0.304} & \cellcolor{oursblue}\textbf{0.655} \\
\bottomrule
\end{tabular}}
\begin{tablenotes}
\scriptsize
\item \hspace{-0.4cm}\textbf{Bold}: best, \ul{underline}: second best.
\end{tablenotes}
\end{threeparttable}
\end{table*}

\subsection{Interpreting the Ablation Progression}
\label{sec:supp_ablation_interp} 
FID-I and FID-C measure distributional quality in Inception and CLIP image-feature spaces, respectively; DINOv2 measures paired visual-semantic fidelity, while CLIPScore and T5-Cos assess text alignment. We therefore interpret temporal and fidelity metrics jointly, complemented by qualitative and operation-level evaluations. From 30\% completion onward on FS-COCO, the full system (PCM + LS-KVI + BAC) obtains the strongest FID-C and remains at or above the other ablation variants on CLIPScore and T5-Cos across the reported stages (\cref{tab:supp_ablation,tab:ablation}). Its FID-C remains low and stable across 30--100\% completion (0.144, 0.141, 0.142, 0.141, 0.143), while CLIPScore and T5-Cos remain competitive throughout. Together, the full system maintains strong FID-I while improving FID-C and textual alignment across later completion stages; the qualitative BAC analysis separately examines difficult sketch--prompt configurations.
\subsection{Inference and Reproducibility Details}
\label{sec:supp_reproducibility}

\paragraph{Inference configuration.}
All ProgressNet experiments use FLUX.1-schnell with Union ControlNet at
$1024\times1024$ resolution. Unless otherwise stated, the structural and
memory-conditioning scales are fixed to $\lambda_s=0.5$ and
$\lambda_m=0.65$, respectively. Live progressive generation uses two inference steps in bfloat16 on a single NVIDIA H200 GPU.

\paragraph{Sketch preprocessing.}
The line-based sketch condition is converted to a binary structural map
before being resized to the $1024\times1024$ ControlNet input resolution.
We convert the sketch to grayscale, threshold it at intensity $200$,
invert the binary map so that drawn strokes form the active condition,
and resize using nearest-neighbour interpolation. 
\begin{figure*}[!htb]
  \centering
  \includegraphics[width=\linewidth]{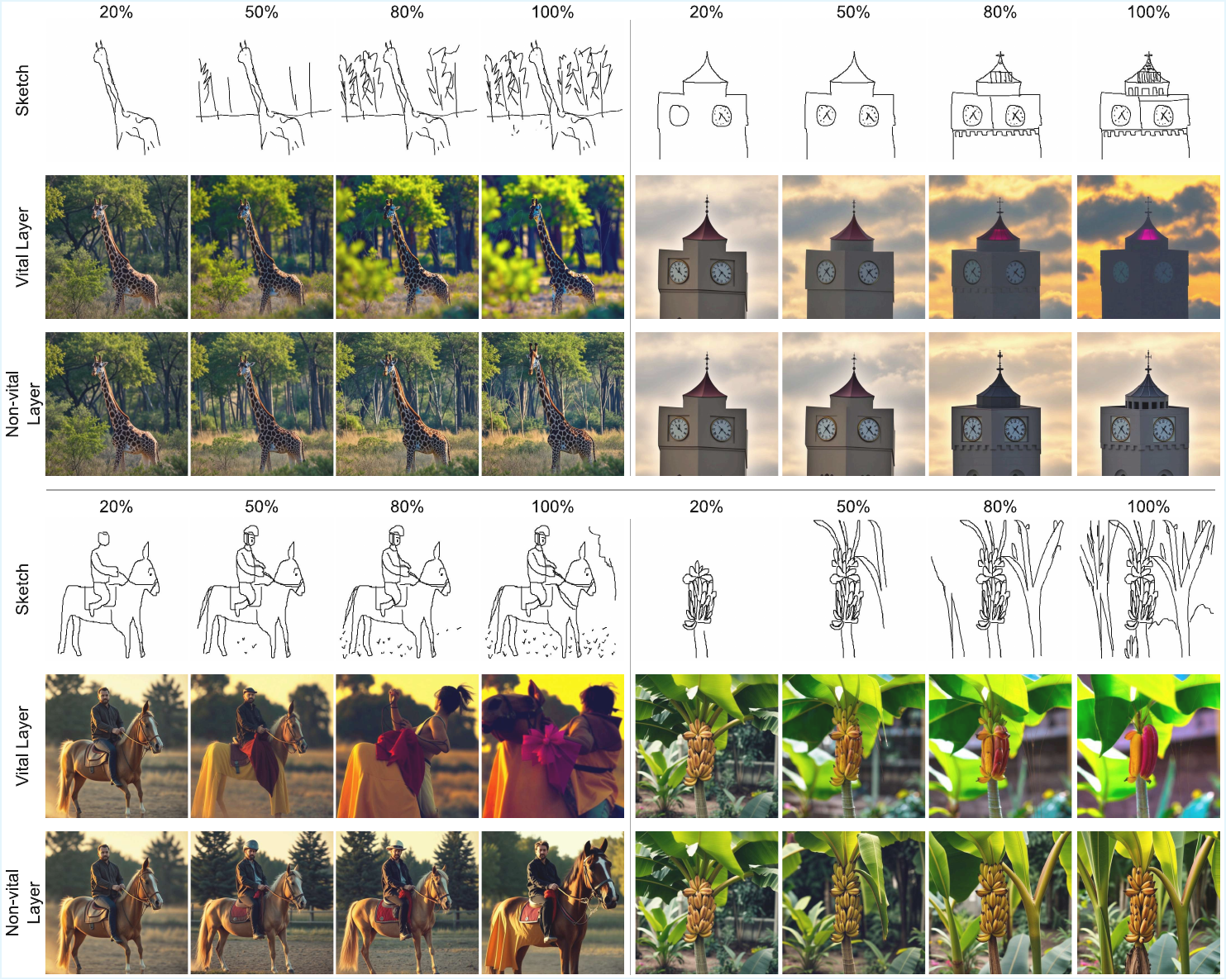}
  \caption{Extended qualitative comparison of vital-layer vs.\ non-vital-layer K/V injection across four FS-COCO samples at progressive sketch completions ($20\%$, $50\%$, $80\%$, $100\%$). Vital-layer (\citep{avrahami2025stableflow}) injection progressively corrupts appearance, shifting illumination (clock tower darkens), altering identity (horse rider changes clothing and pose), and inducing color drift (banana flower saturates), while non-vital-layer ($\{5,6,7,10,13,14\}$) injection preserves consistent appearance, lighting, and identity throughout the progression. This extends the single-sample comparison in \cref{fig:lskvi_exp} of the main paper.}
  \label{fig:supp_lskvi_qual}
  \vspace{-20pt}
\end{figure*}

% ---------------------- TEXT ----------------------
% \noindent\textit{Layer selection.}\quad
% Table~\ref{tab:sensitivity} (LS-KVI rows) compares injection at the vital layers $\{0,1,2,18\}$ identified by StableFlow~\citep{avrahami2025stableflow} against our non-vital set $\{5,6,11,10,13,14\}$. While vital-layer injection yields marginally higher CLIPScore ($+0.000$--$0.007$) by directly reinforcing the semantic pathway, it incurs monotonic FID-I degradation as completion increases ($170.7{\to}187.3$, $\Delta{=}16.6$), compared with a stable trajectory for our default ($173.4{\to}177.4$, $\Delta{=}4.0$)---a $4{\times}$ larger quality drift. DINOv2 similarly declines under vital injection ($0.406{\to}0.396$) while remaining flat for non-vital ($0.405{\to}0.402$). 
\section{Component Implementation and LS-KVI Analysis}
\label{sec:supp_components}
\subsection{Previous-Concept Memory implementation}
\label{sec:supp_pcm_impl}
PCM derives its commitment mask $C_t$ from the processed current sketch. We combine a dilated stroke-proximity support with flood-filled enclosed regions; the resulting mask marks content that should remain committed under subsequent interaction. At the two extremes, $C_t{=}1$ denotes fully committed content retained from $I_{t-1}$, whereas $C_t{=}0$ denotes fully relaxed regions during a prompt update; intermediate values provide a soft blend between the previous output and the neutral prior. For stroke addition, the previous output is used unchanged as memory. For stroke removal, evidence inside the erased region is withdrawn by neutral white filling before tile conditioning. For prompt updates, committed content is preserved while uncommitted regions are replaced by the neutral gray prior, as defined in \cref{eq:pcm_memory} of the main paper. The same mask construction is used for all datasets and all interaction sequences.
\paragraph{Commitment-mask construction.}
The soft commitment mask combines two complementary supports. For
stroke-proximity support, pixels with grayscale intensity below $200$
are treated as sketch pixels, dilated by $24$ pixels using an elliptical
kernel, and Gaussian-smoothed with a $41\times41$ kernel. For enclosed
regions, we first morphologically close sketch gaps using a
$17\times17$ elliptical kernel, corresponding to an $8$-pixel closing
radius. We then flood-fill the exterior from the image boundary and
retain the non-exterior region as the enclosed support, followed by
Gaussian smoothing with a $21\times21$ kernel. The final commitment
mask is the pixelwise maximum of the proximity and enclosed-region
supports and is clipped to $[0,1]$. These parameters are held fixed
throughout all reported experiments.
\subsection{Choice of vital and non-vital layers}
\label{sec:supp_layer_choice}
The mechanism is made visible in Fig.~\ref{fig:supp_lskvi_qual}, which extends the single-sample comparison of \cref{fig:lskvi_exp} across four diverse FS-COCO scenes. In each case, vital-layer injection progressively overrides appearance attributes that should remain stable: the clock tower's warm sunset illumination darkens into an unnatural palette; the horse rider's clothing, skin tone, and pose drift across completions; and the banana flower saturates in color as strokes accumulate. 
% These artifacts are consistent with the quantitative degradation of FID-I in Tab.~\ref{tab:sensitivity}: vital layers encode the generation pathway that determines global appearance, and once their keys and values are replaced with cached versions, each new stroke forces the model to reconcile conflicting identity signals, producing the progressive corruption visible in the middle rows of Fig.~\ref{fig:supp_lskvi_qual}.

In contrast, non-vital-layer injection (bottom rows) maintains consistent lighting, identity, and color across the four examples. Empirically, injecting at these lower-vitality layers preserves appearance more effectively while allowing structural adaptation. The giraffe retains its pose and forest backdrop; the clock tower preserves warm tones and adapts precise details correctly; the horse rider maintains consistent attire and posture; and the banana flower evolves structurally without color shift. This is precisely the property exploited by LS-KVI under progression: temporal smoothing at layers whose individual absence leaves generation quality intact (\cref{sec:lskvi}, main paper), enabling the concept to structurally evolve with new strokes while preserving what the image \emph{was}. We therefore inject at non-vital layers throughout. Under the FLUX vitality analysis, the six lowest-vitality layers selected for injection are $\{5,6,7,10,13,14\}$, with $\alpha_{\mathrm{kv}}{=}0.3$ in all reported experiments. The cache is refreshed after each generation and cleared on stroke removal so that correspondence pathways associated with erased content are not reintroduced.
\paragraph{K/V tensor and cache implementation.}
LS-KVI attaches forward hooks to the key and value
projections of joint attention at the selected transformer blocks. The
full projected key and value tensors are retained rather than spatially
or channel-wise subsampled. For a selected layer $\ell$, the current
turn uses
\[
\mathbf{K}^{\prime t}_{\ell}=\mathbf{K}^{t}_{\ell}+\alpha_{\mathrm{kv}}\mathbf{K}^{t-1}_{\ell},
\qquad
\mathbf{V}^{\prime t}_{\ell}=\mathbf{V}^{t}_{\ell}+\alpha_{\mathrm{kv}}\mathbf{V}^{t-1}_{\ell},
\]
with $\alpha_{\mathrm{kv}}=0.3$. During generation, the current
projection outputs are captured for the next turn; after the generation
finishes, the captured tensors are committed as the new reference cache.
The cache is indexed by transformer layer and is refreshed after every
completed turn. On stroke removal, the cached K/V state is cleared before
the subsequent generation so that features associated with erased
content cannot be re-injected.

\paragraph{Layer-selection protocol.}
The injection subset is selected once using the residual-bypass vitality
analysis described in \cref{sec:lskvi}: each candidate transformer layer
is bypassed individually over progressive sketch stages and ranked by
the resulting output deviation. LS-KVI uses the lowest-vitality layers
and keeps this subset and $\alpha_{\mathrm{kv}}=0.3$ fixed across the
reported datasets rather than re-selecting layers per dataset.
\begin{table}[!htb]
\centering
\caption{Erasure-specific LS-KVI ablation on 100 FS-COCO sequences.}
\label{tab:supp_erasure}
\vspace{-4pt}
\setlength{\tabcolsep}{5pt}
\renewcommand{\arraystretch}{1.08}
\resizebox{\linewidth}{!}{%
\begin{tabular}{lccccc}
\toprule
LS-KVI state
& Return LPIPS $\downarrow$
& Recovery $\uparrow$
& DINOv2 $\uparrow$
& Erased-region L1 $\downarrow$
& Outside-change L1 $\downarrow$ \\
\midrule
Retain previous K/V
& $0.295$
& $0.019$
& $0.854$
& $0.121$
& $\mathbf{0.096}$ \\

Invalidate previous K/V
& $\mathbf{0.204}$
& $\mathbf{0.303}$
& $\mathbf{0.885}$
& $\mathbf{0.085}$
& $0.123$ \\
\bottomrule
\end{tabular}
}
\vspace{-4pt}
\end{table}

\paragraph{Erasure-specific LS-KVI state invalidation.}
We quantitatively evaluate stroke removal on 100 FS-COCO sequences using a controlled
$S_{50}\!\rightarrow\!S_{70}\!\rightarrow\!S_{50}$ add--erase cycle, where the final sketch is
identical to the initial $50\%$ state. All generation settings and preceding history are fixed;
we vary only whether the previous LS-KVI cache is retained or invalidated during removal.
Return LPIPS~\citep{zhang2018unreasonable} and DINOv2~\citep{oquab2023dinov2} measure recovery toward the original $I_{50}$ output, while
Recovery denotes the LPIPS improvement relative to the preceding $I_{50}\!\rightarrow\!I_{70}$
change. We additionally report error inside the erased region and appearance change outside it.
Invalidating stale LS-KVI state reduces return LPIPS by $30.9\%$ and improves recovery in
$93/100$ sequences (paired Wilcoxon $p=7.1{\times}10^{-15}$). It also reduces erased-region
error in $91/100$ cases, showing that stale temporal correspondences can resist destructive
updates.

\section{Empirical Validation of Banded Adaptive Control}
\label{sec:supp_bac}
\vspace{-12pt}
We examine how control responses vary with network depth,
solver step, and progressive stage, and whether related
patterns persist across backbones and sketch domains.
These diagnostics motivate 
\begin{wrapfigure}{l}{0.4\textwidth}
  % \vspace{-10pt}
  \centering
  \hspace*{-4pt}
  \includegraphics[width=\linewidth]{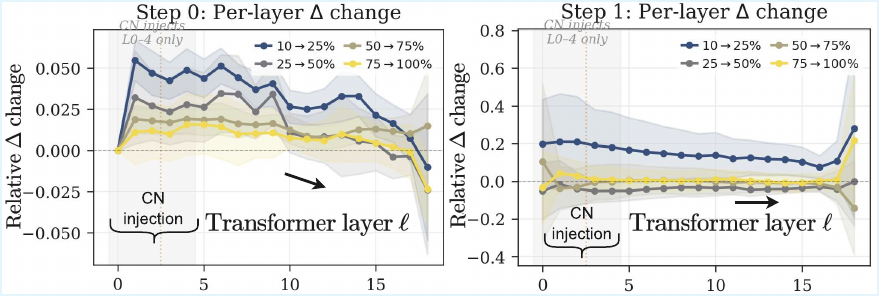}
  \captionsetup{font=small}
  \vspace{-18pt}
  \caption{Layerwise relative change in the measured control-induced perturbation across progressive sketch transitions at solver steps 0 and 1. The response varies systematically with backbone depth and solver step.}
  \label{fig:supp_perlayer}
  \vspace{-12pt}
\end{wrapfigure}
the banded, step-specific
design; the component ablations evaluate its effects
on generated outputs.
% ---------------------- FIGURE A (wrapfig, left) ----------------------
%%%%%%%%%%%%%%%% ----------------- FINAL
% \paragraph{channel-wise JSD definition.} Let $\mathbf{c}\in\mathbb{R}^{C\times H\times W}$ denote a
% ControlNet feature tensor. For channel $q$, we convert the spatial
% activation map $\mathbf{c}_q$ into a normalized spatial distribution
% $\mathbf{p}_q=\Phi(\mathbf{c}_q)$ using the same per-channel normalization in all experiments. Given consecutive features
% $\mathbf{c}^{t-1}$ and $\mathbf{c}^{t}$, we define

% \[
% \operatorname{JSD}_{\mathrm{cw}}
% (\mathbf{c}^{t-1},\mathbf{c}^{t})
% =
% \frac{1}{C}
% \sum_{q=1}^{C}
% \operatorname{JSD}
% \left(
% \Phi(\mathbf{c}^{t-1}_q),
% \Phi(\mathbf{c}^{t}_q)
% \right).
% \]
%%%%%%%%%%%%%%%% ----------------- FINAL
\textbf{Implemented BAC comparison.}
For each ControlNet block $k$ and solver step $s$, BAC compares the cached
memory-conditioning features from the two preceding turns. We compute
channel-wise Jensen--Shannon divergence (JSD) between their activation
distributions and average the resulting divergences within the early and late
ControlNet bands. Following \cref{eq:bac_cached} of the main paper, these
band-wise values are clipped to $[f_{\min},f_{\max}]$ to obtain the four
step- and band-specific BAC correction factors. 
\subsection{Layer- and Step-Dependent Control Responses}
\label{sec:supp_propagation}

\cref{fig:supp_perlayer} reports layerwise changes in the measured
control-induced perturbation across progressive stages.
The profiles differ across backbone depth and solver
step. In standard Diffusers FLUX, ControlNet residual
samples are distributed across backbone blocks rather
than injected only into blocks $\ell_{0\text{-}4}$.
The resulting profiles provide evidence of depth- and
step-dependent control responses across progressive stages.

For FLUX's five ControlNet blocks, BAC uses $\mathcal{K}_{\mathrm{early}}{=}\{0,1,2\}$ and $\mathcal{K}_{\mathrm{late}}{=}\{3,4\}$, following the early/late partition used in the perturbation analysis.
% ---------------------- FIGURE B (wrapfig, left) ----------------------
% \begin{wrapfigure}{r}{0.4\textwidth}
%   \centering
%   \hspace*{-4pt}
%   \includegraphics[width=\linewidth]{images/supp/fig2_cropped.pdf}
%   % \captionsetup{font=small}
%   \vspace{-18pt}
%   \caption{Mean ControlNet JSD$_{\mathrm{cw}}$ vs.\ mean transformer $|\Delta|$ change. Step~0: $r{=}0.18$, $p{=}0.068$. Step~1: $r{=}0.28$, $p{=}0.006$. At the later solver step, ControlNet redistribution is significantly associated with downstream perturbation.}
%   \label{fig:supp_cn_tf}
%   % \vspace{-4pt}
% \end{wrapfigure}
The implemented BAC factor follows the retrospective cached-feature comparison defined above. $\operatorname{JSD}_{\mathrm{cw}}$ is computed independently at each ControlNet block and solver step, after which block-wise divergences are aggregated within the early and late bands to obtain the correction factors. No additional forward pass is required during generation.

\begin{figure*}[!t]
  \centering
  \includegraphics[width=\linewidth]{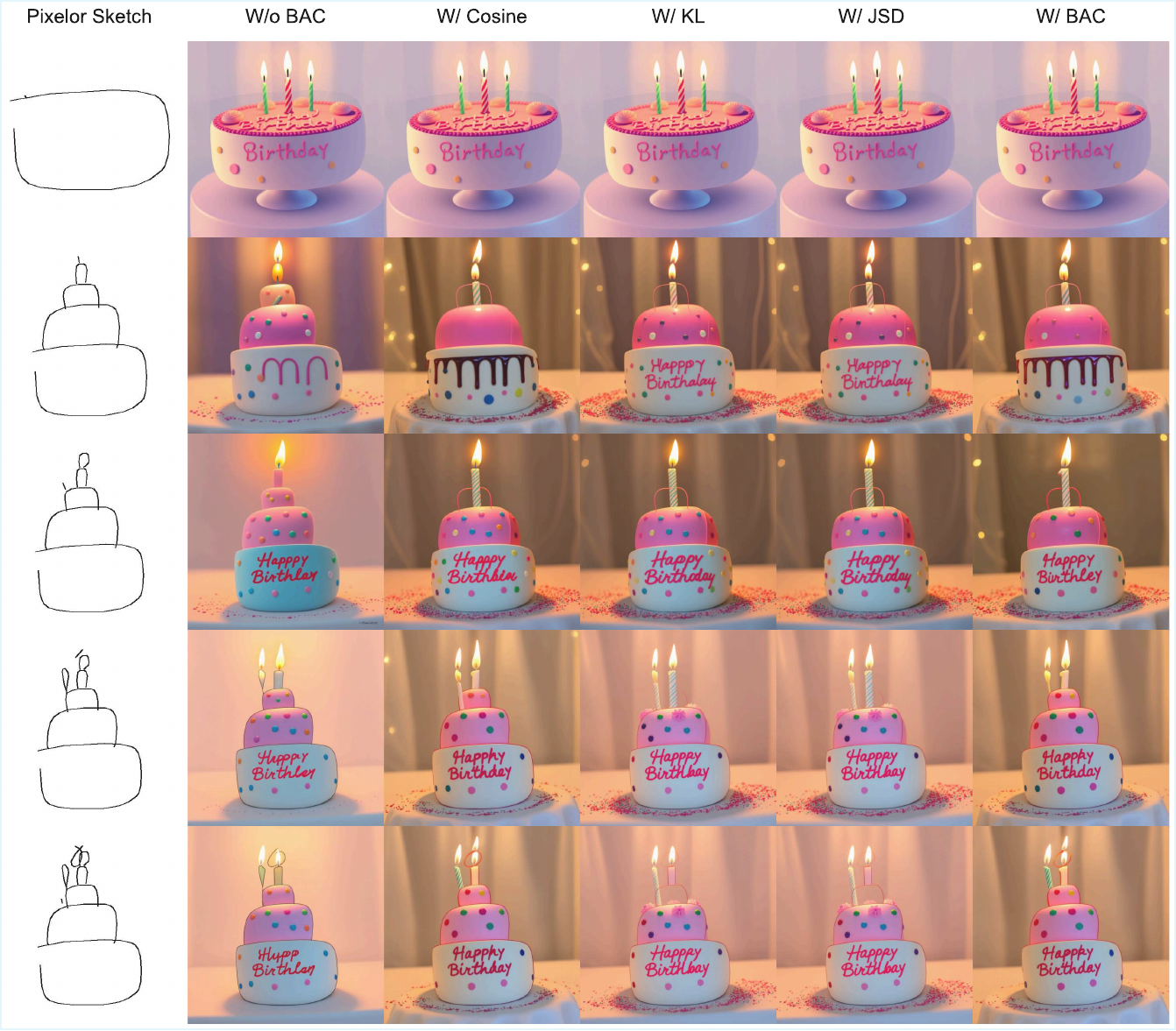}
  \caption{Comparison of BAC correction-factor variants on an out-of-distribution sketch from Pixelor~\citep{bhunia2020pixelor}. We evaluate cosine similarity, KL divergence, Jensen--Shannon divergence (JSD), and channel-wise JSD. The first column (W/o BAC) applies no correction factor. KL and JSD introduce visible artifacts at full completion, while disabling correction altogether retains a solid background with dark sketching artifacts. Compared to cosine similarity, our channel-wise JSD comparison produces smoother
modulation without imprinting strong sketch-edge details into the generated image.}
  \label{fig:supp_bac}
  \vspace{-12pt}
\end{figure*}

\paragraph{BAC Correction-Factor Variants.}
We compare four divergence measures for computing the BAC 
correction factor: cosine similarity, KL divergence, Jensen--Shannon divergence (JSD), and channel-wise JSD. Fig.~\ref{fig:supp_bac} presents qualitative results on an out-of-distribution sketch from a stroke-based drawing agent, Pixelor~\citep{bhunia2020pixelor}. Without any correction (W/o BAC), the generation retains a solid background with visible sketch-edge artifacts. KL and standard JSD show stronger visible distortions at higher completion levels. Cosine similarity yields cleaner outputs but still transfers fine sketch details into the generated image. 
%%%%%--------------------- FINAL
% channel-wise JSD, which carefully preserves per-channel structure before computing divergence, produces the smoothest modulation, suppressing sketch artifacts without over-constraining the generation. This motivates our default choice of channel-wise JSD for all BAC computations.
%%%%%--------------------- FINAL
\vspace{8pt}

\subsection{Cross-backbone validation on SD3.5 Large Turbo}
\label{sec:supp_sd35}
To test whether the band- and step-asymmetric control behavior motivating BAC
is specific to FLUX.1-schnell, we repeat the diagnostic on SD3.5 Large Turbo using 3,000 FS-COCO samples. SD3.5 uses an MMDiT backbone, four solver steps, and separate Canny and blur ControlNets, providing a substantially different control architecture.

\begin{table}[!htb]
\centering
\caption{Late/early JSD ratio across FLUX and SD3.5. Ratios above one indicate stronger redistribution in later ControlNet layers.}
\label{tab:supp_sd35_bands}
\vspace{-5pt}
\small
\setlength{\tabcolsep}{4pt}
\begin{tabular}{lccc}
\toprule
Transition & FLUX (5 layers) & SD3.5 Canny (19) & SD3.5 Blur (19) \\
\midrule
10$\to$20\%  & 3.2$\times$ & 3.3$\times$ & 2.7$\times$ \\
20$\to$30\%  & 2.0$\times$ & 2.0$\times$ & 2.7$\times$ \\
30$\to$50\%  & 2.2$\times$ & 2.2$\times$ & 2.3$\times$ \\
50$\to$70\%  & 2.3$\times$ & 2.0$\times$ & 2.4$\times$ \\
80$\to$100\% & 2.2$\times$ & 1.8$\times$ & 2.6$\times$ \\
\bottomrule
\end{tabular}
\end{table}
\begin{table}[!htb]
\centering
\caption{Step asymmetry at the $10{\to}20\%$ transition. SD3.5 exhibits progressive amplification across its four solver steps.}
\label{tab:supp_sd35_steps}
\vspace{-5pt}
\small
\setlength{\tabcolsep}{4pt}
\begin{tabular}{lcccccc}
\toprule
Control branch & Step 0 & Step 1 & Step 2 & Step 3 & Last/First & Last higher \\
\midrule
SD3.5 Canny & 0.005 & 0.007 & 0.009 & 0.021 & 3.97$\times$ & 99.3\% \\
SD3.5 Blur  & 0.118 & 0.147 & 0.168 & 0.212 & 1.79$\times$ & 100\% \\
FLUX (2-step) & 0.134 & 0.206 & -- & -- & 1.54$\times$ & 97.0\% \\
\bottomrule
\end{tabular}
\end{table}
Late-layer redistribution remains stronger on SD3.5 despite the different architecture and the use of separate control branches. The SD3.5 comparison was recorded at the five transitions shown in \cref{tab:supp_sd35_bands}; the additional $70{\to}80\%$ transition is included in the FLUX cross-dataset analysis below. The inferred band boundary is also modality-dependent: the Canny branch separates at layer 11/19 ($\sim58\%$ depth), close to FLUX's 3/5 ($60\%$) split, whereas the blur-memory branch separates earlier at layer 5/19 ($\sim26\%$). This distinction cannot be observed in FLUX's Union ControlNet because structure and memory share one control stack.
\subsection{Cross-dataset BAC validation}
\label{sec:supp_crossdataset}
We next test whether the same redistribution pattern depends on FS-COCO. The BAC diagnostic is run on FLUX.1-schnell over four sketch domains totaling 52,368 samples: FS-COCO~\citep{chowdhury2022fs} (10,000), FrISS~\citep{kutuk2025class} (1,010), Photo-Sketching~\citep{li2019photo} (5,000), and Sketchy~\citep{sangkloy2016sketchy} (36,358). These datasets span freehand scenes, abstract text-associated scenes, edge-aligned contour tracings, and single-object freehand drawings.

\begin{table*}[!t]
\centering
\caption{Late/early JSD ratio at solver step 0 across four sketch domains. The same band asymmetry persists at every evaluated transition.}
\label{tab:supp_bac_crossdataset_bands}
\vspace{-5pt}
\small
\setlength{\tabcolsep}{5pt}
\begin{tabular}{lcccccc}
\toprule
Dataset & 10$\to$20 & 20$\to$30 & 30$\to$50 & 50$\to$70 & 70$\to$80 & 80$\to$100 \\
\midrule
FS-COCO          & 3.2$\times$ & 2.0$\times$ & 2.2$\times$ & 2.3$\times$ & 2.3$\times$ & 2.2$\times$ \\
FrISS            & 3.4$\times$ & 2.0$\times$ & 2.1$\times$ & 2.1$\times$ & 2.1$\times$ & 2.1$\times$ \\
Photo-Sketching  & 3.2$\times$ & 2.0$\times$ & 2.1$\times$ & 2.1$\times$ & 2.1$\times$ & 2.2$\times$ \\
Sketchy          & 3.3$\times$ & 1.8$\times$ & 2.2$\times$ & 2.0$\times$ & 1.9$\times$ & 2.0$\times$ \\
\bottomrule
\end{tabular}
\end{table*}

\begin{table}[!t]
\centering
\caption{Step-1/step-0 JSD amplification at the $10{\to}20\%$ transition. The existence of step asymmetry is stable across domains, while its magnitude remains content-dependent.}
\label{tab:supp_bac_crossdataset_steps}
\vspace{-5pt}
\small
\setlength{\tabcolsep}{4pt}
\begin{tabular}{lcc}
\toprule
Dataset & Step 1 / Step 0 & Samples with step 1 higher \\
\midrule
FS-COCO         & 1.54$\times$ & 97.0\% \\
FrISS           & 1.59$\times$ & 97.7\% \\
Photo-Sketching & 1.44$\times$ & 93.8\% \\
Sketchy         & 2.15$\times$ & 96.3\% \\
\bottomrule
\end{tabular}
\end{table}

\begin{table*}[!t]
\centering
\caption{Transition ranking by mean JSD across datasets (rank 1 indicates the largest redistribution). The ordering is preserved across all four domains.}
\label{tab:supp_bac_transition_rank}
\vspace{-5pt}
\small
\setlength{\tabcolsep}{4pt}
\begin{tabular}{cllll}
\toprule
Rank & FS-COCO & FrISS & Photo-Sketching & Sketchy \\
\midrule
1 & 20$\to$30 (.362) & 20$\to$30 (.365) & 20$\to$30 (.371) & 20$\to$30 (.390) \\
2 & 80$\to$100 (.247) & 80$\to$100 (.241) & 80$\to$100 (.241) & 80$\to$100 (.256) \\
3 & 30$\to$50 (.238) & 30$\to$50 (.228) & 30$\to$50 (.232) & 30$\to$50 (.256) \\
4 & 50$\to$70 (.225) & 50$\to$70 (.217) & 50$\to$70 (.214) & 50$\to$70 (.234) \\
5 & 70$\to$80 (.193) & 70$\to$80 (.203) & 70$\to$80 (.198) & 70$\to$80 (.212) \\
6 & 10$\to$20 (.134) & 10$\to$20 (.118) & 10$\to$20 (.128) & 10$\to$20 (.104) \\
\bottomrule
\end{tabular}
\end{table*}

Together, the SD3.5 and cross-dataset analyses show that the qualitative BAC signature used in the main paper, stronger late-band redistribution and solver-step asymmetry---persists beyond one backbone and one sketch domain.
% NOTE: the pre-rebuttal quantitative BAC-metric table is not inserted here because its
% Photo-Sketching FID scale is inconsistent with the current main-paper Table 1 protocol.
% Reconcile that evaluation before adding the Cosine/KL/JSD/JSD-CW quantitative table.

\vspace{-8pt}
\section{User Study Details}
\label{sec:supp_user}
\vspace{-6pt}
We recruited participants
% 42 participants (16 with formal artistic training and 26 without)
from open communities through an online Gradio interface. Both studies were conducted double-blind: method identities were hidden, presentation order was randomized per trial, and participants were not told which system produced an output.

\paragraph{Study 1: overall generation quality.}
Participants evaluated six methods (Ours, Flux+ControlNet~\citep{flux2024}, SDXS~\citep{song2024sdxs}, Sketch-a-Sketch~\citep{sarukkai2024block}, StableFlow~\citep{avrahami2025stableflow}, and Conditional Balance~\citep{cohen2025conditional}) across 10 diverse samples. For each sample, participants (i) ranked the six methods by overall visual preference, (ii) rated temporal coherency using a continuous 0--1 slider while viewing progressive-generation videos, and (iii) rated generation quality on a five-point Likert scale from Very Poor to Excellent. These correspond to the \textit{Pref.}, \textit{Coher.}, and \textit{Qual.} columns in \cref{tab:user_study}. The reported values aggregate responses across participants and evaluated samples and are presented as normalized criterion-wise scores, with higher values indicating stronger user preference.

\paragraph{Study 2: operation-level capabilities.}
Participants first selected the better output for 12 erasure samples and three editing-pipeline samples, evaluating stroke-removal and sketch-modification fidelity. They then viewed side-by-side videos of complete progressive sessions and judged prompt-guided progression, fine-grained drawing quality, fine-grained erasure quality, and content preservation across edits. These map to the \textit{Erase}, \textit{Edit}, \textit{Prog.}, \textit{F-Draw}, \textit{F-Erase}, and \textit{Preserv.} columns in \cref{tab:user_study}. As in Study~1, responses are aggregated across participants and evaluated samples and reported as normalized criterion-wise scores, with higher values indicating stronger outcomes. Scores are normalized independently within each criterion across methods.

\vspace{-10pt}
\section{Extended Related Work}
\label{sec:supp_related}

\paragraph{Controllable Image Generation.}
ControlNet~\citep{zhang2023controlnet} established spatial conditioning through auxiliary branches attached to frozen diffusion backbones, and subsequent works expanded controllability through unified multi-condition models~\citep{zhao2023unicontrolnet,qin2023unicontrol}, lightweight conditioning adapters~\citep{mou2024t2iadapter,ye2023ipadapter}, feedback-based control architectures~\citep{li2024controlnetplusplus,zavadski2024controlnetxs}, grounded and regional generation~\citep{li2023gligen,xie2023boxdiff,bar2023multidiffusion}, and training-free feature manipulation~\citep{mo2024freecontrol,tumanyan2023plug}. Sketch-based conditioning has similarly progressed through dedicated
sketch-to-image and interactive systems~\citep{sarukkai2024block,song2024sdxs} toward increasingly general conditional generators. These developments largely assume a condition that is already sufficiently specified before generation begins.

The transition from UNet diffusion to DiT and rectified-flow architectures~\citep{peebles2023scalable,esser2024sd3,flux2024} has further changed how spatial conditions are injected. Ctrl-Adapter~\citep{lin2025ctrladapter} bridges pretrained ControlNets to newer diffusion backbones, OminiControl~\citep{tan2025ominicontrol} introduces lightweight universal control for FLUX, and SemanticDraw~\citep{lee2025semanticdraw} supports real-time multi-region interaction. ControlNet-XS~\citep{zavadski2024controlnetxs} and ControlNet++~\citep{li2024controlnetplusplus} likewise revisit the interaction between the conditioning pathway and the generator. While these methods considerably expand \emph{what} can be controlled, their conditioning remains predominantly generation-centric rather than progression-centric: the model receives the current condition, but does not explicitly reason about what should remain stable from the preceding interaction turn. ProgressNet instead treats the previous concept as part of the state of generation and explicitly distinguishes stroke addition, removal, and prompt reinterpretation.

\paragraph{Real-Time and Iterative Image Editing.}
Diffusion editing has been extensively studied through inversion and attention manipulation~\citep{hertz2022prompttoprompt,meng2022sdedit,couairon2022diffedit,kawar2023imagic,mokady2023nulltext}, inpainting~\citep{avrahami2023blended,Lugmayr_2022_CVPR,suvorov2022lama}, instruction-based editing~\citep{brooks2023instructpix2pix}, and personalization~\citep{hu2022lora,ruiz2023dreambooth,gal2023textualinversion,kumari2023customdiffusion}. Such methods enable semantic or regional modification but generally solve each edit as a fresh optimization, inversion, or generation problem rather than maintaining a persistent state across a long interaction sequence.

Few-step and distilled generation substantially reduce this cost. Latent Consistency Models~\citep{luo2023latent}, SDXS~\citep{song2024sdxs}, NitroFusion~\citep{Chen_2025_CVPR}, SDXL-Turbo~\citep{sauer2024fast}, and StreamDiffusion~\citep{kodaira2023streamdiffusion} push inference toward interactive rates. However, low latency alone does not imply temporal consistency: regenerating from a changed sketch can alter previously established identity, appearance, geometry, or background even when the newly introduced stroke is local. Among interactive sketch systems, Block-and-Detail~\citep{sarukkai2024block} stabilizes generation through coarse-to-fine re-noising, whereas SDXS~\citep{song2024sdxs} provides one-step image-conditioned synthesis; both remain vulnerable to drift when the conditioning evolves progressively.

Recent training-free editing methods are particularly related to our temporal component. StableFlow~\citep{avrahami2025stableflow} identifies vital transformer layers and reuses internal features to preserve content during editing, while FluxSpace~\citep{dalva2025fluxspace} exposes semantic directions in rectified-flow representations and FLUX.1 Kontext~\citep{bfl2025fluxkontext} enables iterative in-context editing. Related feature-reuse mechanisms have also been explored for consistency through
cross-image or temporal attention~\citep{alaluf2024crossimage,cao2023masactrl,
qi2023fatezero}. These works demonstrate that internal attention features provide a strong substrate for consistency, but progressive sketching requires a different balance: preserving all influential layers too strongly can suppress newly drawn structure. This motivates our Layer-Selective K/V Injection, which preserves temporal information only through non-vital layers while leaving structurally important layers free to adapt.

\paragraph{Condition Misalignment and Adaptive Control.}
A separate literature studies the failure of text and visual conditions to remain semantically aligned. Text-to-image models exhibit semantic neglect, incorrect attribute binding, and word-to-concept mismatch~\citep{rassin2022seeingdouble,wang2024tasc}, with such failures analyzed through cross-attention~\citep{tang2023daam} and latent-space representations~\citep{ramos2025beyondnoise}. Structured Diffusion~\citep{feng2023trainingfree} decomposes prompts for improved compositionality, while Attend-and-Excite~\citep{chefer2023attend}, Linguistic Binding~\citep{rassin2023linguistic}, and Attention Refocusing~\citep{phung2023attentionrefocusing} intervene on attention to improve semantic correspondence.

More directly related to ProgressNet, SmartControl~\citep{liu2024smartcontrol} adapts ControlNet behavior under rough or unreliable visual conditions, while Text-Anchored Score Composition~\citep{wang2024tasc} explicitly addresses conflict between textual and spatial conditions. Conditional Balance~\citep{cohen2025conditional} controls competing conditioning signals through selective feature manipulation, and training-free alternatives such as Ctrl-X~\citep{lin2024ctrlx} and FreeControl~\citep{mo2024freecontrol} regulate structure and appearance without conventional task-specific training. Collectively, these methods show that a single fixed global control scale is insufficient when text and structural evidence disagree.

Progressive sketching makes this conflict substantially more difficult because the disagreement itself changes over time. A sparse early contour may be highly ambiguous or even contradict the prompt, whereas later strokes can progressively resolve the intended geometry. The appropriate strength of control can therefore vary across interaction turns, architectural layers, and denoising steps. Our pilot analysis further shows that ControlNet perturbations are step- and band-asymmetric and that large inter-turn feature redistribution accompanies structurally disruptive stroke updates. 
%%%%%%%%%%%% -------------------- FINAL
% ProgressNet exploits this behavior through Banded Adaptive Control (BAC), using channel-wise Jensen--Shannon divergence between consecutive ControlNet features as an online measure of inter-turn control redistribution. Unlike learned control policies~\citep{liu2024smartcontrol} or manually selected control locations~\citep{cohen2025conditional}, BAC derives its correction directly from the evolving feature distribution and requires no additional training.
%%%%%%%%%%%% -------------------- FINAL
ProgressNet exploits this behavior through Banded Adaptive Control (BAC),
deriving step- and band-specific correction factors from retrospective
comparisons of cached tile-memory ControlNet features across preceding turns. Unlike learned control policies~\citep{liu2024smartcontrol} or manually selected control
locations~\citep{cohen2025conditional}, BAC derives its modulation directly from
the evolving control representation and requires no additional training.
\section{Nano Banana Pro}
\label{sec:supp_nano}
\vspace{-8pt}
Nano Banana Pro, a commercial image-generation and editing model by Google DeepMind, is included as a qualitative reference in Fig.~8 of the main paper. For fine-grained spatial-control comparisons, the same sketch and text instruction were supplied through its public chat interface, with an explicit request to follow the sketch structure. Because the system is proprietary and does not expose the controlled inference protocol required by our completion-level evaluation, we do not include it in the quantitative tables. The comparison is therefore intended only to illustrate differences in interactive sketch adherence under the available interface, not to establish a controlled benchmark ranking.
\vspace{-8pt}
\section{Additional Qualitative Examples}
\label{sec:supp_qual}
\vspace{-8pt}
We provide additional qualitative comparisons against the strongest competing methods, with all examples drawn from FS-COCO~\citep{chowdhury2022fs}, which offers a diverse set of single-object, multi-object, and scene-level sketches paired with reference photographs. This diversity allows a thorough evaluation of temporal coherence, sketch faithfulness, and semantic fidelity across several completion levels. For readability, these extended figures focus on SDXS, FLUX+ControlNet, and StableFlow, which represent one-step image-conditioned synthesis, stateless structural control, and strong temporal feature locking, respectively; the full six-method comparison is reported quantitatively in the main paper.

\begin{figure*}[!b]
  \centering
  \includegraphics[width=\linewidth]{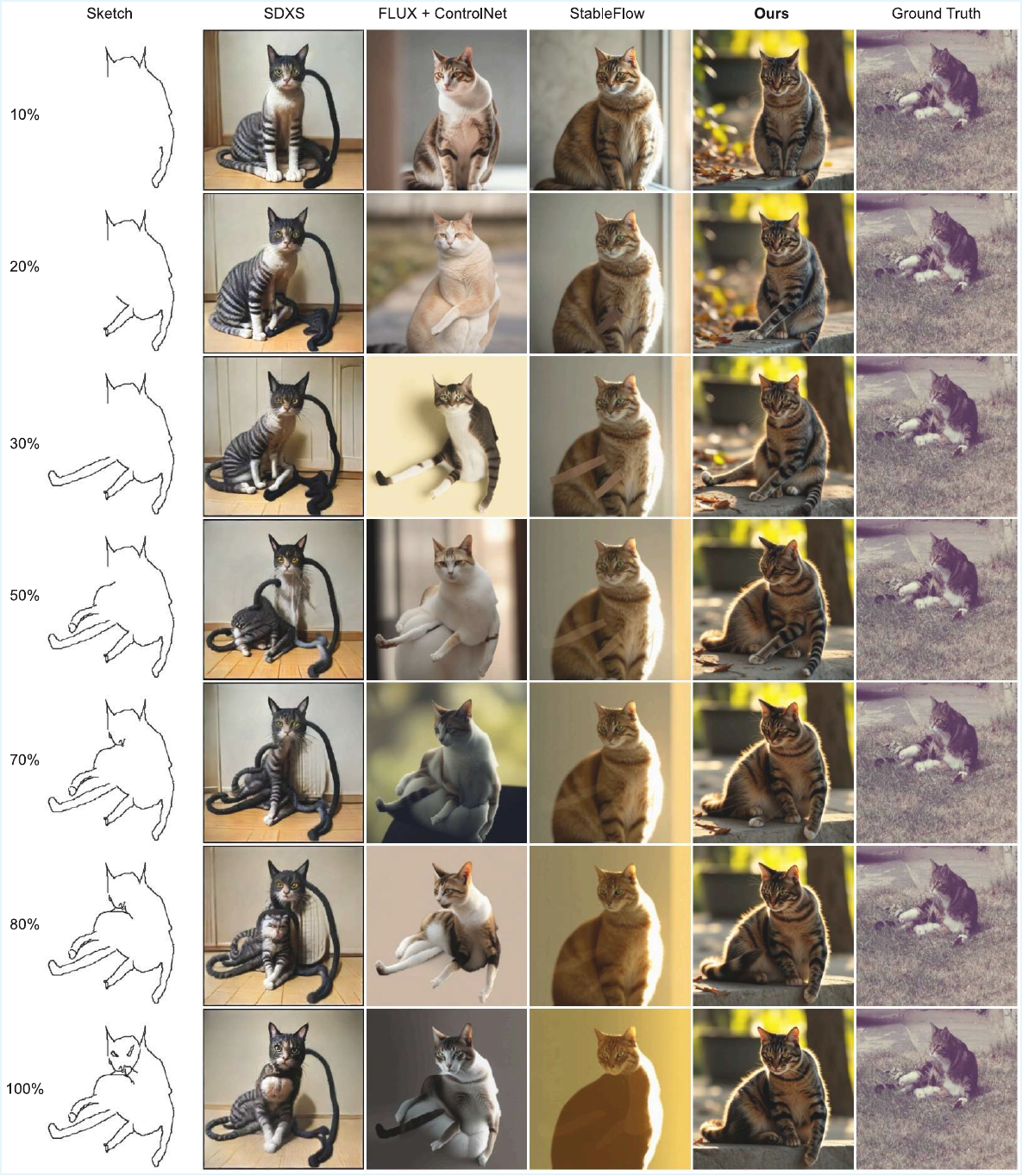}
  \caption{Progressive sketch-to-image generation, extension of \cref{fig:qualitative_main_part1}. Prompt: \texttt{In  this scene the cat sitting.} StableFlow shows stronger stroke-region amplification, while SDXS contains visible artifacts. The baseline forms a cat but changes noticeably in appearance and structure across turns. Our method remains responsive to stroke additions while preserving the sketch structure and overall appearance.}
  \label{fig:supp_qual1}
\end{figure*}

\begin{figure*}[!htb]
  \centering
  \includegraphics[width=\linewidth]{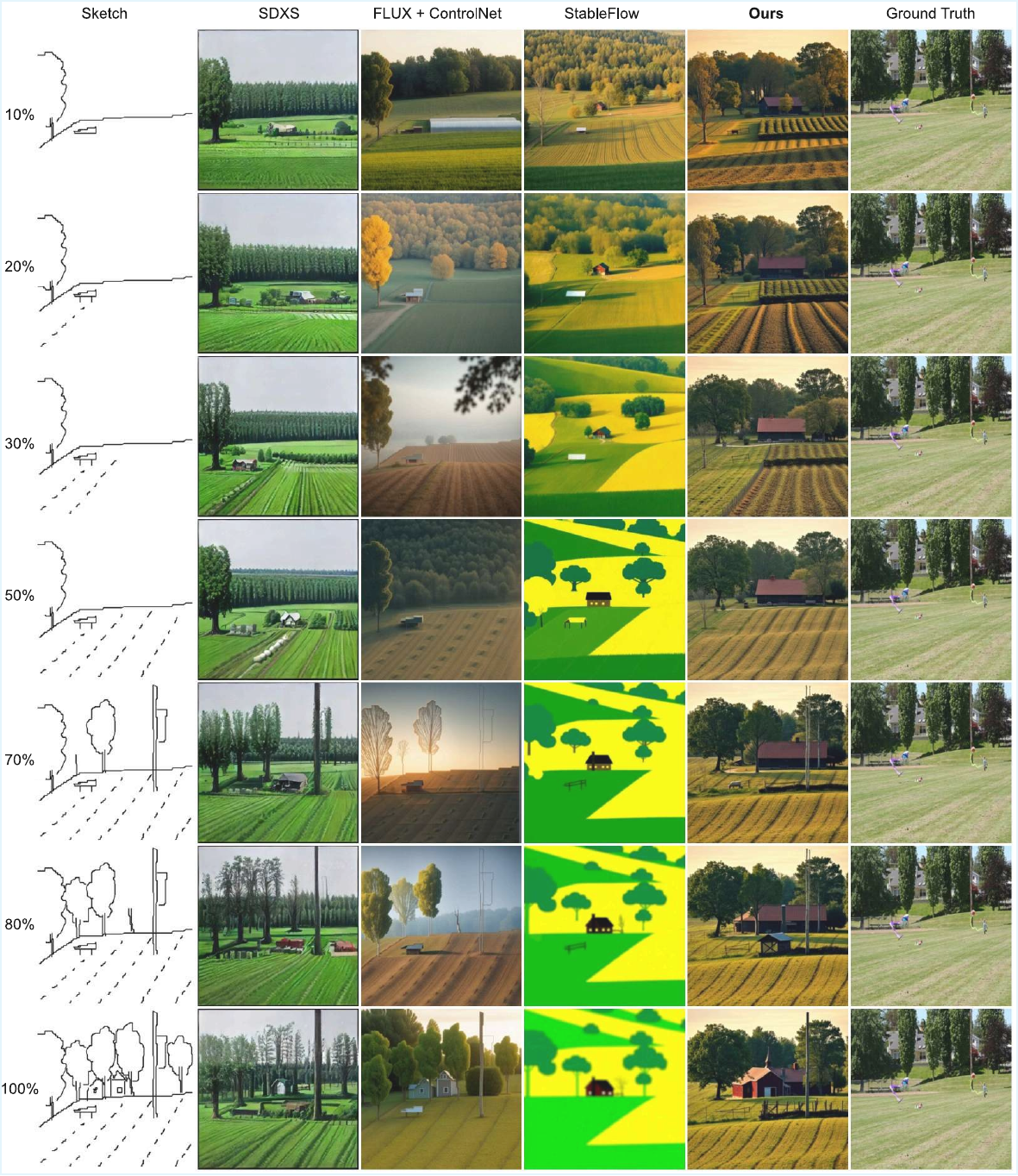}
  \caption{Progressive sketch-to-image generation. Prompt: \texttt{farm with trees in surrounding}. StableFlow shows stronger green saturation, while the baseline contains visible artifacts. SDXS captures the overall structure but retains less fine detail. Our approach adapts to sketch changes while preserving scene appearance and details.}
  \label{fig:supp_qual2}
\end{figure*}

\begin{figure*}[!htb]
  \centering
  \includegraphics[width=\linewidth]{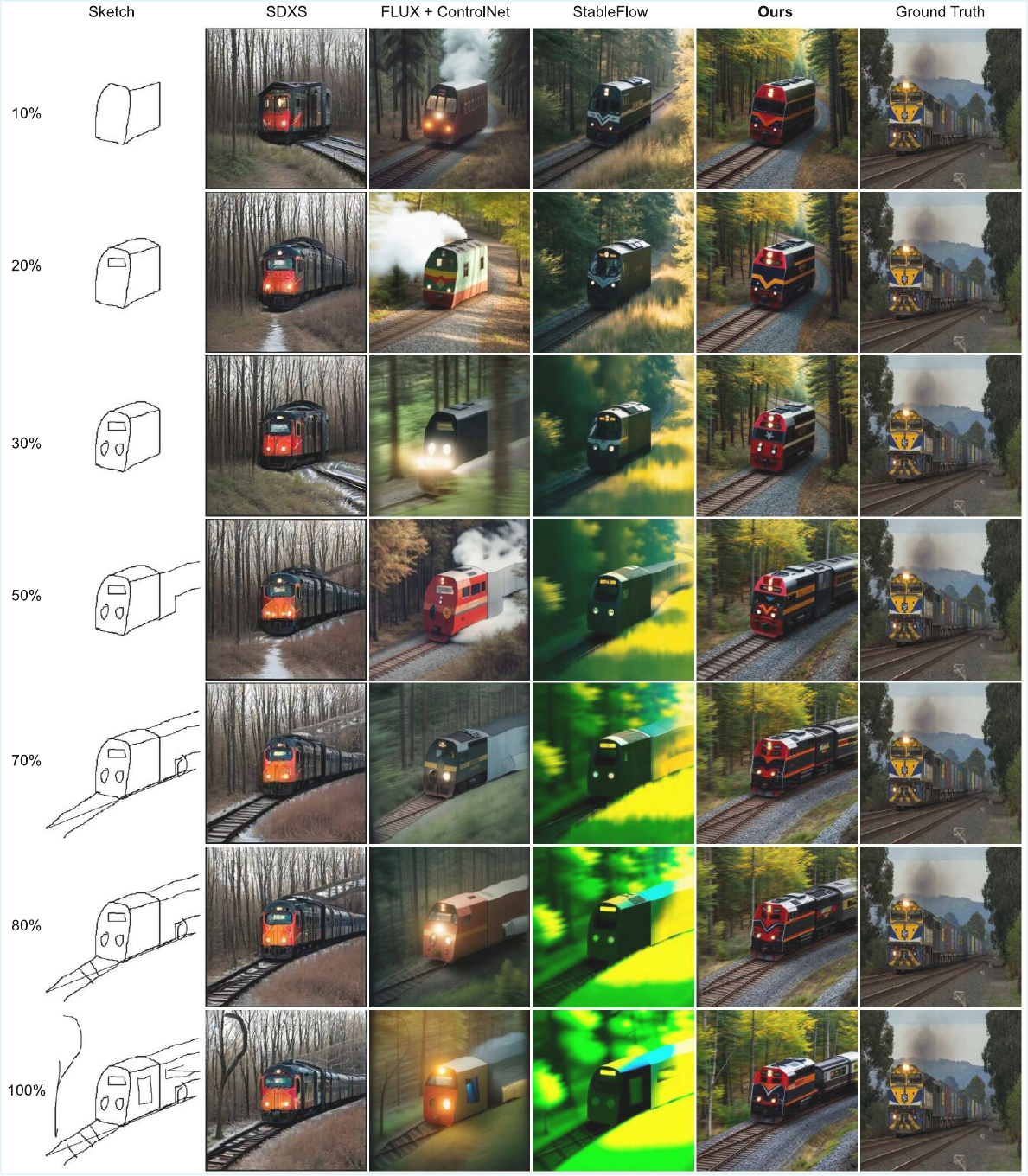}
  \caption{Progressive sketch-to-image generation. Prompt: \texttt{a train moving through the woods.} Both the baseline and SDXS contain visible artifacts, while StableFlow shows stronger green saturation. Our method adapts to the final tree-branch updates while retaining a consistent scene appearance.}
  \label{fig:supp_qual3}
\end{figure*}

\begin{figure*}[!htb]
  \centering
  \includegraphics[width=\linewidth]{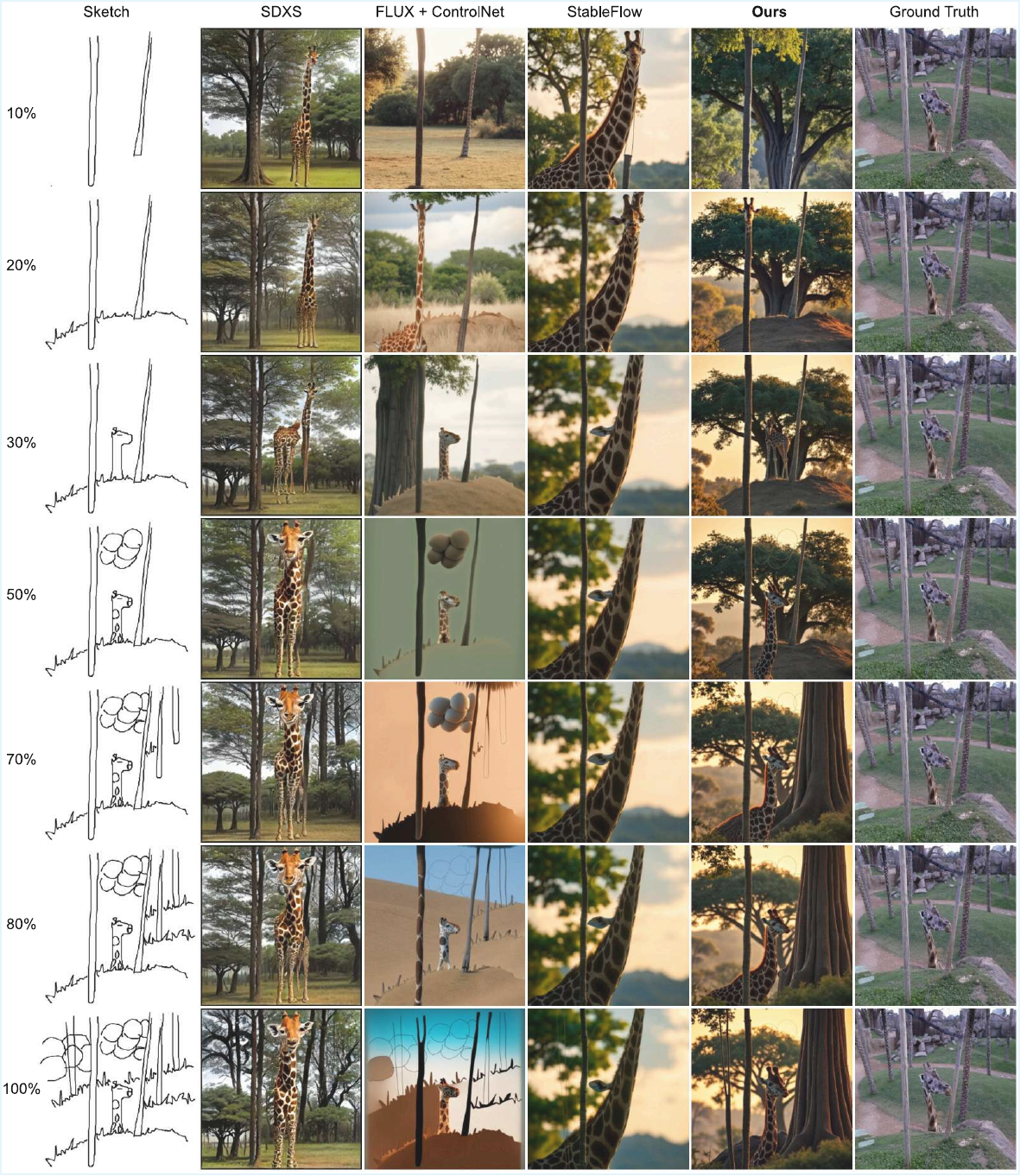}
  \caption{Progressive sketch-to-image generation. Prompt: \texttt{a giraffe is standing near a tree}. This is a challenging scenario where the evolving sketch initially differs from the prompt's general subject. StableFlow remains close to the initial structure, while SDXS shows limited adherence to later sketch updates and the baseline contains visible artifacts. Our model progressively adapts to the structural changes, following the initial tree strokes before forming the giraffe structure as additional evidence arrives.}
  \label{fig:supp_qual4}
\end{figure*}

\begin{figure*}[!htb]
  \centering
  \includegraphics[width=\linewidth]{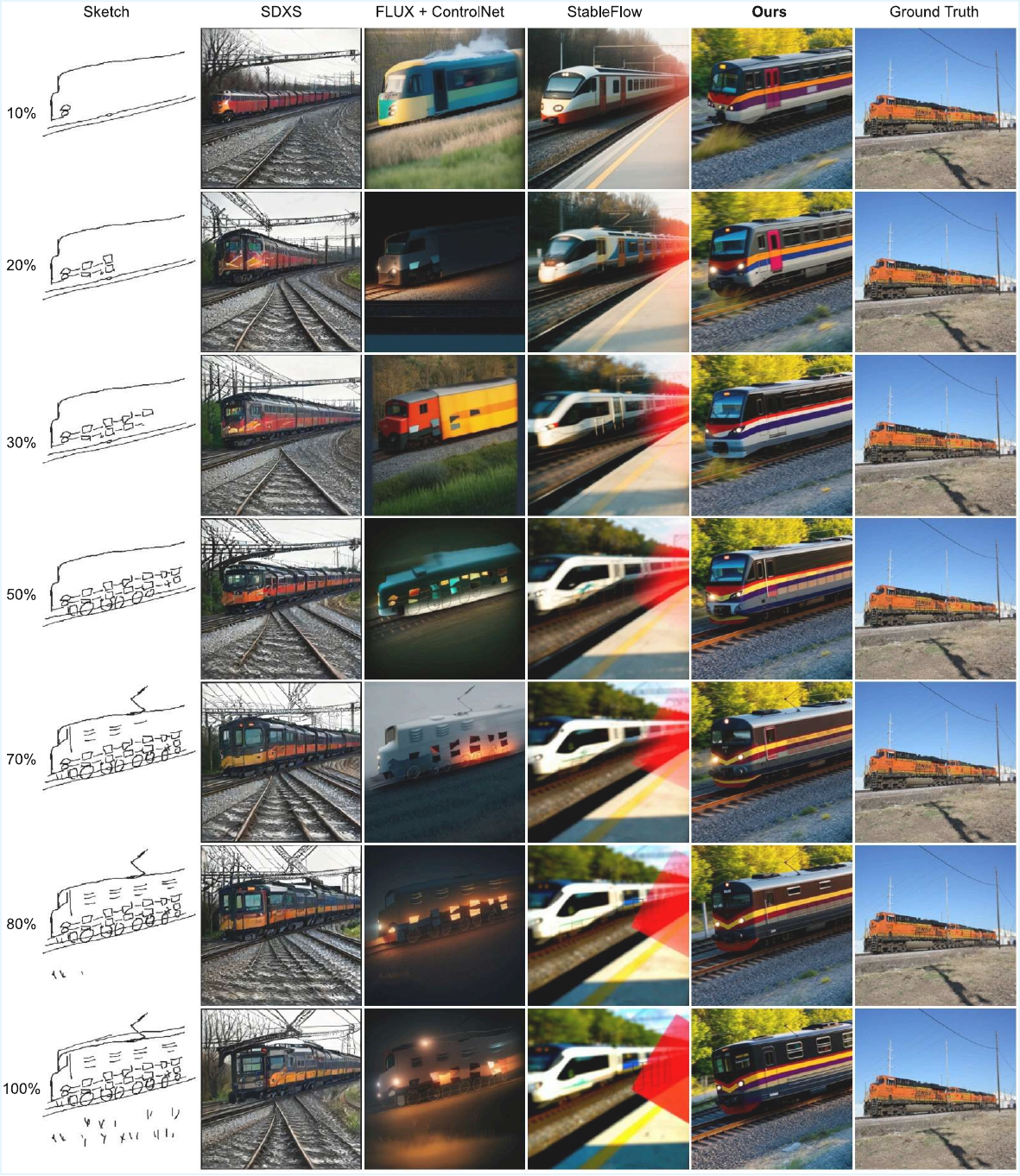}
  \caption{Progressive sketch-to-image generation. Prompt: \texttt{A train on a rail track}. StableFlow shows a reddish triangular artifact together with strong persistence of the initial structure and color palette. The baseline follows the sketch closely but produces a flat appearance, while SDXS preserves the main structure with limited detail. Our method progressively integrates specific details (e.g., the pantograph at 70\% completion) and preserves them across subsequent steps.}
  \label{fig:supp_qual5}
\end{figure*}

\begin{figure*}[!htb]
  \centering
  \includegraphics[width=\linewidth]{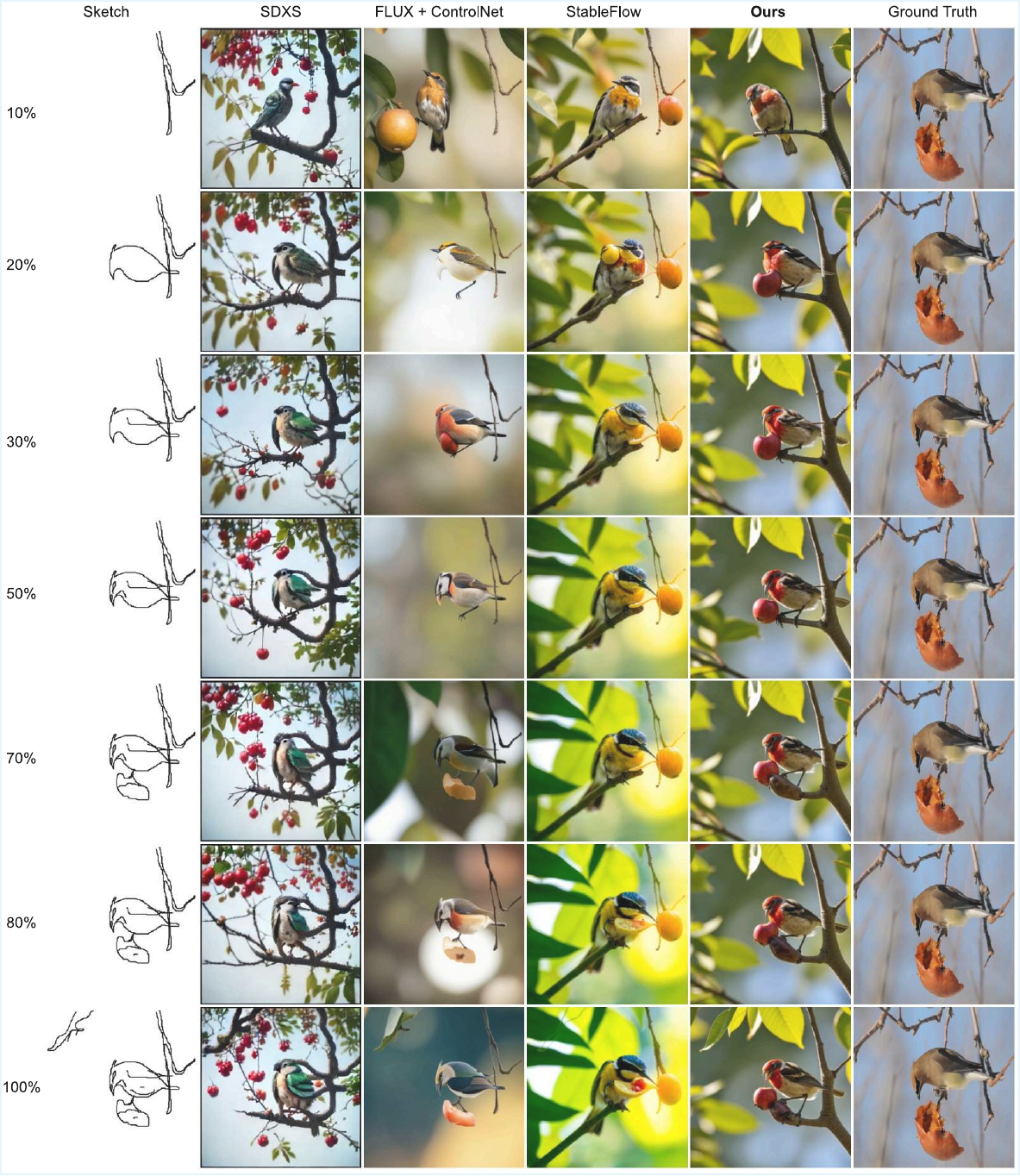}
  \caption{Progressive sketch-to-image generation. Prompt: \texttt{A bird is sitting on the branch of a tree  eating a fruit.} The baseline shows visible artifacts around imperfect strokes, while SDXS introduces extraneous elements. StableFlow exhibits stronger visual amplification, whereas our method balances the evolving sketch structure and prompt semantics while retaining the intended visual content.}
  \label{fig:supp_qual6}
\end{figure*}

\begin{figure*}[!htb]
  \centering
  \includegraphics[width=\linewidth]{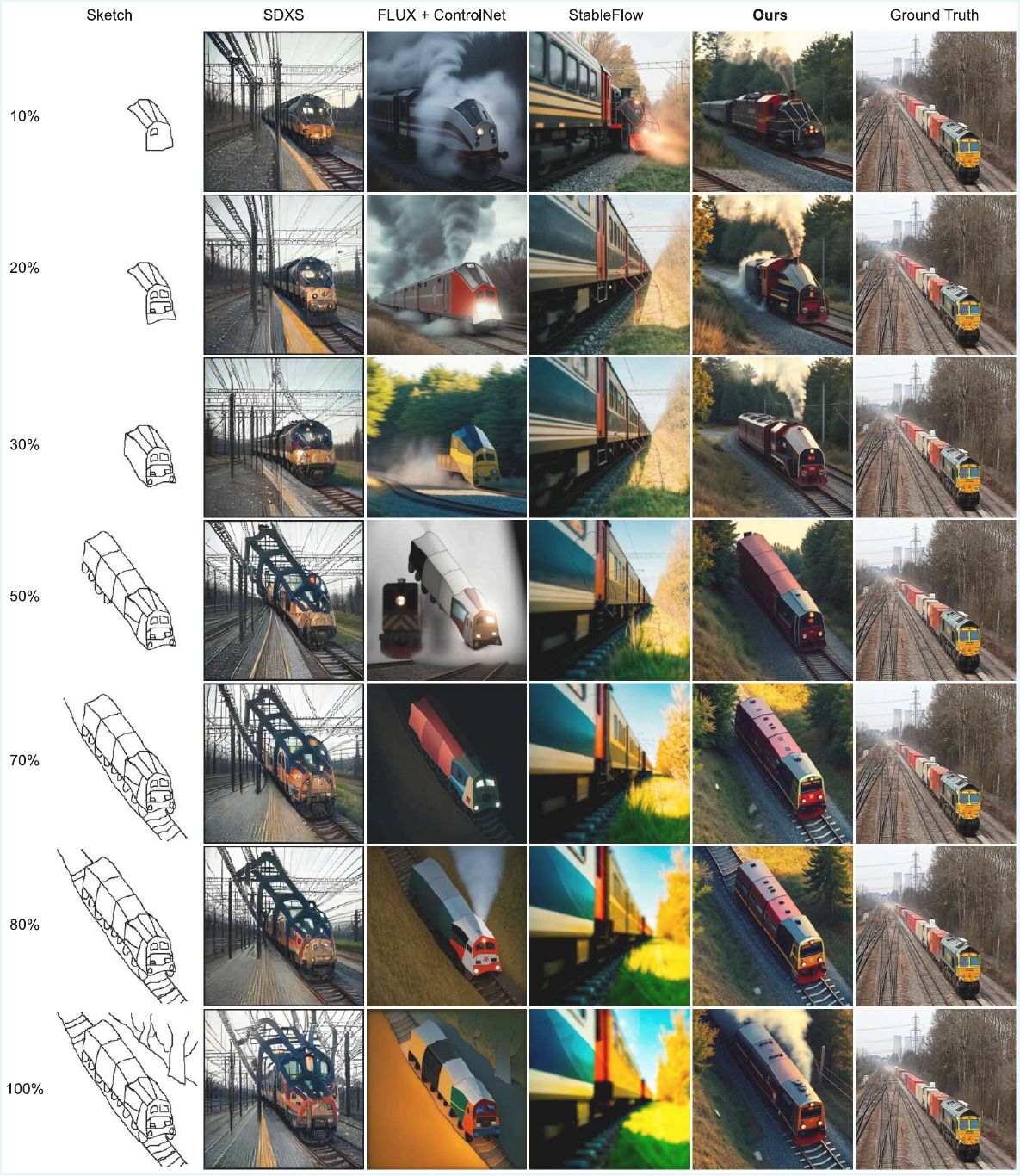}
  \caption{Progressive sketch-to-image generation. Prompt: \texttt{a train on the tracks.} Our method maintains a coherent progression across sketch completion levels while adapting the top-down view to later strokes. SDXS and the baseline show more visible artifacts, while StableFlow preserves the initial structure strongly and retains the saturated colors established at the 10\% stage.}
  \label{fig:supp_qual7}
\end{figure*}

\begin{figure*}[!htb]
  \centering
  \includegraphics[width=\linewidth]{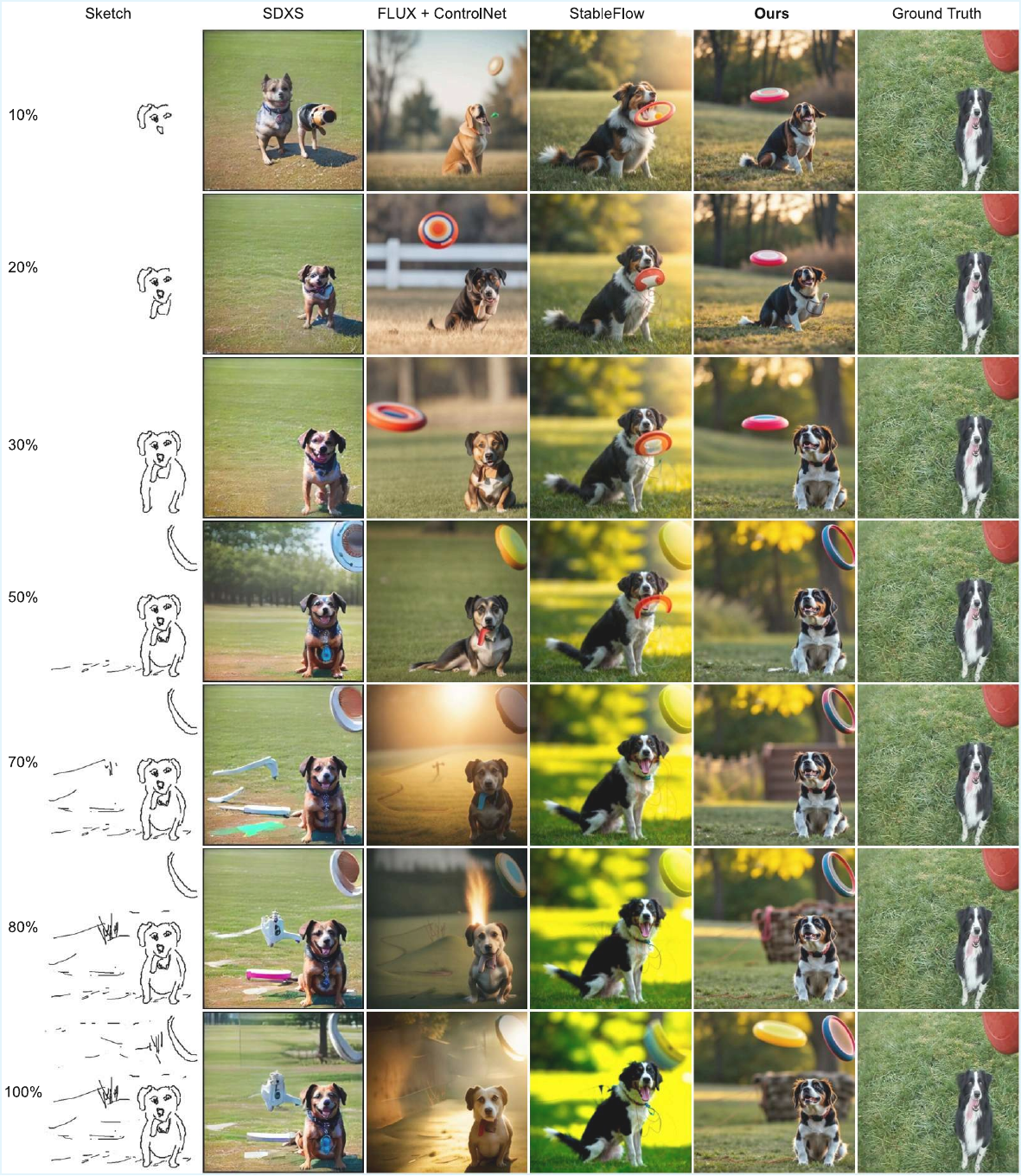}
  \caption{Progressive sketch-to-image generation. Prompt: \texttt{A dog sitting and eagerly waiting to catch the frisbee}. StableFlow shows stronger background saturation and visible artifacts at 80\% and 100\% progression. SDXS produces a high-quality image with some flawed details. Compared to the baseline, our method shows stronger sketch adherence while preserving the dog's facial structure.}
  \label{fig:supp_qual8}
\end{figure*}

\begin{figure*}[!htb]
  \centering
  \includegraphics[width=\linewidth]{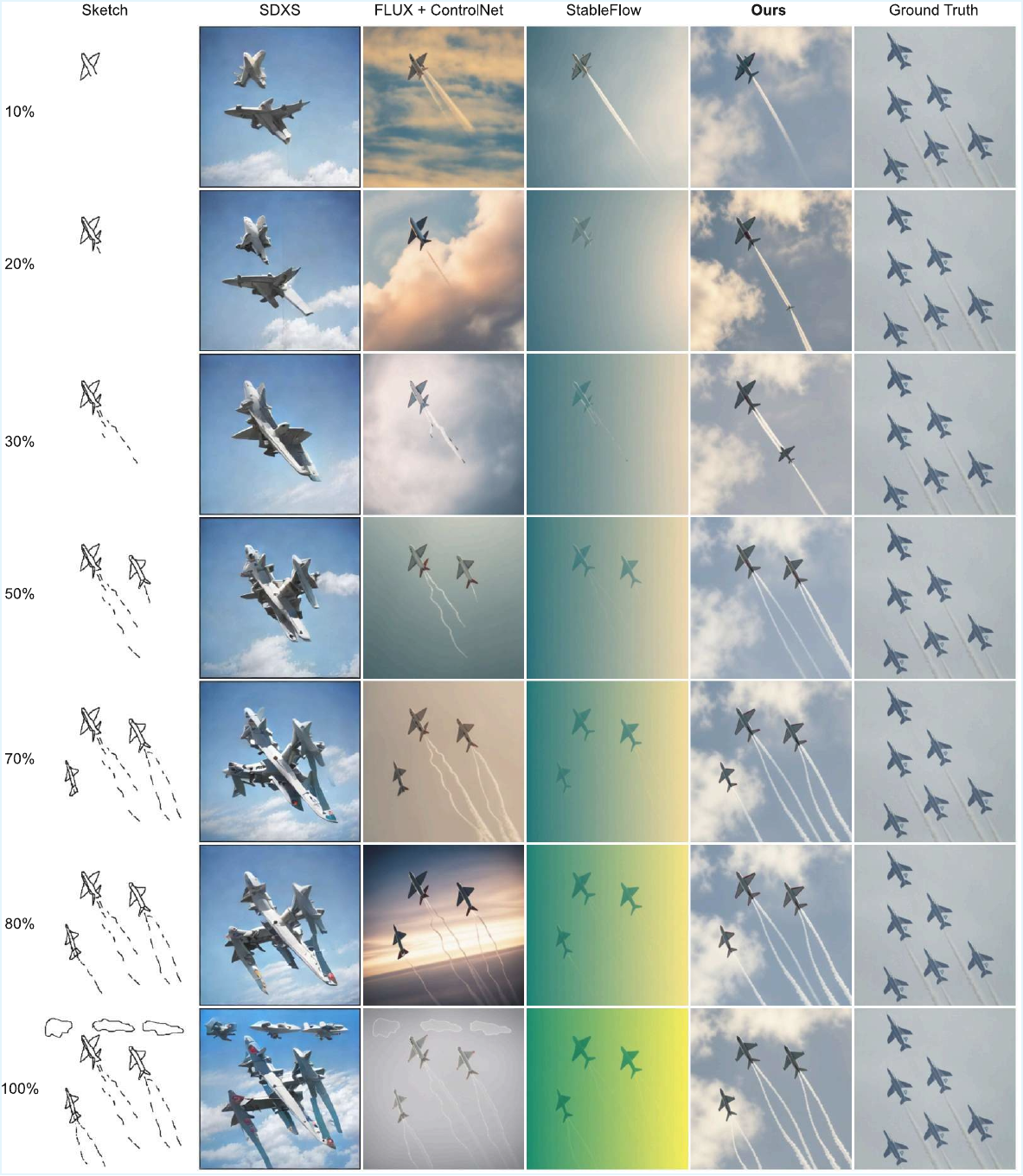}
  \caption{Progressive sketch-to-image generation. Prompt: \texttt{Jetplanes  flying in the sky}. StableFlow shows strong persistence of the initial structure together with color distortion and weaker adherence to later strokes. SDXS and the baseline contain visible artifacts under the evolving sketch condition. Our method adapts to the evolving sketch while preserving the emerging contrails and reducing visible stroke artifacts.}
  \label{fig:supp_qual9}
\end{figure*}

\begin{figure*}[!htb]
  \centering
  \includegraphics[width=\linewidth]{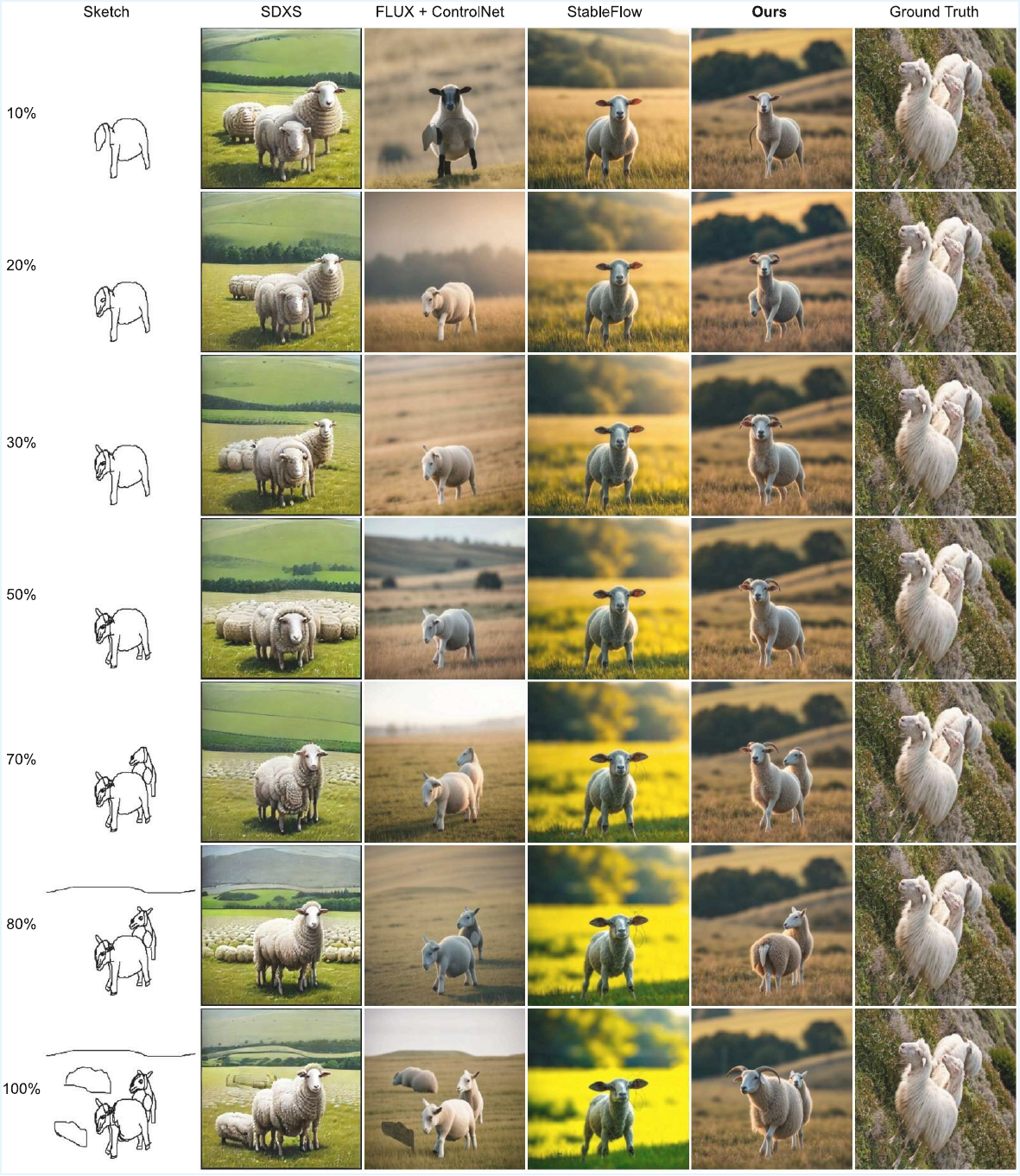}
  \caption{Progressive sketch-to-image generation. Prompt: \texttt{sheep standing in a field.} StableFlow remains focused on a single, greenish sheep, while the baseline follows the rough sketch structure but produces an unrealistic animal. SDXS produces a high-quality image with weaker preservation of the underlying sketch structure. Our method maintains sketch fidelity, introducing the second sheep at 70\% completion and refining the scene across later stages.}
  \label{fig:supp_qual10}
\end{figure*}

\begin{figure*}[!htb]
  \centering
  \includegraphics[width=\linewidth]{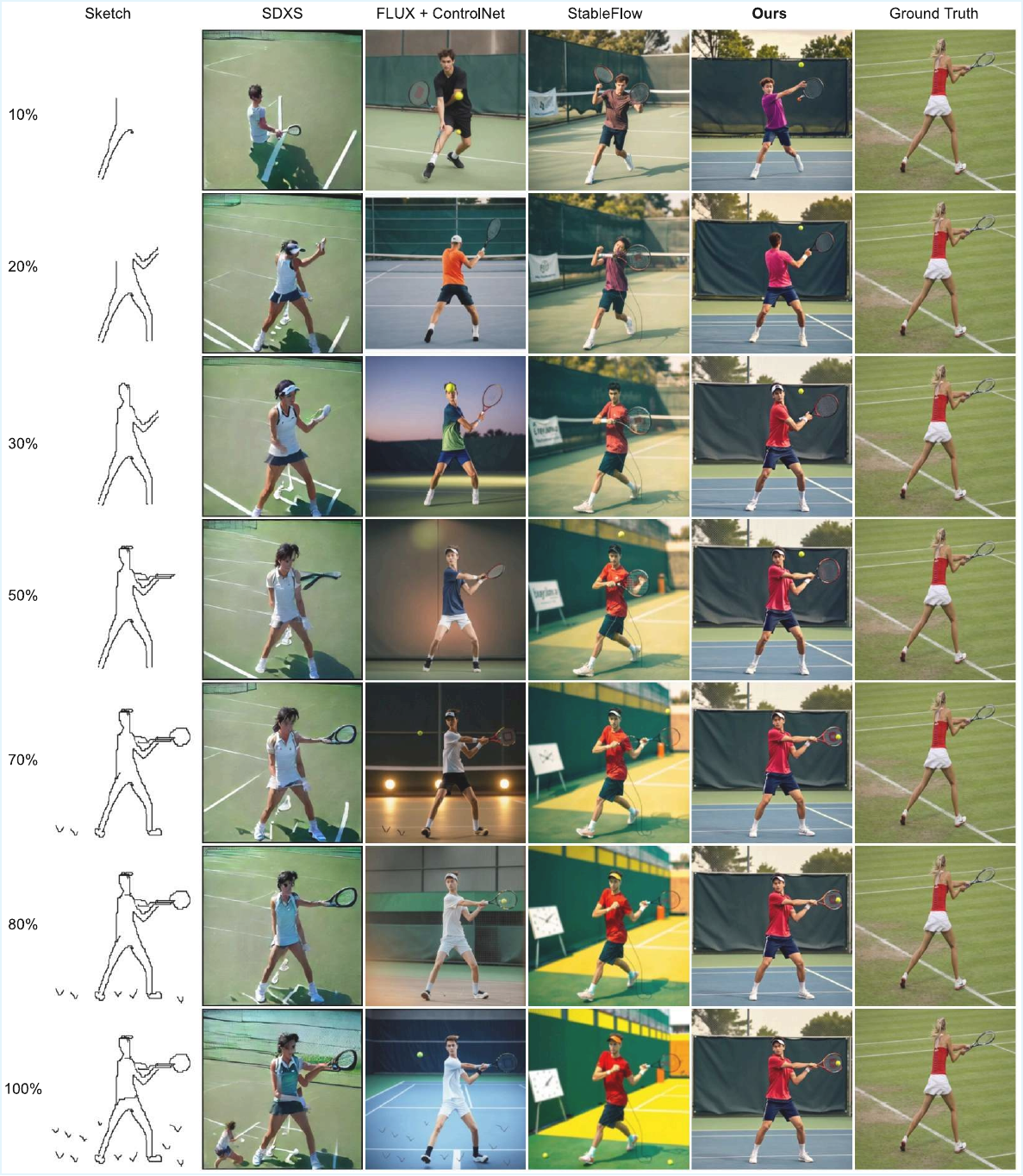}
  \caption{Progressive sketch-to-image generation. Prompt: \texttt{player playing a tennis game}. StableFlow shows strong persistence of structure and color together with visible artifacts, while SDXS follows the sketch more loosely and contains image-level artifacts. The baseline also shows substantial artifacts. Our method deviates slightly from the exact pose in the ground truth but retains the overall sketch structure with fewer visible artifacts.}
  \label{fig:supp_qual11}
\end{figure*}

\begin{figure*}[!htb]
  \centering
  \includegraphics[width=\linewidth]{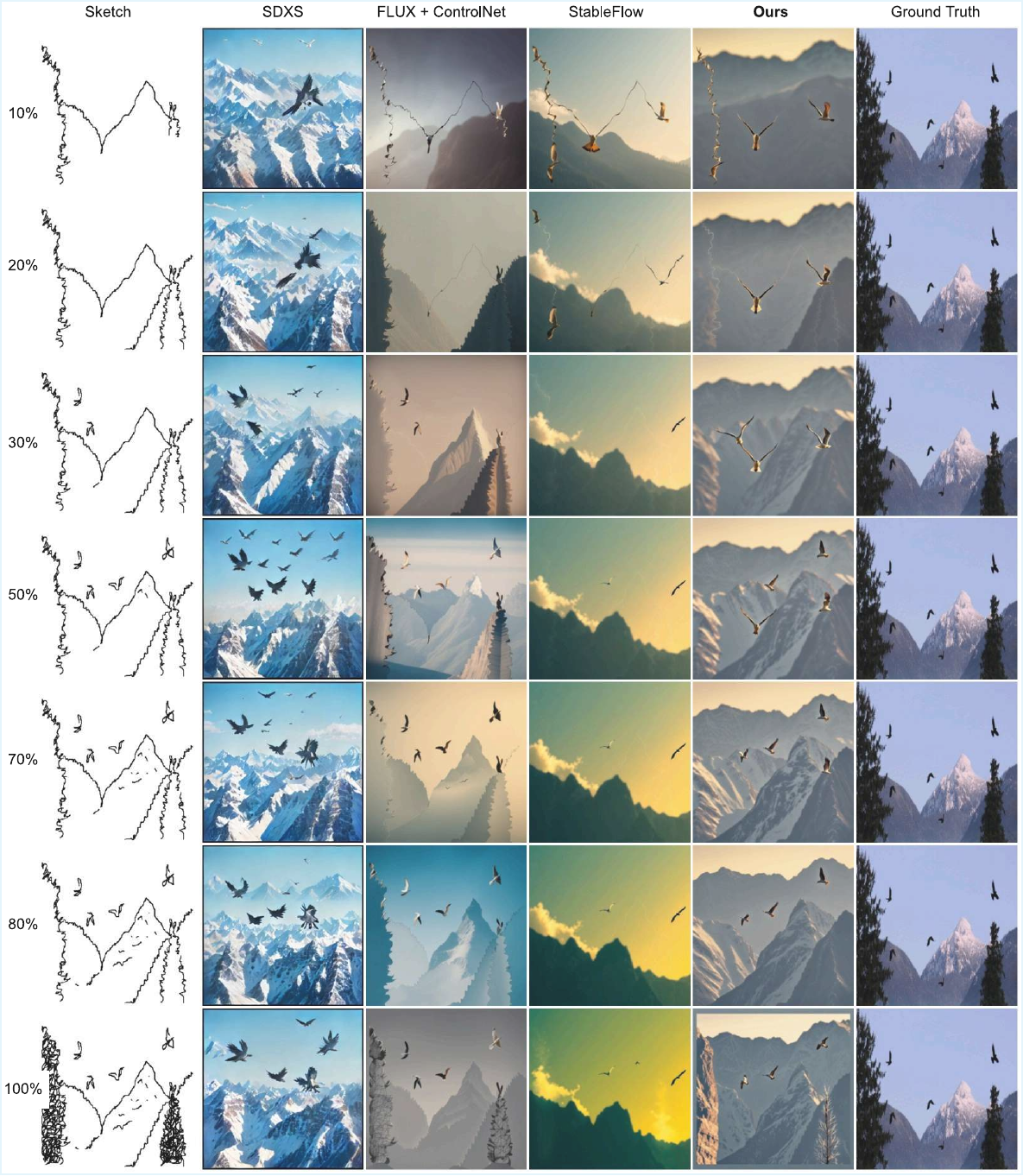}
  \caption{Progressive sketch-to-image generation. Prompt: \texttt{Few birds are flying over the mountains}. Our method forms birds from rough strokes while suppressing many of the initial stroke artifacts. The baseline reflects the sketch structure but yields a less photorealistic interpretation. StableFlow shows stronger color saturation and weaker agreement with the final sketch, while SDXS contains substantial image-level artifacts.}
  \label{fig:supp_qual12}
\end{figure*}

\begin{figure*}[!htb]
  \centering
  \includegraphics[width=\linewidth]{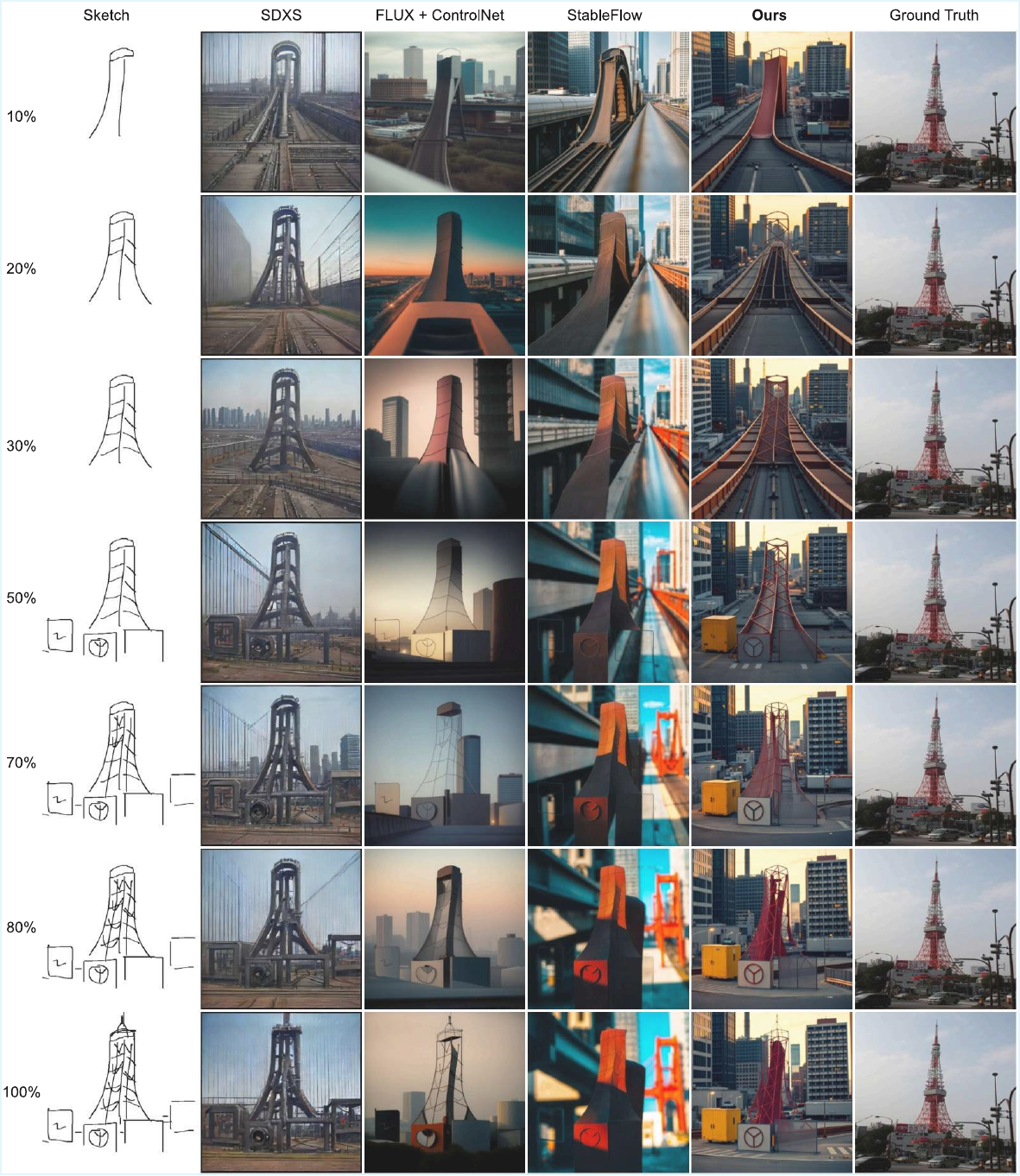}
  \caption{Progressive sketch-to-image generation. Prompt: \texttt{Industrial belt in the city}. This represents a challenging scenario where the prompt's open-ended nature makes interpretation difficult. Our method interprets the circular symbol and box-like structures consistently across the progression. The competing outputs show different limitations, including stronger saturation for StableFlow and visible artifacts for the baseline and SDXS.}
  \label{fig:supp_qual13}
\end{figure*}

\begin{figure*}[!htb]
  \centering
  \includegraphics[width=\linewidth]{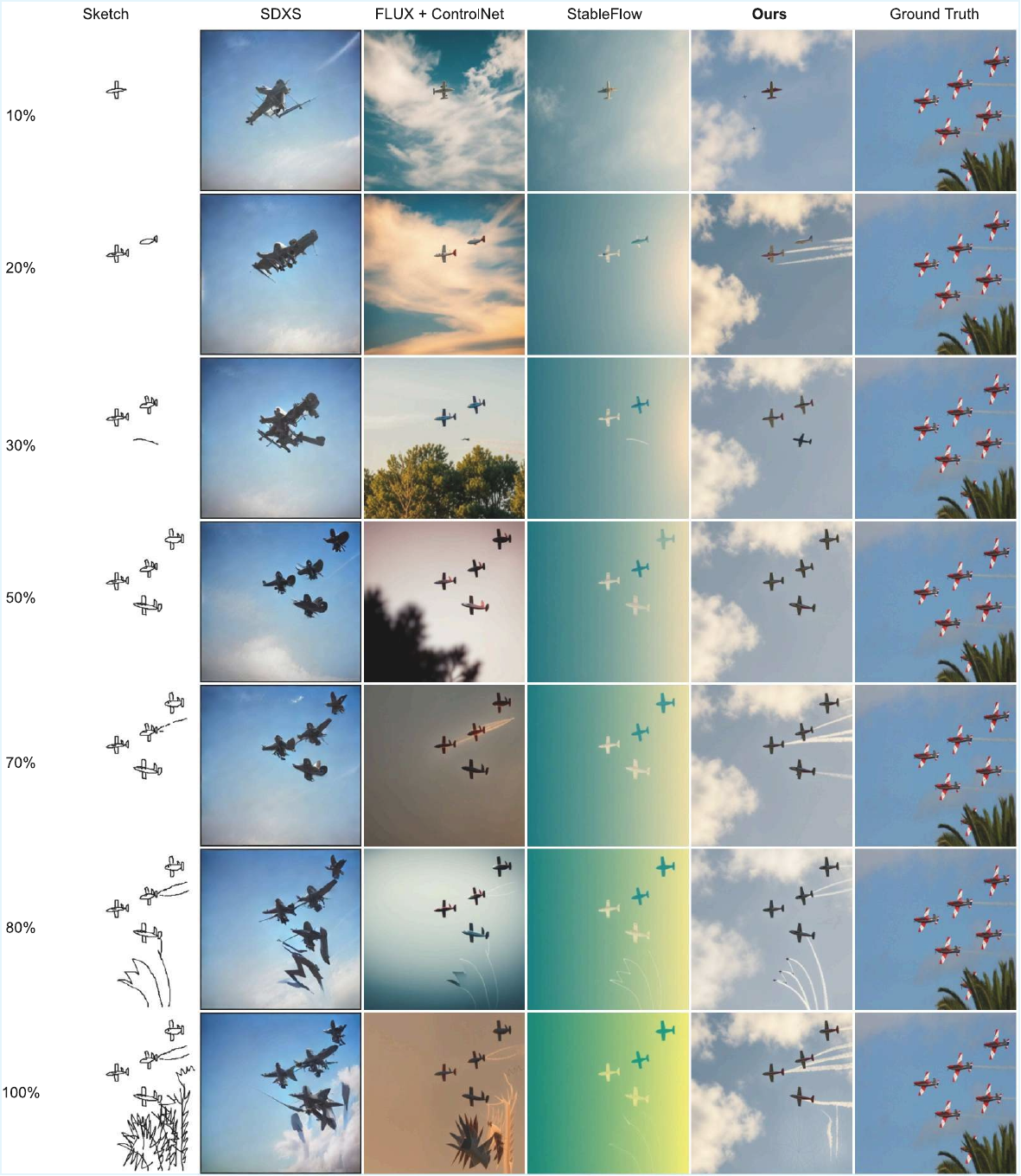}
  \caption{Progressive sketch-to-image generation. Prompt: \texttt{aeroplanes are flying in the sky}. StableFlow shows stronger color saturation, while the baseline contains visible imperfections around sketch-aligned regions. SDXS changes noticeably across turns and retains tree-like artifacts. Our method translates the evolving strokes into jets while gradually reducing the early tree-like structures.}
  \label{fig:supp_qual14}
\end{figure*}

\begin{figure*}[!htb]
  \centering
  \includegraphics[width=\linewidth]{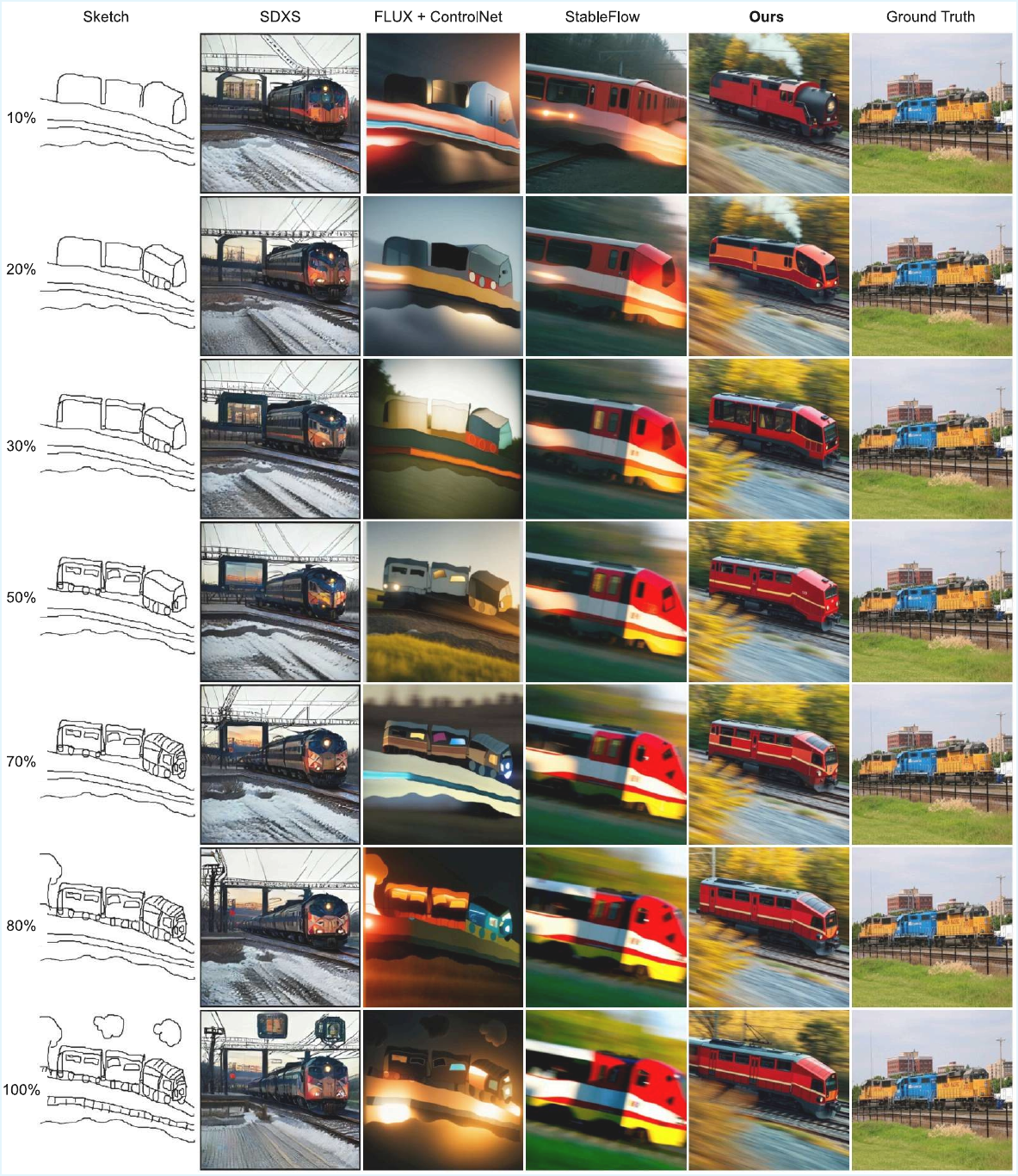}
  \caption{Progressive sketch-to-image generation. Prompt: \texttt{a train moving on tracks.} The baseline contains cartoon-like artifacts, while StableFlow strongly preserves the initial structure and shows limited later adaptation together with color distortion. SDXS shows weak sketch adherence. Our method provides smoother progression while balancing the evolving sketch with scene realism.}
  \label{fig:supp_qual15}
\end{figure*}

\begin{figure*}[!htb]
  \centering
  \includegraphics[width=\linewidth]{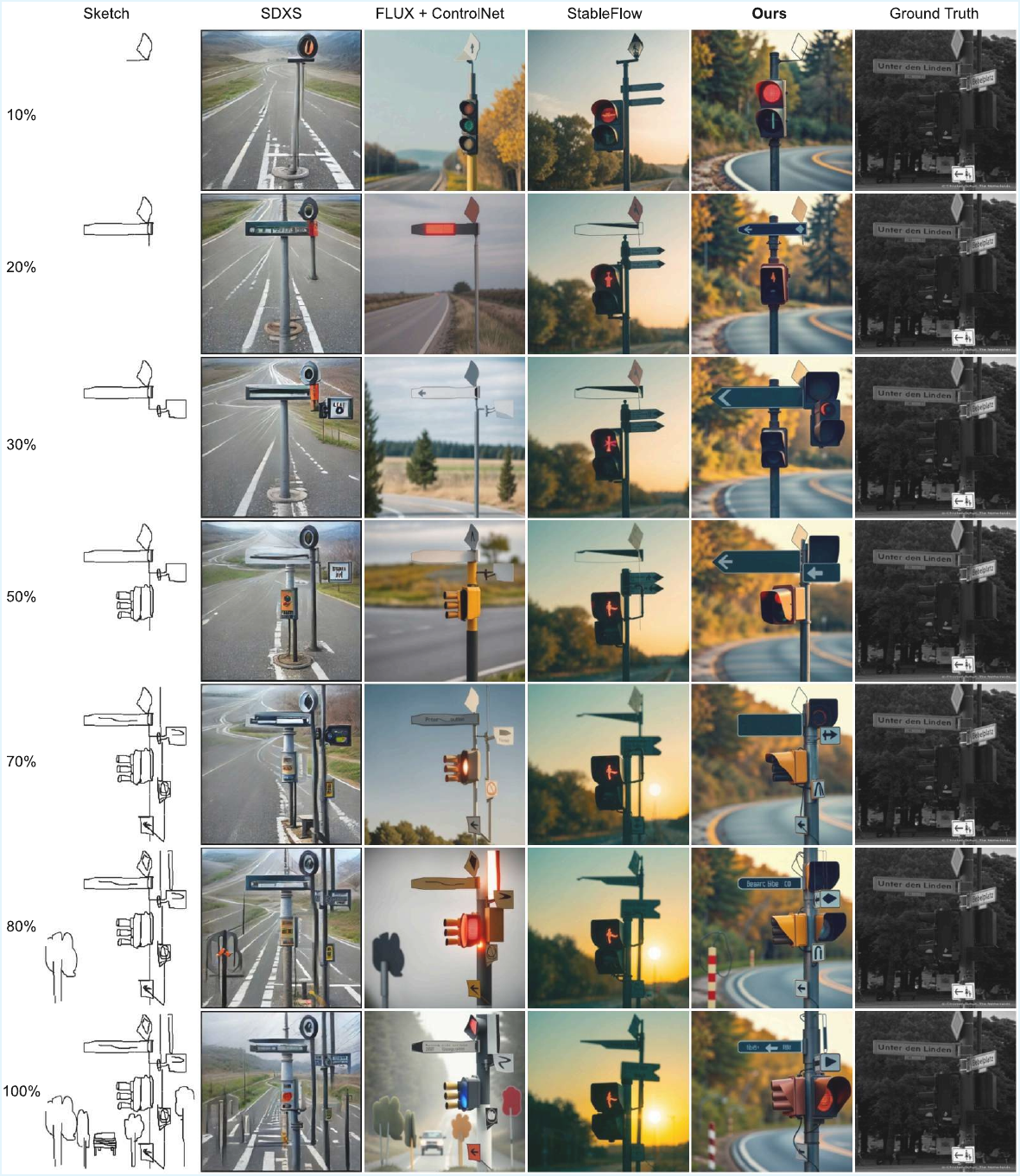}
  \caption{Progressive sketch-to-image generation. Prompt: \texttt{signal and direction pole on road}. In this scene-level generation, StableFlow strongly preserves the initial structure and shows color distortion as details are added. The baseline reflects several sketch imperfections in the final image, while SDXS contains limited fine detail. Our method incorporates progressive details while maintaining a coherent road scene with fewer visible artifacts.}
  \label{fig:supp_qual16}
\end{figure*}

\begin{figure*}[!htb]
  \centering
  \includegraphics[width=\linewidth]{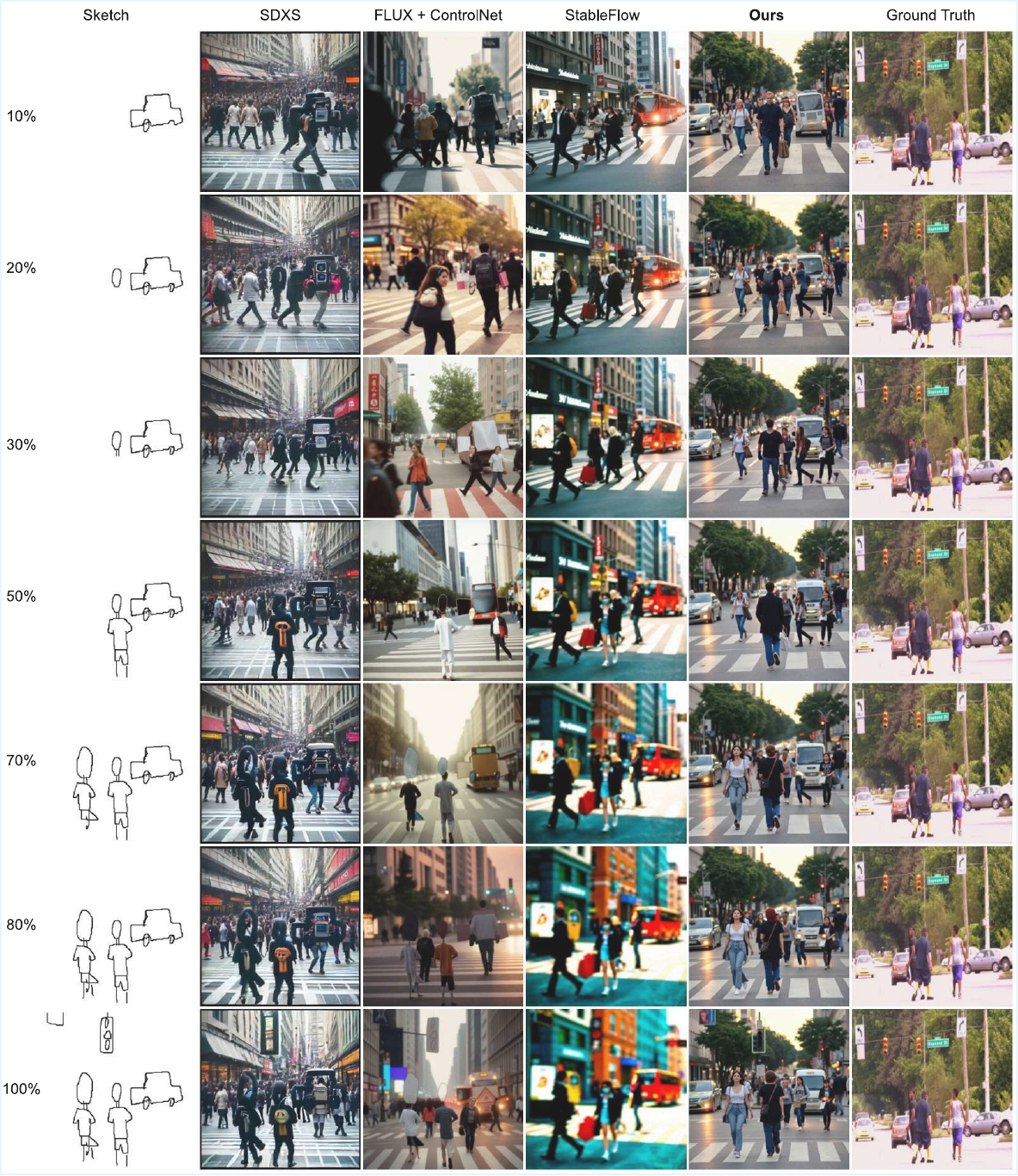}
  \caption{Progressive sketch-to-image generation. Prompt: \texttt{People are crossing the busy streets.} In this complex, multi-element scene, the baseline becomes more cartoon-like and contains visible artifacts under rough strokes. SDXS shows distorted figures, while StableFlow has stronger color saturation and adapts less to later strokes. Our method maintains a coherent traffic scene with central figures that remain aligned with the evolving strokes.}
  \label{fig:supp_qual17}
\end{figure*}

\begin{figure*}[!htb]
  \centering
  \includegraphics[width=\linewidth]{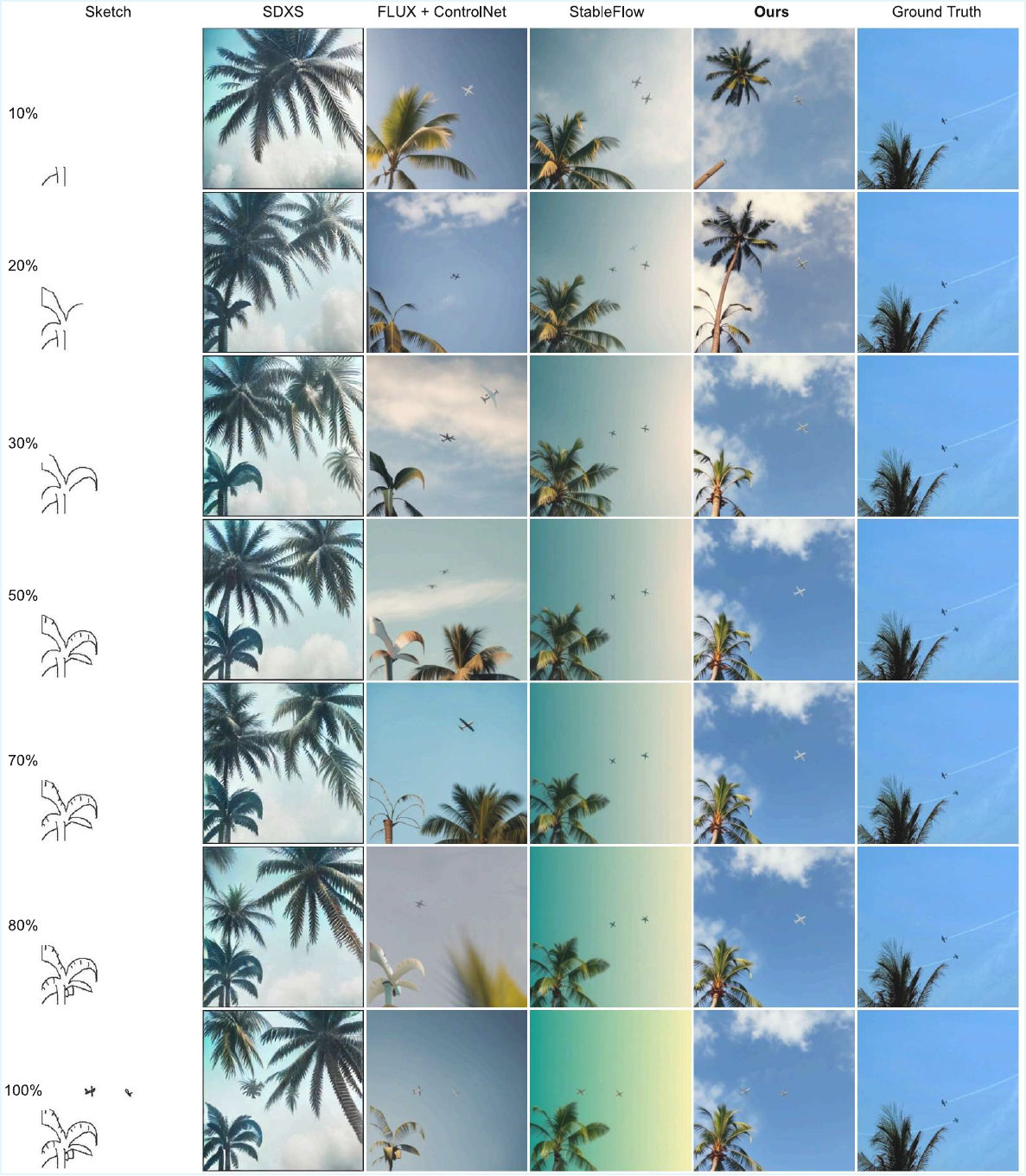}
  \caption{Progressive sketch-to-image generation. Prompt: \texttt{Two planes flying in sky where a coconut tree is also in the frame.} StableFlow exhibits color distortion while maintaining the broad sketch structure, the baseline shows artifacts in the tree leaves, and SDXS shows weaker sketch adherence. Our method progresses from a single airplane to multiple planes as the corresponding strokes are introduced.}
  \label{fig:supp_qual18}
\end{figure*}

\begin{figure*}[!htb]
  \centering
  \includegraphics[width=\linewidth]{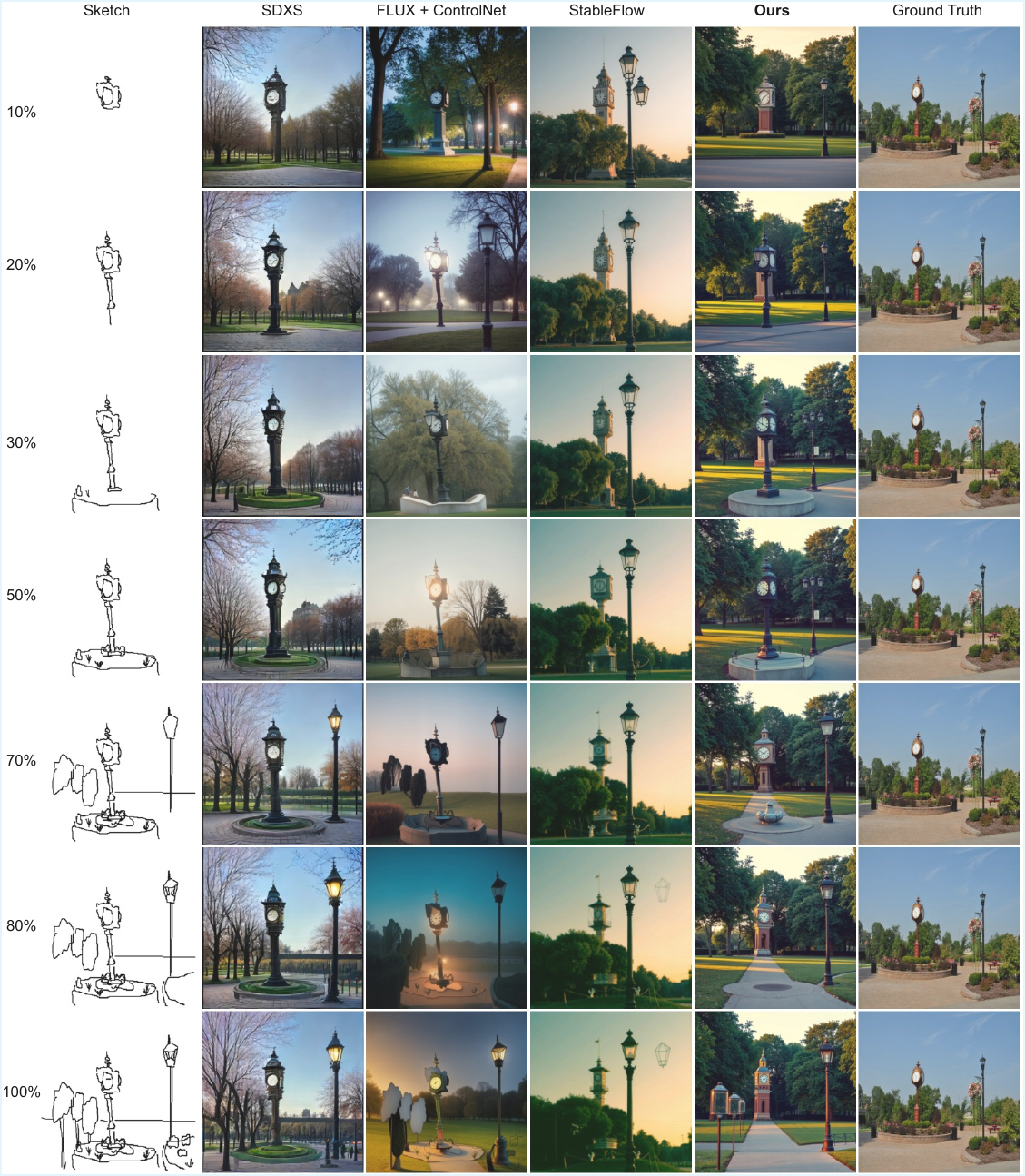}
  \caption{Progressive sketch-to-image generation. Prompt: \texttt{a park with clock tower and a streetlamp nearby } This example features a vague description of a park; as the user draws trees (despite the prompt focusing on a park and clock), competing methods drift and generate visible artifacts under the evolving sketch condition. Our method demonstrates flexibility by interpreting rough tree strokes as lamp-like structures appropriate for a park scene, while smoothly incorporating user details added at the 70\% stage.}
  \label{fig:supp_qual19}
\end{figure*}

\begin{figure*}[!htb]
  \centering
  \includegraphics[width=\linewidth]{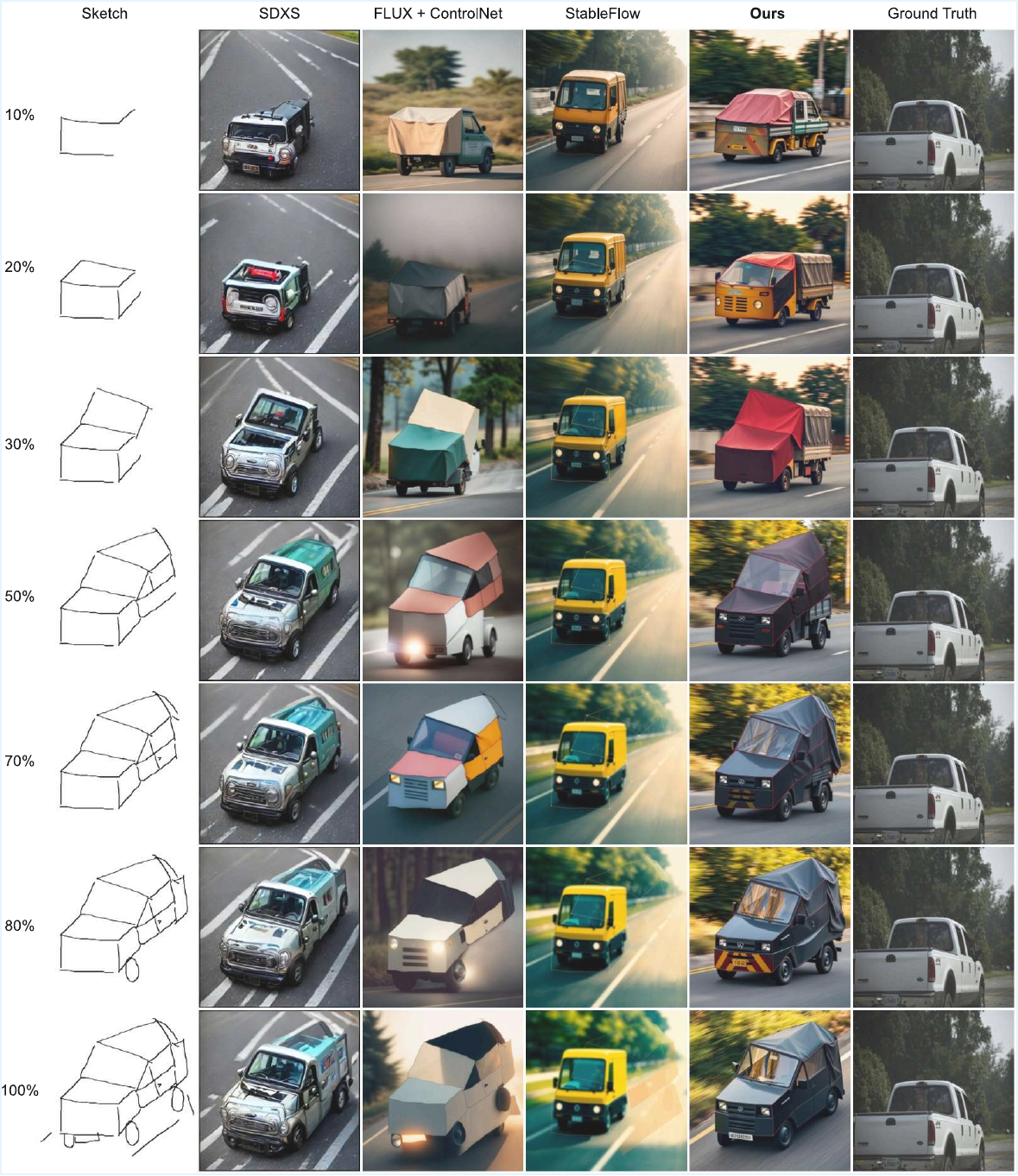}
  \caption{Progressive sketch-to-image generation. Prompt: \texttt{A mini truck on road}. This is a challenging task because the upper section of the truck is only partially specified. Our method interprets the incomplete strokes as a plausible vehicle structure and updates it as more evidence arrives. The competing outputs show different behaviors: the baseline contains substantial artifacts, StableFlow strongly preserves the initial structure, and SDXS preserves the overall intent with less fine detail. Our approach updates the structure as new strokes are provided.}
  \label{fig:supp_qual20}
\end{figure*}

\begin{figure*}[!htb]
  \centering
  \includegraphics[width=\linewidth]{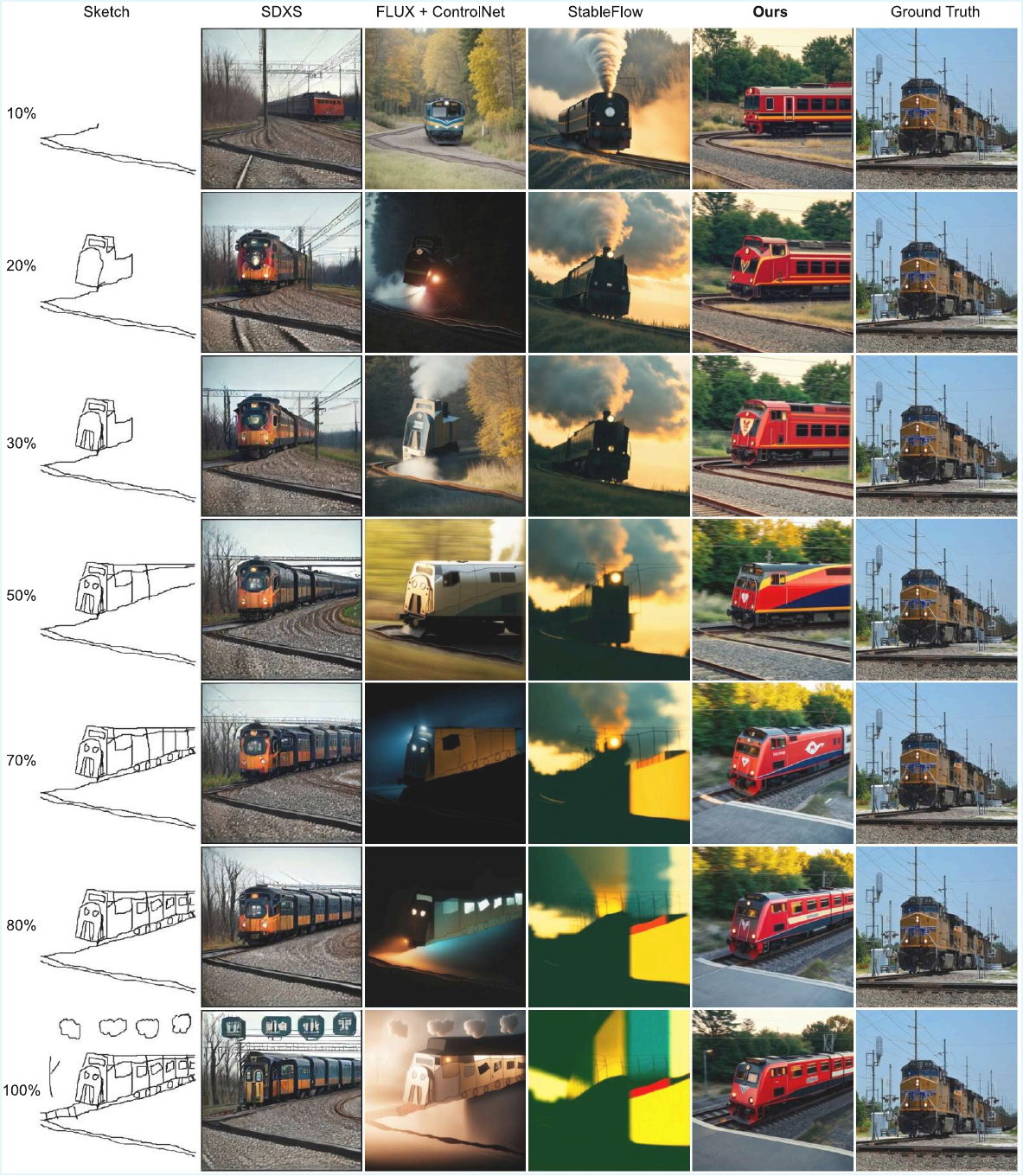}
  \caption{Progressive sketch-to-image generation. Prompt: \texttt{a train on a track.} StableFlow strongly preserves the initial structure and exhibits substantial color amplification. SDXS retains the broad structure but shows limited response to small corrections, while the baseline contains heavy artifacts. Our method forms a coherent train structure and extends the sky when additional cloud strokes are introduced.}
  \label{fig:supp_qual21}
\end{figure*}

\begin{figure*}[!htb]
  \centering
  \includegraphics[width=\linewidth]{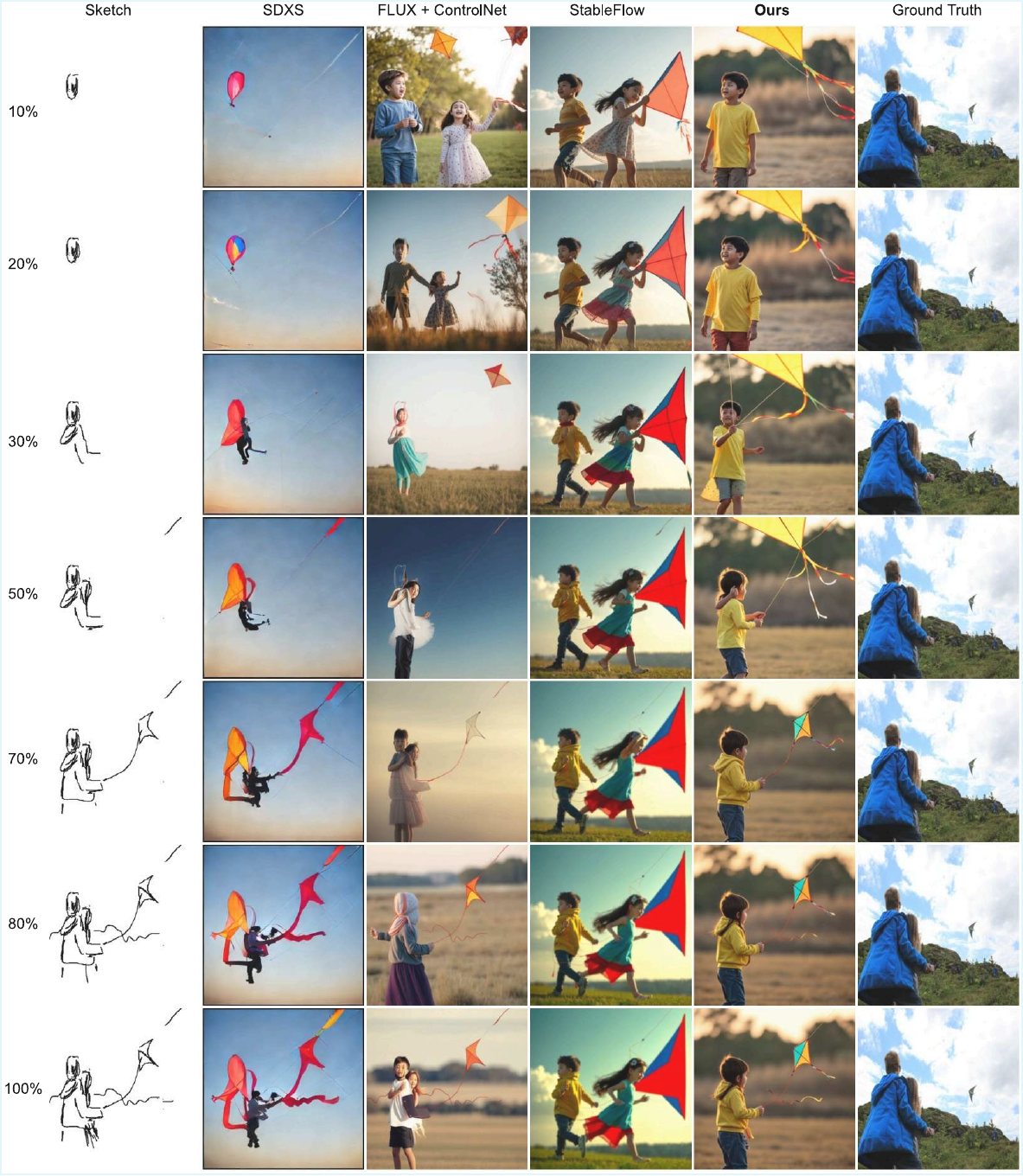}
  \caption{Progressive sketch-to-image generation. Prompt: \texttt{A boy and a girl flying a kite}. This case is challenging because the sketch strokes overlap heavily. Heavy stroke overlap remains challenging at the final stages, although the overall scene composition remains photorealistic and temporally stable. SDXS, the baseline, and StableFlow show more drift and visible artifacts, while our method remains comparatively stable in overall composition.}
  \label{fig:supp_qual22}
\end{figure*}

\begin{figure*}[!htb]
  \centering
  \includegraphics[width=\linewidth]{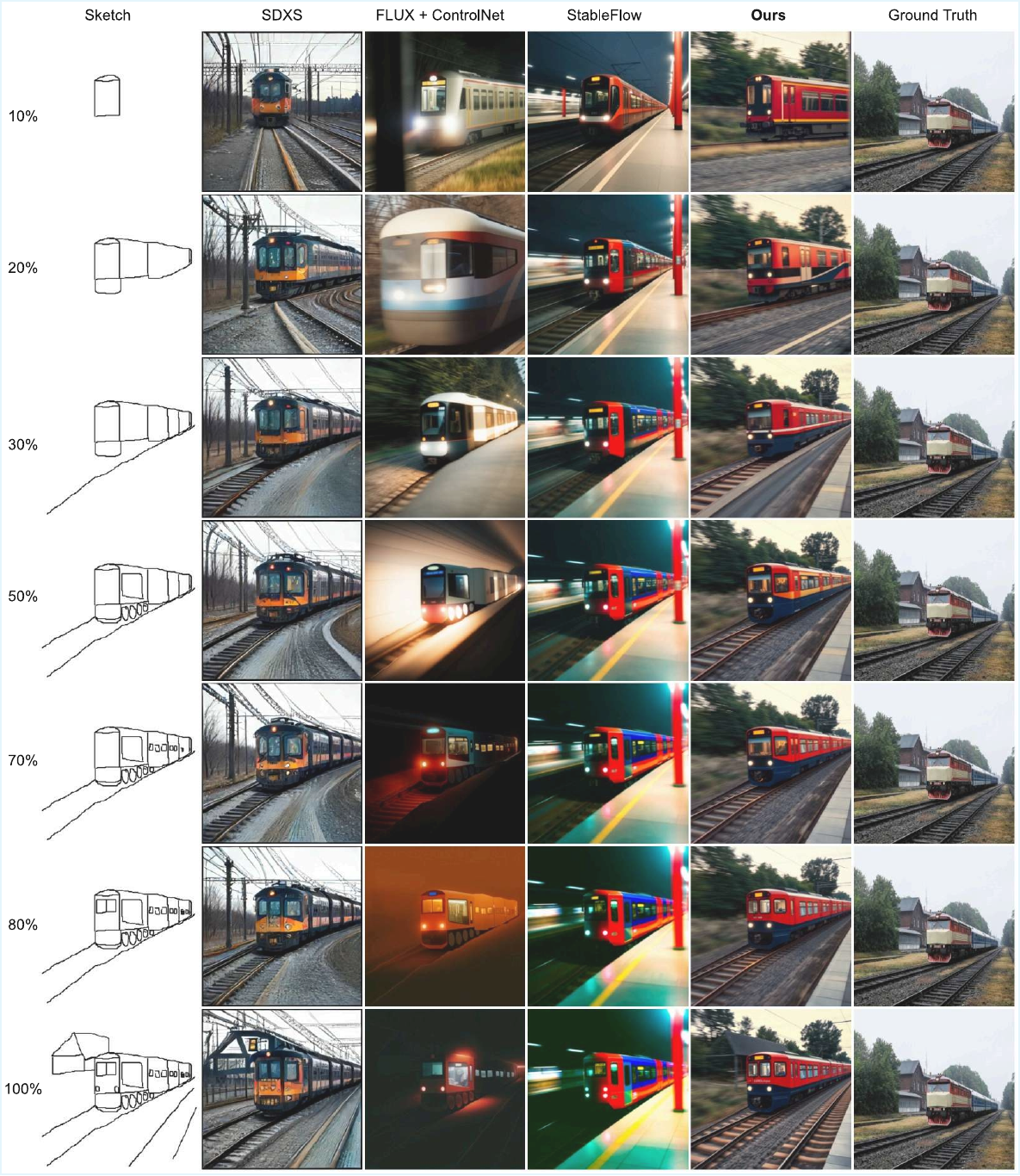}
  \caption{Progressive sketch-to-image generation. Prompt: \texttt{A train moving  on  track}. This demonstrates faithful generation where our method progressively incorporates an extra line intended by the user at 100\% completion. SDXS also reacts to the added line but loses some fine detail, whereas our method integrates the new information while preserving more of the established scene.}
  \label{fig:supp_qual23}
\end{figure*}
\clearpage

\end{document}